\documentstyle[11pt,epsf,theapa]{article}
\newtheorem{theorem}{Theorem}

\newtheorem{lemma}{Lemma}
\newtheorem{definition}{Definition}
\newtheorem{corollary}{Corollary}
\def\limplies{\; \Rightarrow \;}

\def\lforall#1{\forall \: #1 \;}

\def\comment#1{}

\def\tuple#1{$\langle #1\rangle$}
\def\mtuple#1{\langle #1\rangle}

\def\centerps#1{\begin{center}
\leavevmode
\epsfbox{#1}
\end{center}}
\def\argmax{\mathop{\rm argmax}}

\def\qed{{\bf Q.E.D.}\vspace{.1in}}

\begin{document}
\setcounter{page}{1}
\title{Hierarchical Reinforcement Learning with the MAXQ Value
Function Decomposition}
\author{Thomas G. Dietterich\\
Department of Computer Science\\
Oregon State University\\
Corvallis, OR 97331\\
{\tt tgd@cs.orst.edu}}
\date{\today}

\maketitle

\begin{abstract}
\noindent This paper presents a new approach to hierarchical
reinforcement learning based on decomposing the target Markov decision
process (MDP) into a hierarchy of smaller MDPs and decomposing the
value function of the target MDP into an additive combination of the
value functions of the smaller MDPs.  The decomposition, known as the
MAXQ decomposition, has both a procedural semantics---as a subroutine
hierarchy---and a declarative semantics---as a representation of the
value function of a hierarchical policy.  MAXQ unifies and extends
previous work on hierarchical reinforcement learning by Singh,
Kaelbling, and Dayan and Hinton.  It is based on the assumption that
the programmer can identify useful subgoals and define subtasks that
achieve these subgoals.  By defining such subgoals, the programmer
constrains the set of policies that need to be considered during
reinforcement learning.  The MAXQ value function decomposition can
represent the value function of any policy that is consistent with the
given hierarchy.  The decomposition also creates opportunities to
exploit state abstractions, so that individual MDPs within the
hierarchy can ignore large parts of the state space.  This is
important for the practical application of the method.  This paper
defines the MAXQ hierarchy, proves formal results on its
representational power, and establishes five conditions for the safe
use of state abstractions.  The paper presents an online model-free
learning algorithm, MAXQ-Q, and proves that it converges wih
probability 1 to a kind of locally-optimal policy known as a
recursively optimal policy, even in the presence of the five kinds of
state abstraction.  The paper evaluates the MAXQ representation and
MAXQ-Q through a series of experiments in three domains and shows
experimentally that MAXQ-Q (with state abstractions) converges to a
recursively optimal policy much faster than flat Q learning.  The fact
that MAXQ learns a representation of the value function has an
important benefit: it makes it possible to compute and execute an
improved, non-hierarchical policy via a procedure similar to the
policy improvement step of policy iteration.  The paper demonstrates
the effectiveness of this non-hierarchical execution experimentally.
Finally, the paper concludes with a comparison to related work and a
discussion of the design tradeoffs in hierarchical reinforcement
learning.
\end{abstract}

\newpage

\section{Introduction}

A central goal of artificial intelligence is to develop techniques for
constructing robust, autonomous agents that are able to achieve good
performance in complex, real-world environments.  One fruitful line of
research views agents from an ``economic'' perspective
\cite{bsw-epmas-97}: An agent interacts with an environment and
receives real-valued rewards and penalties.  The agent's goal is to
maximize the total reward it receives.  The economic view makes it
easy to formalize traditional goals of achievement (``land this
airplane'').  But it also makes it easy to formulate goals of
prevention (``don't crash into any other airplanes'') and goals of
maintenance (``keep the air-traffic control system working as long as
possible'').  Goals of achievement can be represented by giving a
positive reward for achieving the goal.  Goals of prevention can be
represented by giving a negative reward when bad events occur, and
goals of maintenance can be represented by giving a positive reward
for each time step that the desireable state is maintained.
Furthermore, the economic formalism makes it possible to incorporate
uncertainty---we can require the agent to maximize the {\it expected
value} of the total reward in the face of random events in the world.

This brief review shows that the economic approach is very
expressive---a difficult research challenge, however, is to develop
efficient and scalable methods for reasoning, planning, and learning
within the economic AI framework.  The area of Stochastic Planning
studies methods for finding optimal or near-optimal plans to maximize
expected total reward in the case where the agent has complete
knowledge of the probabilistic behavior of the environment and the
reward function.  The basic methods for this case were developed in
the 1950s in the field of ``Dynamic Programming.''  Unfortunately,
these methods require time polynomial in the number of states in the
state space, which makes them prohibitively expensive for most AI
problems.  Hence, recent research has focused on methods
 that can exploit structure within the planning
problem to work more efficiently \cite{bdh-dtp:sacl-99}.

The area of Reinforcement Learning \cite{bt-ndp-96,sb-rl-98} studies
methods for learning optimal or near-optimal plans by interacting
directly with the external environment (as opposed to analyzing a
user-provided model of the environment).  Again, the basic methods in
reinforcement learning are based on dynamic programming algorithms.
However, reinforcement learning methods offer two important advantages
over classical dynamic programming.  First, the methods are online.
This permits them to focus their attention on the parts of the state
space that are important and ignore the rest of the space.  Second,
the methods can employ function approximation algorithms (e.g., neural
networks) to represent their knowledge.  This allows them to
generalize across the state space so that the learning time scales
much better.

Despite the recent advances in both probabilistic planning and
reinforcement learning, there are still many shortcomings.  The
biggest of these is the lack of a fully satisfactory method for
incorporating hierarchies into these algorithms.  Research in
classical planning has shown that hierarchical methods such as
hierarchical task networks \cite{aij:CurrieTate91}, macro actions
\cite{fhn-legrp-72,k-mo:wml-85}, and state abstraction methods
\cite{s-phas-74,knoblock-90} can provide exponential reductions in the
computational cost of finding good plans.  However, all of the basic
algorithms for probabilistic planning and reinforcement learning are
``flat'' methods---they treat the state space as one huge flat search
space.  This means that the paths from the start state to the goal
state are very long, and the length of these paths determines the cost
of learning and planning, because information about future rewards
must be propagated backward along these paths.

Many researchers
\cite{s-tlcsest-92,l-rlrnn-93,k-hrl:pr-93,dh-frl-93,hmbkd-hsmdpm-98,pr-rlhm-98,sps-bmsm:lprkmts-98}
have experimented with different methods of hierarchical reinforcement
learning and hierarchical probabilistic planning.  This research has
explored many different points in the design space of hierarchical
methods, but several of these systems were designed for specific
situations.  We lack crisp definitions of the main approaches and a
clear understanding of the relative merits of the different methods.

This paper formalizes and clarifies one approach and attempts to
understand how it compares with the other techniques.  The approach,
called the MAXQ method, provides a hierarchical decomposition of the
given reinforcement learning problem into a set of subproblems.  It
simultaneously provides a decomposition of the value function for the
given problem into a set of value functions for the subproblems.
Hence, it has both a declarative semantics (as a value function
decomposition) and a procedural semantics (as a subroutine hierarchy). 

A review of previous research shows that there are several important
design decisions that must be made when constructing a hierarchical
reinforcement learning system.  As a way of providing an overview of
the results in this paper, let us review these issues and see how the
MAXQ method approaches each of them.

The first issue is how subtasks should be specified.  Hierarchical
reinforcement learning involves breaking the target Markov decision
problem into a hierarchy of subproblems or subtasks.  There are three
general approaches to defining these subtasks.  One approach is to
define each subtask in terms of a fixed policy that is provided by the
programmer.  The ``option'' method of Sutton, Precup, and Singh
\citeyear{sps-bmsm:lprkmts-98} takes this approach.  The second
approach is to define each subtask in terms of a non-deterministic
finite-state controller.  The Hierarchy of Abstract Machines (HAM)
method of Parr and Russell \citeyear{pr-rlhm-98} takes this approach.
This method permits the programmer to provide a ``partial policy''
that constrains the set of permitted actions at each point, but does
not specify a complete policy for each subtask.  The third approach is
to define each subtask in terms of a termination predicate and a local
reward function.  These define what it means for the subtask to be
completed and what the final reward should be for completing the
subtask.  The MAXQ method described in this paper follows this
approach, building upon previous work by Singh
\citeyear{s-tlcsest-92}, Kaelbling \citeyear{k-hrl:pr-93}, Dayan and
Hinton \citeyear{dh-frl-93}, and Dean and Lin \citeyear{dl-dtpsd-95}.

An advantage of the ``option'' and partial policy approaches is that
the subtask can be defined in terms of an amount of effort or a course
of action rather than in terms of achieving a particular goal
condition.  However, the ``option'' approach (at least in the simple
form described here), requires the programmer to provide complete
policies for the subtasks, which can be a difficult programming task
in real-world problems.  On the other hand, the termination predicate
method requires the programmer to guess the relative desirability of
the different states in which the subtask might terminate.  This can
also be difficult, although Dean and Lin show how these guesses can be
revised automatically by the learning algorithm.

A potential drawback of all hierarchical methods is that the learned
policy may be suboptimal.  The programmer-provided hierarchy
constrains the set of possible policies that can be considered.  If
these constraints are poorly chosen, the resulting policy will be
suboptimal.  Nonetheless, the learning algorithms that have been
developed for the ``option'' and partial policy approaches guarantee
that the learned policy will be the best possible policy consistent
with these constraints.

The termination predicate method suffers from an additional source of
suboptimality.  The learning algorithm described in this paper
converges to a form of local optimality that we call {\it recursive
optimality}.  This means that the policy of each subtask is locally
optimal given the policies of its children.  But there might exist
better hierarchical policies where the policy for a subtask must be
locally suboptimal so that the overall policy is optimal.  This
problem can be avoided by careful definition of termination predicates
and local reward functions, but this is an added burden on the
programmer.  (It is interesting to note that this problem of recursive
optimality has not been noticed previously.  This is because previous
work focused on subtasks with a single terminal state, and in such
cases, the problem does not arise.)

The second design issue is whether to employ state abstractions within
subtasks.  A subtask employs state abstraction if it ignores some
aspects of the state of the environment.  For example, in many robot
navigation problems, choices about what route to take to reach a goal
location are independent of what the robot is currently carrying.
With few exceptions, state abstraction has not been explored
previously.  We will see that the MAXQ method creates many
opportunities to exploit state abstraction, and that these
abstractions can have a huge impact in accelerating learning.  We will
also see that there is an important design tradeoff:  the successful
use of state abstraction requires that subtasks be defined in terms of
termination predicates rather than using the option or partial policy
methods.  This is why the MAXQ method must employ termination
predicates, despite the problems that this can create. 

The third design issue concerns the non-hierarchical ``execution'' of
a learned hierarchical policy.  Kaelbling \citeyear{k-hrl:pr-93} was
the first to point out that a value function learned from a
hierarchical policy could be evaluated incrementally to yield a
potentially much better non-hierarchical policy.  Dietterich
\citeyear{d-mmhrl-98} and Sutton, Singh, Precup, and Ravindran
\citeyear{sspr-istaa-99} generalized this to show how arbitrary
subroutines could be executed non-hierarchically to yield improved
policies.  However, in order to support this non-hierarchical
execution, extra learning is required.  Ordinarily, in hierarchical
reinforcement learning, the only states where learning is required at
the higher levels of the hierarchy are states where one or more of the
subroutines could terminate (plus all possible initial states).  But
to support non-hierarchical execution, learning is required in all
states (and at all levels of the hierarchy).  In general, this
requires additional exploration as well as additional computation and
memory.  As a consequence of the hierarchical decomposition of the
value function, the MAXQ method is able to support either form of
execution, and we will see that there are many problems where the
improvement from non-hierarchical execution is worth the added cost.

The fourth and final issue is what form of learning algorithm to
employ.  An important advantage of reinforcement learning algorithms
is that they typically operate online.  However, finding online
algorithms that work for general hierarchical reinforcement learning
has been difficult, particularly within the termination predicate
family of methods.  Singh's method relied on each subtask having a
unique terminal state; Kaelbling employed a mix of online and batch
algorithms to train her hierarchy; and work within the ``options''
framework usually assumes that the policies for the subproblems are
given and do not need to be learned at all.  The best previous online
algorithms are the HAMQ Q learning algorithm of Parr and Russell (for
the partial policy method) and the Feudal Q algorithm of Dayan and
Hinton.  Unfortunately, the HAMQ method requires ``flattening'' the
hierarchy, and this has several undesirable consequences.  The Feudal
Q algorithm is tailored to a specific kind of problem, and it does not
converge to any well-defined optimal policy.

In this paper, we present a general algorithm, called MAXQ-Q, for
fully-online learning of a hierarchical value function.  We show
experimentally and theoretically that the algorithm converges to a
recursively optimal policy.  We also show that it is substantially
faster than ``flat'' (i.e., non-hierarchical) Q learning when state
abstractions are employed.  Without state abstractions, it gives
performance similar to (or even worse than) the HAMQ algorithm.

The remainder of this paper is organized as follows.  After
introducing our notation in Section 2, we define the MAXQ value
function decomposition in Section 3 and illustrate it with a simple
example Markov decision problem.  Section 4 presents an analytically
tractable version of the MAXQ-Q learning algorithm called the MAXQ-0
algorithm and proves its convergence to a recursively optimal policy.
It then shows how to extend MAXQ-0 to produce the MAXQ-Q algorithm,
and shows how to extend the theorem similarly.  Section 5 takes up the
issue of state abstraction and formalizes a series of five conditions
under which state abstractions can be safely incorporated into the
MAXQ representation.  State abstraction can give rise to a
hierarchical credit assignment problem, and the paper briefly
discusses one solution to this problem.  Finally, Section 7 presents
experiments with three example domains.  These experiments give some
idea of the generality of the MAXQ representation.  They also provide
results on the relative importance of temporal and state abstractions
and on the importance of non-hierarchical execution.  The paper
concludes with further discussion of the design issues that were
briefly described above, and in particular, it tackles the question of
the tradeoff between the method of defining subtasks (via termination
predicates) and the ability to exploit state abstractions. 

Some readers may be disappointed that MAXQ provides no way of learning
the structure of the hierarchy.  Our philosophy in developing MAXQ
(which we share with other reinforcement learning researchers, notably
Parr and Russell) has been to draw inspiration from the development of
Belief Networks \cite{PRNN:Pearl:1988}.  Belief networks were first
introduced as a formalism in which the knowledge engineer would
describe the structure of the networks and domain experts would
provide the necessary probability estimates.  Subsequently, methods
were developed for learning the probability values directly from
observational data.  Most recently, several methods have been
developed for learning the structure of the belief networks from data,
so that the dependence on the knowledge engineer is reduced.

In this paper, we will likewise require that the programmer provide
the structure of the hierarchy.  The programmer will also need to make
several important design decisions.  We will see below that a MAXQ
representation is very much like a computer program, and we will rely
on the programmer to design each of the modules and indicate the
permissible ways in which the modules can invoke each other.  Our
learning algorithms will fill in ``implementations'' of each module in
such a way that the overall program will work well.  We believe that
this approach will provide a practical tool for solving large
real-world MDPs.  We also believe that it will help us understand the
structure of hierarchical learning algorithms.  It is our hope that
subsequent research will be able to automate most of the work that we
are currently requiring the programmer to do.

\section{Formal Definitions}

\subsection{Markov Decision Problems and Semi-Markov Decision Problems}

We employ the standard definitions for Markov Decision Problems and
Semi-Markov Decision Problems. 

In this paper, we restrict our attention to situations in which an
agent is interacting with a fully-observable stochastic environment.
This situation can be modeled as a Markov Decision Problem (MDP)
\tuple{S,A,P,R,P_0} defined as follows:
\begin{itemize}
\item $S$: this is the set of states of the environment.  At each point in
time, the agent can observe the complete state of the environment.
\item $A$: this is a finite set of actions.  Technically, the set of
available actions depends on the current state $s$, but we will
suppress this dependence in our notation.
\item $P$: When an action $a \in A$ is performed, the environment makes a
probabilistic transition from its current state $s$ to a resulting
state $s'$ according to the probability distribution $P(s'|s,a)$.
\item $R$: Similarly, when action $a$ is performed and the environment
makes its transition from $s$ to $s'$, the agent receives a
real-valued (possibly stochastic) reward $R(s'|s,a)$.  To simplify the
notation, it is customary to treat this reward as being given at the
time that action $a$ is initiated, even though it may in general
depend on $s'$ as well as on $s$ and $a$.
\item $P_0$: This is the starting state distribution. When the MDP is
initialized, it is in state $s$ with probability $P_0(s)$.
\end{itemize}
A {\it policy}, $\pi$, is a mapping from states to actions that tells
what action $a = \pi(s)$ to perform when the environment is in state
$s$.

We will consider two settings:  Episodic and Infinite-Horizon.

In the episodic setting, all rewards are finite and there is at least
one zero-cost absorbing terminal state.  An absorbing terminal state
is a state in which all actions lead back to the same state with
probability 1 and zero reward.  We will only consider problems where
all deterministic policies are ``proper''---that is, all deterministic
policies have a non-zero probability of reaching a terminal state when
started in an arbitrary state.  In this setting, the goal of the agent
is to find a policy that maximizes the expected cumulative reward.  In
the special case where all rewards are non-positive, these problems
are referred to as stochastic shortest path problems, because the
rewards can be viewed as costs (i.e., lengths), and the policy
attempts to move the agent along the path of minimum expected cost.

In the infinite horizon setting, all rewards are also finite.  In
addition, there is a discount factor $\gamma$, and the agent's goal is
to find a policy that minimizes the infinite discounted sum of future
rewards.

The value function $V^{\pi}$ for policy $\pi$ is a function  that
tells, for each state $s$, what the expected cumulative reward will be
of executing that policy.  Let $r_t$ be a random variable that tells
the reward that the agent receives at time step $t$ while following
policy $\pi$.  We can define the value function in the episodic
setting as
\[V^{\pi}(s) = E\left\{r_{t} + r_{t+1} + r_{t+2} +  \cdots | s_t = t,
\pi\right\}.\] 
In the discounted setting, the value function is
\[V^{\pi}(s) = E\left\{\left.r_t + \gamma r_{t+1} + \gamma^2 r_{t+2} + 
\cdots \right| s_t = t, \pi\right\}.\] 
We can see that this equation reduces to the previous one when $\gamma
= 1$.  However, in the infinite horizon case, this infinite sum will not
converge unless $\gamma < 1$. 

The value function satisfies the Bellman equation for a fixed policy:
\[
V^{\pi}(s) = \sum_{s'} P(s'|s,\pi(s))  \left[R(s'|s,\pi(s)) + \gamma
V^{\pi}(s')\right].
\]
The quantity on the right-hand side is called the {\it backed-up
value} of performing action $a$ in state $s$.  For each possible
successor state $s'$, it computes the reward that would be received
and the value of the resulting state and then weights those according
to the probability of ending up in $s'$. 

The optimal value function $V^*$ is the value function that
simultaneously maximizes the expected cumulative reward in all states
$s \in S$.  Bellman \citeyear{b-dp-57} proved that it is the unique solution to
what is now known as the Bellman equation:
\begin{equation}
\label{eq-bellman}
V^*(s) = \max_a \;\sum_{s'} P(s'|s,a)  \left[R(s'|s,a) +
\gamma V^{*}(s')\right]. 
\end{equation}
There may be many optimal policies that achieve this value.  Any
policy that chooses $a$ in $s$ to achieve the maximum on the
right-hand side of this equation is an optimal policy.  We will denote
an optimal policy by $\pi^*$.  Note that all optimal policies are
``greedy'' with respect to the backed-up value of the available
actions. 

Closely related to the value function is the so-called {\it
action-value function}, or $Q$ function \cite{w-ldr-89}.  This
function, $Q^{\pi}(s,a)$, gives the expected cumulative reward of
performing action $a$ in state $s$ and then following policy $\pi$
thereafter.  The $Q$ function also satisfies a Bellman equation:
\[Q^{\pi}(s,a) = \sum_{s'} P(s'|s,a)  \left[R(s'|s,a) +
\gamma Q^{\pi}(s',\pi(s'))\right].\] 
The optimal action-value function is written $Q^*(s,a)$, and it
satisfies the equation
\begin{equation}
\label{eq-q-bellman}
Q^{*}(s,a) = \sum_{s'} P(s'|s,a)  \left[R(s'|s,a) +
\gamma \max_{a'} Q^{*}(s',a')\right].
\end{equation}
Note that any policy that is greedy with respect to $Q^*$ is an
optimal policy.  There may be many such optimal policies---they differ
only in how they break ties between actions with identical $Q^*$
values.

An {\it action order}, denoted $\omega$, is a total order over the
actions within an MDP.  That is, $\omega$ is an anti-symmetric,
transitive relation such that $\omega(a_1,a_2)$ is true iff $a_1$ is
preferred to $a_2$.  An {\it ordered greedy policy}, $\pi_{\omega}$ is
a greedy policy that breaks ties using $\omega$.  For example, suppose
that the two best actions at state $s$ are $a_1$ and $a_2$, that
$Q(s,a_1) = Q(s,a_2)$, and that $\omega(a_1,a_2)$.  Then the ordered
greedy policy $\pi_{\omega}$ will choose $a_1$: $\pi_{\omega}(s) =
a_1$.  Note that although there may be many optimal policies for a
given MDP, the ordered greedy policy, $\pi^*_{\omega}$, is unique.

A discrete-time {\it semi-Markov Decision Process} (SMDP) is a
generalization of the Markov Decision Process in which the actions can
take a variable amount of time to complete.  In particular, let the
random variable $N$ denote the number of time steps that action $a$
takes when it is executed in state $s$.  We can extend the state
transition probability function to be the joint distribution of the
result states $s'$ and the number of time steps $N$ when action $a$ is
performed in state $s$: $P(s',N|s,a)$.  Similarly, the reward function
can be changed to be $R(s',N|s,a)$.\footnote{This formalization is
slightly different than the standard formulation of SMDPs, which
separates $P(s'|s,a)$ and $F(t|s,a)$, where $F$ is the cumulative
distribution function for the probability that $a$ will terminate in
$t$ time units, where $t$ is real-valued rather than integer-valued.
In our case, it is important to consider the joint distribution of
$s'$ and $N$, but we do not need to consider actions with arbitrary
real-valued durations.}

It is straightforward to modify the Bellman equation to define the
value function for a fixed policy $\pi$ as
\[
V^{\pi}(s) = \sum_{s',N} P(s',N|s,\pi(s))  \left[R(s',N|s,\pi(s))
+ \gamma^N V^{\pi}(s')\right].
\]
The only change is that the expected value on the right-hand side is
taken with respect to both $s'$ and $N$, and $\gamma$ is raised to the
power $N$ to reflect the variable amount of time that may elapse while
executing action $a$. 

Note that because expectation is a linear operator, we can write each
of these Bellman equations as the sum of the expected reward for
performing action $a$ and the expected value of the resulting state
$s$.  For example, we can rewrite the equation above as
\begin{equation}
\label{eq-expected-smdp-bellman}
V^{\pi}(s) =  \overline{R}(s,\pi(s)) + \sum_{s',N} P(s',N|s,\pi(s))
\gamma^N V^{\pi}(s').
\end{equation}
where $\overline{R}(s,\pi(s))$ is the expected reward of performing
action $\pi(s)$ in state $s$, where the expectation is taken with
respect to $s'$ and $N$.

Note that for the episodic case, there is no difference between a MDP
and a Semi-Markov Decision Process. 

\subsection{Reinforcement Learning Algorithms}

A reinforcement learning algorithm is an algorithm that is given
access to an unknown MDP via the following reinforcement learning
protocol.  At each time step $t$, the algorithm is told the current
state $s$ of the MDP and the set of actions $A(s) \subseteq A$ that
are executable in that state.  The algorithm chooses an action $a \in
A(s)$, and the MDP executes this action (which causes it to move to
state s') and returns a real-valued reward $r$.  If $s$ is an
absorbing terminal state, the set of actions $A(s)$ contains only the
special action {\sf reset}, which causes the MDP to move to one of its
initial states, drawn according to $P_0$.

The learning algorithm is evaluated based on its observed cumulative
reward.  The cumulative reward of a good learning algorithm should
converge to the cumulative reward of the optimal policy for the MDP. 

In this paper, we will make use of two well-known learning algorithms:
Q learning \cite{w-ldr-89,mach:Watkins+Dayan:1992} and SARSA(0)
\cite{rn-oqucs-94}.  Both of these algorithms maintain a tabular
representation of the action-value function $Q(s,a)$.  Every entry of
the table is initialized arbitrarily.

In Q learning, after the algorithm has observed $s$, chosen $a$,
received $r$, and observed $s'$, it performs the following update:
\[
Q_t(s,a) := (1-\alpha_t) Q_{t-1}(s,a) + \alpha_t [r + \gamma \max_{a'} Q_{t-1}(s',a')],
\]
where $\alpha_t$ is a learning rate parameter. 

Jaakkola, Jordan and Singh
\citeyear{nc:Jaakkola+Jordan+Singh:1994} and Bertsekas and
Tsitsiklis \citeyear{bt-ndp-96} prove that if the agent follows an
``exploration policy'' that tries every action in every state
infinitely often and if
\begin{equation}
\lim_{T\rightarrow \infty} \sum_{t=1}^T \alpha_t = \infty
 \;\;\;\; \rm{and} \;\;\;\;
\lim_{T\rightarrow \infty} \sum_{t=1}^T \alpha_t^2 < \infty
\label{eq-cool}
\end{equation}
then $Q_t$ converges to the optimal action-value function $Q^*$ with
probability 1.  Their proof holds in both settings discussed in this
paper (episodic and infinite-horizon). 

The SARSA(0) algorithm is very similar.  After observing $s$, choosing
$a$, observing $r$, observing $s'$, and choosing $a'$, the algorithm
performs the following update:
\[
Q_t(s,a) := (1-\alpha_t) Q_{t-1}(s,a) + \alpha_t(s,a) [r + \gamma Q_{t-1}(s',a')],
\]
where $\alpha_t$ is a learning rate parameter.  The key difference is
that the Q value of the chosen action $a'$, $Q(s',a')$, appears on the
right-hand side in the place where $Q$ learning uses the $Q$ value of
the best action.  Singh, Jaakkola, Littman, and Szepesv\'{a}ri
\citeyear{sjls-crssoprla-98} provide two important convergence
results: First, if a fixed policy $\pi$ is employed to choose actions,
SARSA(0) will converge to the value function of that policy provided
$\alpha_t$ decreases according to Equation (\ref{eq-cool}).  Second,
if a so-called GLIE policy is employed to choose actions, SARSA(0)
will converge to the value function of the optimal policy, provided
again that $\alpha_t$ decreases according to Equation (\ref{eq-cool}).
A GLIE policy is defined as follows:

\begin{definition}  A GLIE (greedy in the limit with infinite
exploration) policy is any policy satisfying
\begin{enumerate}
\item Each action is executed infinitely often in every state that is
visited infinitely often.
\item In the limit, the policy is greedy with respect to the $Q$-value
function with probability 1.
\end{enumerate}
\end{definition}

\section{The MAXQ Value Function Decomposition} \label{sec-decomp}

At the center of the MAXQ method for hierarchical reinforcement
learning is the MAXQ value function decomposition.  MAXQ describes how
to decompose the overall value function for a policy into a collection
of value functions for individual subtasks (and subsubtasks,
recursively).

\subsection{A Motivating Example}
\label{sec-taxi}

\begin{figure}
\centerps{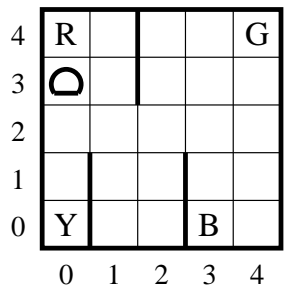}
\caption{The Taxi Domain}
\label{fig-taxi}
\end{figure}

To make the discussion concrete, let us consider the following simple
example.  Figure~\ref{fig-taxi} shows a 5-by-5 grid world inhabited by
a taxi agent.  There are four specially-designated locations in this
world, marked as R(ed), B(lue), G(reen), and Y(ellow).  The taxi
problem is episodic.  In each episode, the taxi starts in a
randomly-chosen square.  There is a passenger at one of the four
locations (chosen randomly), and that passenger wishes to be
transported to one of the four locations (also chosen randomly).  The
taxi must go to the passenger's location (the ``source''), pick up the
passenger, go to the destination location (the ``destination''), and
put down the passenger there.  (To keep things uniform, the taxi must
pick up and drop off the passenger even if he/she is already located
at the destination!)  The episode ends when the passenger is deposited
at the destination location.

There are six primitive actions in this domain: (a) four navigation
actions that move the taxi one square {\sf North}, {\sf South}, {\sf
East}, or {\sf West}, (b) a {\sf Pickup} action, and (c) a {\sf
Putdown} action.  Each action is deterministic.  There is a reward of
$-1$ for each action and an additional reward of $+20$ for
successfully delivering the passenger.  There is a reward of $-10$ if
the taxi attempts to execute the {\sf Putdown} or {\sf Pickup} actions
illegally.  If a navigation action would cause the taxi to hit a wall,
the action is a no-op, and there is only the usual reward of $-1$.

We seek a policy that maximizes the total reward per episode.  There
are 500 possible states: 25 squares, 5 locations for the passenger
(counting the four starting locations and the taxi), and 4
destinations.

This task has a simple hierarchical structure in which there are two
main sub-tasks: Get the passenger and Deliver the passenger.  Each of
these subtasks in turn involves the subtask of navigating to one of
the four locations and then performing a {\sf Pickup} or {\sf Putdown}
action.

This task illustrates the need to support temporal abstraction, state
abstraction, and subtask sharing.  The temporal abstraction is
obvious---for example, the process of navigating to the passenger's
location and picking up the passenger is a temporally extended action
that can take different numbers of steps to complete depending on the
distance to the target.  The top level policy (get passenger; deliver
passenger) can be expressed very simply if these temporal abstractions
can be employed.  

The need for state abstraction is perhaps less obvious.  Consider the
subtask of getting the passenger.  While this subtask is being solved,
the destination of the passenger is completely irrelevant---it cannot
affect any of the nagivation or pickup decisions.  Perhaps more
importantly, when navigating to a target location (either the source
or destination location of the passenger), only the identity of the
target location is important.  The fact that in some cases the taxi is
carrying the passenger and in other cases it is not is irrelevant.

Finally, support for subtask sharing is critical.  If the system could
learn how to solve the navigation subtask once, then the solution
could be shared by both of the ``Get the passenger'' and ``Deliver the
passenger'' subtasks.  We will show below that the MAXQ method
provides a value function representation and learning algorithm that
supports temporal abstraction, state abstraction, and subtask sharing.

To construct a MAXQ decomposition for the taxi problem, we must
identify a set of individual subtasks that we believe will be
important for solving the overall task.  In this case, let us define
the following four tasks:
\begin{itemize}
\item {\sf Navigate}$(t)$.  In this subtask, the goal is to move the
taxi from its current location to one of the four target locations,
which will be indicated by the formal parameter $t$.  

\item {\sf Get}.  In this subtask, the goal is to move the taxi from
its current location to the passenger's current location and pick up
the passenger.  

\item {\sf Put}.  The goal of this subtask is to move the taxi from
the current location to the passenger's destination location and drop
off the passenger.  

\item {\sf Root}.  This is the whole taxi task. 
\end{itemize}

Each of these subtasks is defined by a subgoal, and each subtask
terminates when the subgoal is achieved. 

After defining these subtasks, we must indicate for each subtask which
other subtasks or primitive actions it should employ to reach its
goal.  For example, the {\sf Navigate}$(t)$ subtask should use the
four primitive actions {\sf North}, {\sf South}, {\sf East}, and {\sf
West}.   The {\sf Get} subtask should use the {\sf Navigate} subtask
and the {\sf Pickup} primitive action, and so on. 

\begin{figure}
{\epsfxsize=430pt
\centerps{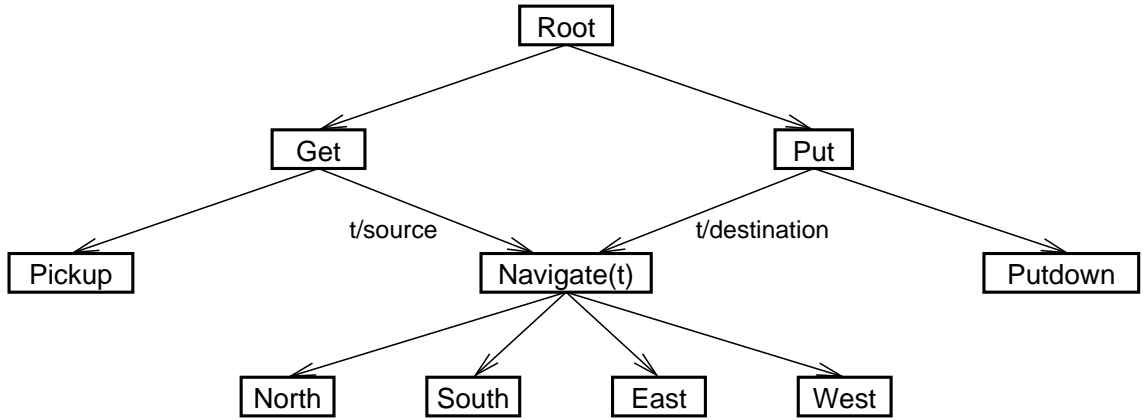}
}
\caption{A task graph for the Taxi problem.}
\label{fig-no-fuel-hierarchy}
\end{figure}

All of this information can be summarized by a directed acyclic graph
called the {\it task graph}, which is shown in
Figure~\ref{fig-no-fuel-hierarchy}.  In this graph, each node
corresponds to a subtask or a primitive action, and each edge
corresponds to a potential way in which one subtask can ``call'' one
of its child tasks.  The notation $formal/actual$ (e.g., $t/source$)
tells how a formal parameter is to be bound to an actual parameter.

Now suppose that for each of these subtasks, we write a policy (e.g.,
as a computer program) to achieve the subtask.  We will refer to the
policy for a subtask as a ``subroutine'', and we can view the parent
subroutine as invoking the child subroutine via ordinary
subroutine-call-and-return semantics.  If we have a policy for each
subtask, then this gives us an overall policy for the Taxi MDP.  The
{\sf Root} subtask executes its policy by calling subroutines that are
policies for the {\sf Get} and {\sf Put} subtasks.  The {\sf Get}
policy calls subroutines for the {\sf Pickup} primitive action and the
{\sf Navigate}$(t)$ subtask.  And so on.  We will call this collection
of policies a {\it hierarchical policy}.  In a hierarchical policy,
each subroutine executes until it enters a terminal state for its
subtask.

\subsection{Definitions}

Let us formalize the discussion so far.  

The MAXQ decomposition takes a given MDP $M$ and decomposes it into a
set of subtasks $\{M_0, M_1, \ldots, M_n\}$ with the convention that
$M_0$ is the root subtask (i.e., solving $M_0$ solves the entire
original MDP $M$).

\begin{definition} An {\em unparameterized subtask} is a three-tuple,
\tuple{T_i, A_i, \tilde{R}_i}, defined as follows: 
{\em 
\begin{enumerate}
\item $T_i(s_i)$ is a termination predicate that partitions $S$ into a
set of {\it active states}, $S_i$ and a set of {\it terminal
states}, $T_i.$ The policy for subtask $M_i$ can only be executed
if the current state $s$ is in $S_i$.

\item $A_i$ is a set of actions that can be performed to achieve
subtask $M_i$.  These actions can either be primitive actions from
$A$, the set of primitive actions for the MDP, or they can be other
subtasks, which we will denote by their indexes $i$.  We will refer to
these actions as the ``children'' of subtask $i$.  If a child subtask
$M_j$ has formal parameters, then it can occur multiple times in
$A_i$, and each such occurrence must specify the actual values that
will be bound to the formal parameters.  The set of actions $A_i$ may
differ from one state to another, so technically, $A_i$ is a
function of $s$.  However, we will suppress this dependence in
our notation.

\item $\tilde{R}_i(s'|s,a)$ is the {\it pseudo-reward function}, which
specifies a {\it pseudo-reward} for each transition from a state $s
\in S_i$ to a terminal state $s' \in T_i$.  This pseudo-reward tells
how desirable each of the terminal states is for this subtask.  It is
typically employed to give goal terminal states a pseudo-reward of 0
and any non-goal terminal states a negative reward.
\end{enumerate}
}

Each primitive action $a$ from $M$ is a primitive subtask in the MAXQ
decomposition such that $a$ is always executable, it always terminates
immediately after execution, and its pseudo-reward function is
uniformly zero.
\end{definition}

If a subtask has formal parameters, then each possible binding of
actual values to the formal parameters specifies a distinct subtask. 
We can think of the values of the formal parameters as being part of
the ``name'' of the subtask.  In practice, of course, we implement a
parameterized subtask by parameterizing the various components of the
task.  If $b$ specifies the actual parameter values for task $M_i$,
then we can define a parameterized termination predicate
$T_i(s,b)$ and a parameterized pseudo-reward function
$\tilde{R}_i(s'|s,a,b)$.  To simplify notation in the rest of the
paper, we will usually omit these parameter bindings from our
notation. 

\begin{table}
\caption{Pseudo-Code for Execution of a Hierarchical Policy}
\label{tab-h-execution}
\vspace*{.1in}
{\footnotesize
\hrule
\begin{center}
\begin{tabbing}
xxxx\=xx\=xxxx\=\kill
1\>$s_t$ is the state of the world at time $t$\\
2\>$K_t$ is the state of the execution stack at time $t$\\[.1in]
3\>while $top(K_t)$ is not a primitive action\\
4\>\>  Let $(i,f_i) := top(K_t)$, where\\
5\>\>\>   $i$ is the name of the ``current'' subroutine, and\\
6 \>\>\>   $f_i$ gives the parameter bindings for $i$\\
7\>\>  Let $(a,f_a) := \pi_i(s,f_i)$, where\\
8\>\>\>   $a$ is the action and $f_a$ gives the parameter bindings
chosen by policy $\pi_i$\\ 
9\>\>  push $(a,f_a)$ onto the stack $K_t$\\[.1in]
10\>Let $(a,nil) := pop(K_t)$ be the primitive action on the top of the stack.\\
11\>Execute primitive action $a$, and update $s_{t+1}$ to be \\
12\>\>the resulting state of the environment.\\[.1in]
13\>while $top(K_t)$ specifies a terminated subtask do\\
14\>\>  $pop(K_t)$\\[.1in]
15\>$K_{t+1} := K_t$ is the resulting execution stack.
\end{tabbing}
\end{center}
}
\hrule
\end{table}

\begin{definition}
A {\em hierarchical policy}, $\pi$, is a set containing a policy for
each of the subtasks in the problem: $\pi = \{\pi_0, \ldots, \pi_n\}.$
\end{definition}

Each subtask policy $\pi_i$ takes a state and returns the name of
a primitive action to execute or the name of a subroutine (and
bindings for its formal parameters) to invoke.  In the terminology of
Sutton, Precup, and Singh \citeyear{sps-bmsm:lprkmts-98}, a subtask
policy is a deterministic ``option'', and its probability of
terminating in state $s$ (which they denote by $\beta(s)$) is 0 if
$s \in S_i$, and 1 if $s \in T_i$.

In a parameterized task, the policy must be parameterized as well so
that $\pi$ takes a state and the bindings of formal parameters and
returns a chosen action and the bindings (if any) of its formal
parameters. 

Table~\ref{tab-h-execution} gives a pseudo-code description of the
procedure for executing a hierarchical policy.  The hierarchical
policy is executed using a stack discipline, as in ordinary
programming languages.  Let $K_t$ denote the contents of the pushdown
stack at time $t$.  When a subroutine is invoked, its name and actual
parameters are pushed onto the stack.  When a subroutine terminates,
its name and actual parameters are popped off the stack.  It is
sometimes useful to think of the contents of the stack as being an
additional part of the state space for the problem.  Hence, a
hierarchical policy implicitly defines a mapping from the current
state $s_t$ and current stack contents $K_t$ to a primitive action
$a$.  This action is executed, and this yields a resulting state
$s_{t+1}$ and a resulting stack contents $K_{t+1}$.  Because of the
added state information in the stack, the hierarchical policy is
non-Markovian with respect to the original MDP.

Because a hierarchical policy maps from states $s$ and stack contents
$K$ to actions, the value function for a hierarchical policy must in
general also assign values to all combinations of states $s$ and stack
contents $K$.  

\begin{definition} A {\em hierarchical value function}, denoted
$V^{\pi}(\mtuple{s,K})$, gives the expected cumulative reward of
following the hierarchical policy $\pi$ starting in state $s$ with
stack contents $K$.
\end{definition}

In this paper, we will primarily be interested only in the ``top
level'' value of the hierarchical policy---that is, the value when the
stack $K$ is empty: $V^{\pi}(\mtuple{s,nil})$.  This is the value of
executing the hierarchical policy beginning in state $s$ and starting
at the top level of the hierarchy. 

\begin{definition} The {\em projected value function}, denoted
$V^{\pi}(s)$, is the value of executing hierarchical policy $\pi$
starting in state $s$ and starting at the root of the task hierarchy.
\end{definition}

\subsection{Decomposition of the Projected Value Function}

Now that we have defined a hierarchical policy and its projected
value function, we can show how that value function can be decomposed
hierarchically.  The decomposition is based on the following theorem:

\begin{theorem}  \label{theorem-subtask}
Given a task graph over tasks $M_0, \ldots, M_n$ and a hierarchical
policy $\pi$, each subtask $M_i$ defines a semi-Markov decision
process with states $S_i$, actions $A_i$, probability transition
function $P^{\pi}_i(s',N|s,a)$, and {\em expected} reward function
$\overline{R}(s,a) = V^{\pi}(a,s)$, where $V^{\pi}(a,s)$ is the
projected value function for child task $M_a$ in state $s$.  If $a$ is
a primitive action, $V^{\pi}(a,s)$ is defined as the expected
immediate reward of executing $a$ in $s$: $V^{\pi}(a,s) =
\sum_{s'}P(s'|s,a) R(s'|s,a)$.
\end{theorem}
\noindent
{\bf Proof:} Consider all of the subroutines that are descendants of
task $M_i$ in the task graph.  Because all of these subroutines are
executing fixed policies (specified by hierarchical policy $\pi$), the
probability transition function $P^{\pi}_i(s',N|s,a)$ is a well
defined, stationary distribution for each child subroutine $a$.  The
set of states $S_i$ and the set of actions $A_i$ are obvious.  The
interesting part of this theorem is the fact that the expected reward
function $\overline{R}(s,a)$ of the SMDP is the projected value
function of the child task $M_a$.

To see this, let us write out the value of $V^{\pi}(i, s)$:
\begin{equation}
\label{eq-r-string}
V^{\pi}(i,s) = E\{r_t + \gamma r_{t+1} + \gamma^2 r_{t+2} + \cdots | s_t = s, \pi\}
\end{equation}
This sum continues until the subroutine for task $M_i$ enters a
state in $T_i$. 

Now let us suppose that the first action chosen by $\pi_i$ is a
subroutine $a$.  This subroutine is invoked, and it executes for a
number of steps $N$ and terminates in state $s'$ according to
$P^{\pi}_i(s',N|s,a)$.  We can rewrite Equation (\ref{eq-r-string})
as
\begin{equation}
\label{eq-r-decompose}
V^{\pi}(i,s) = E\left\{\left.\sum_{u=0}^{N-1} \gamma^u r_{t+u}\;  +
\;\sum_{u=N}^{\infty} \gamma^u r_{t+u}
\right| s_t = s, \pi\right\}
\end{equation}
The first summation on the right-hand side of Equation
(\ref{eq-r-decompose}) is the discounted sum of rewards for executing
subroutine $a$ starting in state $s$ until it terminates, in other
words, it is $V^{\pi}(a,s)$, the projected value function for the
child task $M_a$.  The second term on the right-hand side of the
equation is the value of $s'$ for the current task $i$,
$V^{\pi}(i,s')$, discounted by $\gamma^N$, where $s'$ is the current
state when subroutine $a$ terminates.  We can write
this in the form of a Bellman equation:
\begin{equation}
\label{eq-child-value-as-reward}
V^{\pi}(i,s) = V^{\pi}(\pi_i(s),s) + \sum_{s',N} P^{\pi}_i(s',N|s,\pi_i(s))
\gamma^N V^{\pi}(i,s')
\end{equation}
This has the same form as Equation~(\ref{eq-expected-smdp-bellman}),
which is the Bellman equation for an SMDP, where the first term is the
expected reward $\overline{R}(s,\pi(s))$. \qed

To obtain a hierarchical decomposition of the projected value
function, let us switch to the action-value (or $Q$) representation.
First, we need to extend the $Q$ notation to handle the task
hierarchy.  Let $Q^{\pi}(i,s,a)$ be the expected cumulative reward for
subtask $M_i$ of performing action $a$ in state $s$ and then following
hierarchical policy $\pi$ until subtask $M_i$ terminates.  With this
notation, we can re-state Equation~(\ref{eq-child-value-as-reward}) as
follows:
\begin{equation} \label{eq-q-child-value-as-reward}
Q^{\pi}(i,s,a) = V^{\pi}(a,s) + \sum_{s',N} P^{\pi}_i(s',N|s,a) \gamma^N Q^{\pi}(i,s',\pi(s')),
\end{equation}
The right-most term in this equation is the expected discounted reward
of {\it completing} task $M_i$ after executing action $a$ in state
$s$.  This term only depends on $i$, $s$, and $a$, because the
summation marginalizes away the dependence on $s'$ and $N$.  Let us 
define $C^{\pi}(i,s,a)$ to be equal to this term:

\begin{definition}  The {\em completion function}, $C^{\pi}(i,s,a)$, is
the expected discounted cumulative reward of {\em completing} subtask
$M_i$ after invoking the subroutine for subtask $M_a$ in state $s$.
The reward is discounted back to the point in time where $a$ begins
execution. 
\begin{equation}
\label{eq-decomp-c}
C^{\pi}(i,s,a) = \sum_{s',N} P^{\pi}_i(s',N|s,a) \gamma^N Q^{\pi}(i,s',\pi(s'))
\end{equation}
\end{definition}

With this definition, we can express the $Q$ function recursively as
\begin{equation}
\label{eq-decomp-q}
Q^{\pi}(i,s,a) = V^{\pi}(a,s) + C^{\pi}(i,s,a).
\end{equation}

Finally, we can re-express the definition for $V^{\pi}(i,s)$ as
\begin{equation}
\label{eq-decomp-v}
V^{\pi}(i,s)  =  \left\{ 
\begin{array}{ll}
Q^{\pi}(i,s,\pi_i(s)) & \mbox{ if $i$ is composite}\\
\sum_{s'} P(s'|s,i) R(s'|s,i) & \mbox{ if $i$ is primitive} 
\end{array} \right. 
\end{equation}

We will refer to equations (\ref{eq-decomp-c}), (\ref{eq-decomp-q}),
and (\ref{eq-decomp-v}) as the {\it decomposition equations} for the
MAXQ hierarchy under a fixed hierarchical policy $\pi$.  These
equations recursively decompose the projected value function for the
root, $V^{\pi}(0,s)$ into the projected value functions for the
individual subtasks, $M_1, \ldots, M_n$ and the individual completion
functions $C^{\pi}(j,s,a)$ for $j=1, \ldots, n$.  The fundamental
quantities that must be stored to represent the value function
decomposition are just the $C$ values for all non-primitive subtasks
and the $V$ values for all primitive actions.

\begin{figure}
{\epsfxsize=4in
\centerps{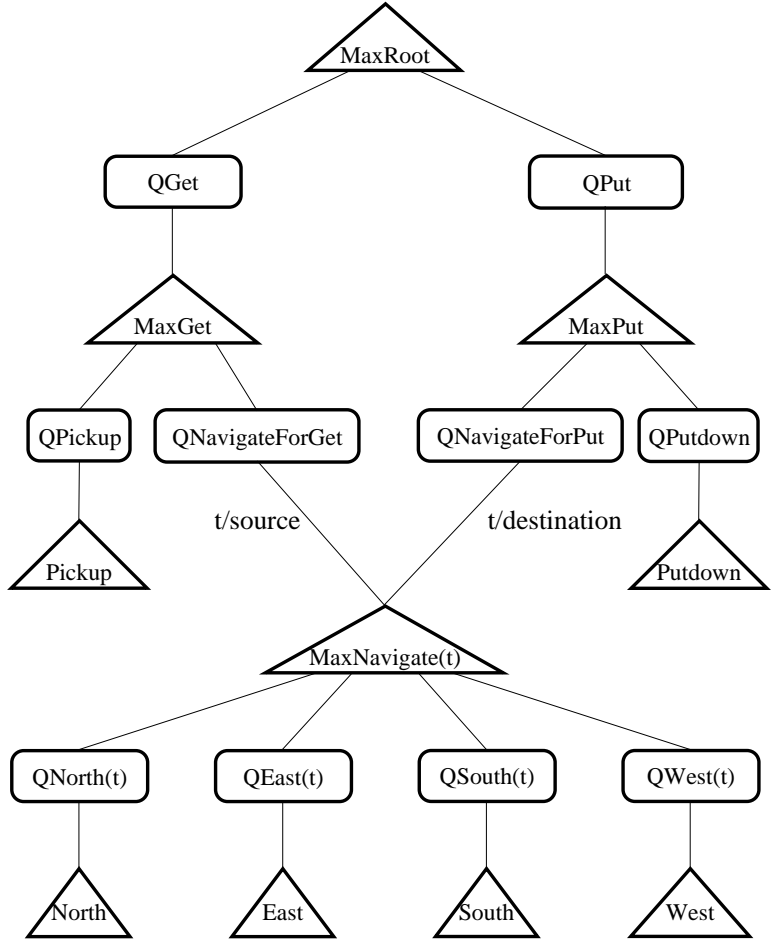}
}
\caption{A MAXQ graph for the Taxi Domain}
\label{fig-taxi-tree}
\end{figure}

To make it easier for programmers to design and debug MAXQ
decompositions, we have developed a graphical representation that we
call the {\it MAXQ graph}.  A MAXQ graph for the Taxi domain is shown
in Figure~\ref{fig-taxi-tree}.  The graph contains two kinds of nodes,
Max nodes and Q nodes.  The Max nodes correspond to the subtasks in
the task decomposition---there is one Max node for each primitive
action and one Max node for each subtask (including the {\sf Root})
task.  Each primitive Max node $i$ stores the value of $V^{\pi}(i,s)$.
The $Q$ nodes correspond to the actions that are available for each
subtask.  Each $Q$ node for parent task $i$, state $s$ and subtask $a$
stores the value of $C^{\pi}(i,s,a)$.

In addition to storing information, the Max nodes and Q nodes can be
viewed as performing parts of the computation described by the
decomposition equations.  Specifically, each Max node $i$ can be
viewed as computing the projected value function $V^{\pi}(i,s)$ for
its subtask.  For primitive Max nodes, this information is stored in
the node.  For composite Max nodes, this information is obtained by
``asking'' the $Q$ node corresponding to $\pi_i(s)$.  Each Q node with
parent task $i$ and child task $a$ can be viewed as computing the
value of $Q^{\pi}(i,s,a)$.  It does this by ``asking'' its child task
$a$ for its projected value function $V^{\pi}(a,s)$ and then adding
its completion function $C^{\pi}(i,s,a)$.

As an example, consider the situation shown in Figure~\ref{fig-taxi},
which we will denote by $s_1$.  Suppose that the passenger is at R and
wishes to go to B.  Let the hierarchical policy we are evaluating be
an optimal policy denoted by $\pi$ (we will omit the superscript * to
reduce the clutter of the notation).  The value of this state under
$\pi$ is 10, because it will cost 1 unit to move the taxi to R, 1 unit
to pickup the passenger, 7 units to move the taxi to B, and 1 unit to
putdown the passenger, for a total of 10 units (a reward of $-10$).
When the passenger is delivered, the agent gets a reward of +20, so
the net value is +10.

\begin{figure}
{\epsfxsize=4in
\centerps{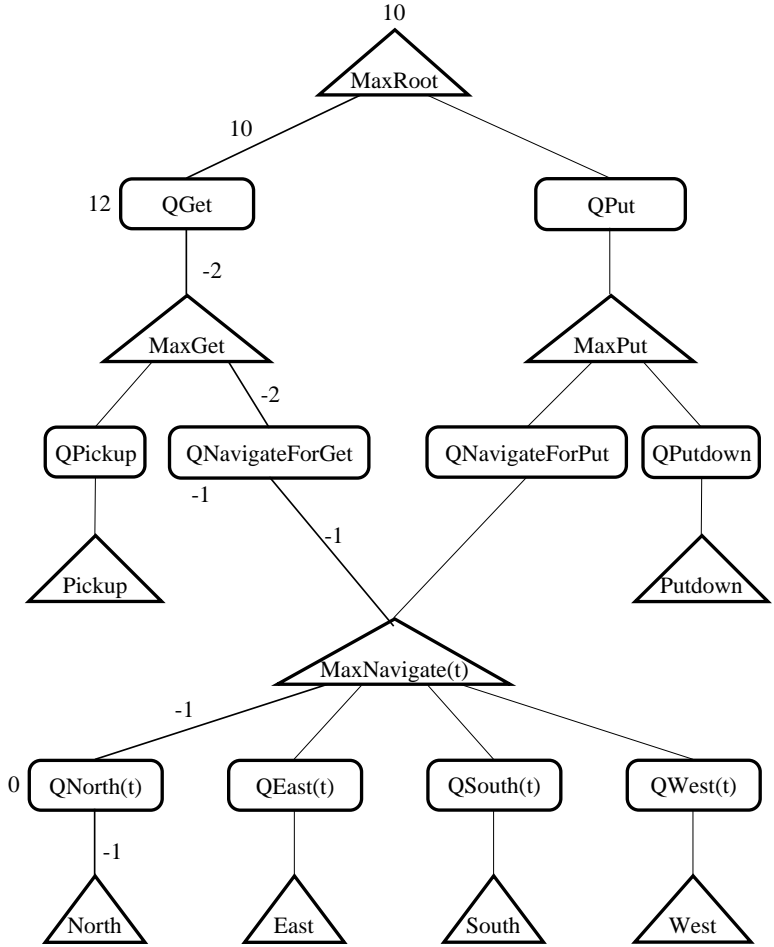}
}
\caption{Computing the value of a state using the MAXQ hierarchy. The
$C$ value of each Q node is shown to the left of the node.  All other
numbers show the values being returned up the graph.  }
\label{fig-decompose}
\end{figure}

Figure~\ref{fig-decompose} shows how the MAXQ hierarchy computes this
value.  To compute the value $V^{\pi}({\sf Root}, s_1)$, {\sf MaxRoot}
consults its policy and finds that $\pi_{\sf Root}(s_1)$ is {\sf Get}.
Hence, it ``asks'' the $Q$ node, {\sf QGet} to compute $Q^{\pi}({\sf
Root}, s_1, {\sf Get})$.  The completion cost for the {\sf Root} task
after performing a {\sf Get}, $C^{\pi}({\sf Root}, s_1, {\sf Get})$,
is 12, because it will cost 8 units to deliver the customer (for a net
reward of $20 - 8 = 12$) after completing the {\sf Get} subtask.
However, this is just the reward {\it after} completing the {\sf Get},
so it must ask {\sf MaxGet} to estimate the expected reward of
performing the {\sf Get} itself.

The policy for {\sf MaxGet} dictates that in $s_1$, the {\sf Navigate}
subroutine should be invoked with $t$ bound to R, so {\sf MaxGet}
consults the Q node, {\sf QNavigateForGet} to compute the expected
reward.  {\sf QNavigateForGet} knows that after completing the {\sf
Navigate}(R) task, one more action (the {\sf Pickup}) will be required
to complete the {\sf Get}, so $C^{\pi}({\sf MaxGet}, s_1, {\sf
Navigate}(R)) = -1$.  It then asks ${\sf MaxNavigate}(R)$ to compute
the expected reward of performing a {\sf Navigate} to location R.  

The policy for ${\sf MaxNavigate}$ chooses the {\sf North} action, so
{\sf MaxNavigate} asks {\sf QNorth} to compute the value.  {\sf
QNorth} looks up its completion cost, and finds that $C^{\pi}({\sf
Navigate}, s_1, {\sf North})$ is 0 (i.e., the {\sf Navigate} task will
be completed after performing the {\sf North} action).  It consults
{\sf MaxNorth} to determine the expected cost of performing the {\sf
North} action itself.  Because {\sf MaxNorth} is a primitive action,
it looks up its expected reward, which is $-1$. 

Now this series of recursive computations can conclude as follows:
\begin{itemize}
\item $Q^{\pi}({\sf Navigate}(R), s_1, {\sf North}) = -1 + 0$
\item $V^{\pi}({\sf Navigate}(R), s_1) = -1$
\item $Q^{\pi}({\sf Get}, s_1, {\sf Navigate}(R)) = -1 + -1$
\\
($-1$ to perform the {\sf Navigate} plus $-1$ to complete the {\sf
Get}. 
\item $V^{\pi}({\sf Get}, s_1) = -2$
\item $Q^{\pi}({\sf Root}, s_1, {\sf Get}) = -2 + 12$
\\
($-2$ to perform the {\sf Get} plus 12 to complete the {\sf Root} task
and collect the final reward). 
\end{itemize}

The end result of all of this is that the value of $V^{\pi}({\sf
Root},s_1)$ is decomposed into a sum of $C$ terms plus the expected
reward of the chosen primitive action:
\begin{eqnarray*}
V^{\pi}({\sf Root},s_1) & = & V^{\pi}({\sf North}, s_1) +
C^{\pi}({\sf Navigate}(R), s_1, {\sf North}) +\\
 & & C^{\pi}({\sf Get}, s_1, {\sf Navigate}(R)) +
C^{\pi}({\sf Root}, s_1, {\sf Get})\\
& = & -1 + 0 + -1 + 12 \\
& = & 10
\end{eqnarray*}

In general, the MAXQ value function decomposition has the form 
\begin{eqnarray}
V^{\pi}(0,s) &=& V^{\pi}(a_m, s) + C^{\pi}({a_{m-1}},s, a_m) + \ldots
+  C^{\pi}({a_1},s,a_2) + C^{\pi}(0, s, a_1),
\label{eq-path}
\end{eqnarray}
where $a_0, a_1, \ldots, a_m$ is the ``path'' of Max nodes chosen by
the hierarchical policy going from the {\sf Root} down to a primitive
leaf node. 

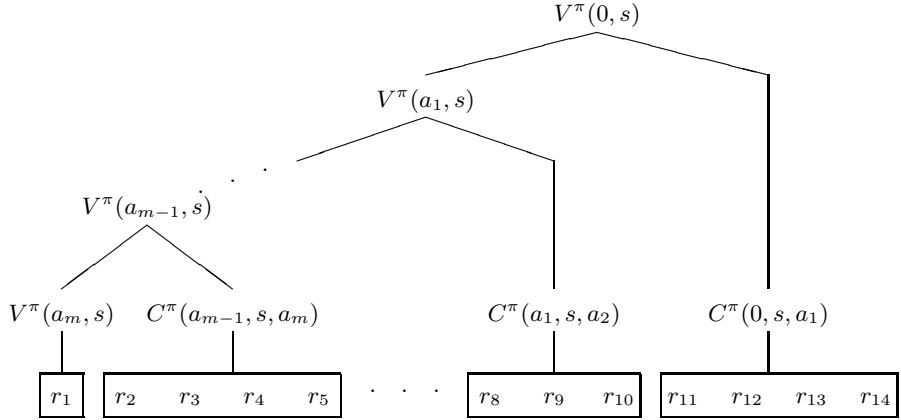
\begin{figure}
\begin{center}
\setlength{\unitlength}{0.00075in}%
\begin{picture}(6116,2918)(2447,-3223)
\footnotesize
\put(5104,-3033){\makebox(0,0)[b]{.}}
\put(4879,-3033){\makebox(0,0)[b]{.}}
\put(5329,-3033){\makebox(0,0)[b]{.}}
\put(2703,-3107){\makebox(0,0)[b]{$r_1$}}
\put(3153,-3107){\makebox(0,0)[b]{$r_2$}}
\put(3603,-3107){\makebox(0,0)[b]{$r_3$}}
\put(4053,-3107){\makebox(0,0)[b]{$r_4$}}
\put(4503,-3107){\makebox(0,0)[b]{$r_5$}}
\put(5703,-3107){\makebox(0,0)[b]{$r_8$}}
\put(6153,-3107){\makebox(0,0)[b]{$r_9$}}
\put(6603,-3107){\makebox(0,0)[b]{$r_{10}$}}
\put(7053,-3107){\makebox(0,0)[b]{$r_{11}$}}
\put(7503,-3107){\makebox(0,0)[b]{$r_{12}$}}
\put(7953,-3107){\makebox(0,0)[b]{$r_{13}$}}
\put(8403,-3107){\makebox(0,0)[b]{$r_{14}$}}
\thinlines
\put(2551,-3211){\framebox(300,300){}}
\put(3001,-3211){\framebox(1650,300){}}
\put(5551,-3211){\framebox(1200,300){}}
\put(6901,-3211){\framebox(1650,300){}}
\put(2701,-2611){\line( 0,-1){300}}
\put(7651,-2611){\line( 0,-1){300}}
\put(3301,-1862){\line(-4,-3){600}}
\put(3301,-1862){\line( 4,-3){600}}
\put(3901,-2611){\line( 0,-1){300}}
\put(6151,-2611){\line( 0,-1){300}}
\put(6151,-1411){\line( 0,-1){900}}
\put(7651,-811){\line( 0,-1){1500}}
\put(6451,-512){\line( 4,-1){1200}}
\put(6451,-512){\line(-4,-1){1200}}
\put(5251,-1112){\line( 3,-1){900}}
\put(5251,-1112){\line(-3,-1){900}}
\put(3901,-1561){\makebox(0,0)[b]{.}}
\put(4126,-1486){\makebox(0,0)[b]{.}}
\put(3676,-1636){\makebox(0,0)[b]{.}}
\put(6451,-437){\makebox(0,0)[b]{$V^{\pi}(0,s)$}}
\put(5251,-1037){\makebox(0,0)[b]{$V^{\pi}({a_1},s)$}}
\put(3301,-1787){\makebox(0,0)[b]{$V^{\pi}({a_{m-1}},s)$}}
\put(2701,-2536){\makebox(0,0)[b]{$V^{\pi}({a_m},s)$}}
\put(3901,-2536){\makebox(0,0)[b]{$C^{\pi}({a_{m-1}},s,a_m)$}}
\put(7651,-2536){\makebox(0,0)[b]{$C^{\pi}(0,s,a_1)$}}
\put(6151,-2536){\makebox(0,0)[b]{$C^{\pi}({a_1},s,a_2)$}}
\end{picture}
\end{center}
\vspace{-.15in}
\caption{The MAXQ decomposition; $r_1, \ldots, r_{14}$ denote the
sequence of rewards received from primitive actions at times $1,
\ldots, 14$.}
\label{fig-path}
\end{figure}

We can summarize the presentation of this section by the following
theorem:

\begin{theorem}  \label{theorem-decomp}
Let $\pi = \{\pi_i; i = 0, \ldots, n \}$ be a hierarchical policy
defined for a given MAXQ graph with subtasks $M_0, \ldots, M_n,$ and let
$i = 0$ be the root node of the graph.  Then there exist values for
$C^{\pi}(i,s,a)$ (for internal Max nodes) and $V^{\pi}(i,s)$ (for
primitive, leaf Max nodes) such that $V^{\pi}(0,s)$ (as computed by
the decomposition equations (\ref{eq-decomp-c}), (\ref{eq-decomp-q}),
and (\ref{eq-decomp-v})) is the expected discounted cumulative reward
of following policy $\pi$ starting in state $s$.
\end{theorem}
\noindent
{\bf Proof:} The proof is by induction on the number of levels in the
task graph.  At each level $i$, we compute values for
$C^{\pi}(i,s,\pi(s))$ (or $V^{\pi}(i,s),$ if $i$ is primitive)
according to the decomposition equations.  We can apply the
decomposition equations again to compute $Q^{\pi}(i,s,\pi(s))$ and
apply Equation (\ref{eq-q-child-value-as-reward}) and Theorem
\ref{theorem-subtask} to conclude that $Q^{\pi}(i,s,\pi(s))$ gives the
value function for level $i$.  When $i=0$, we obtain the value
function for the entire hierarchical policy.  {\bf Q. E. D.}

It is important to note that this representation theorem does not
mention the pseudo-reward function, because the pseudo-reward is used
only during learning.  This theorem captures the {\it representational
power} of the MAXQ decomposition, but it does not address the question
of whether there is a learning algorithm that can find a given policy.
That is the subject of the next section.

\section{A Learning Algorithm for the MAXQ Decomposition}

In order to develop a learning algorithm for the MAXQ decomposition,
we must consider exactly what we are hoping to achieve.  Of course,
for any MDP $M$, we would like to find an optimal policy $\pi^*$.
However, in the MAXQ method (and in hierarchical reinforcement
learning in general), the programmer imposes a hierarchy on the
problem.  This hierarchy constrains the space of possible policies so
that it may not be possible to represent the optimal policy or its
value function.

In the MAXQ method, the constraints take two forms.  First, within a
subtask, only some of the possible primitive actions may be permitted.
For example, in the taxi task, during a ${\sf Navigate}(t)$, only the
{\sf North}, {\sf South}, {\sf East}, and {\sf West} actions are
available---the {\sf Pickup} and {\sf Putdown} actions are not
allowed.  Second, consider a Max node $M_j$ with child nodes
$\{M_{j_1}, \ldots, M_{j_k}\}$.  The policy learned for $M_j$ must
involve executing the learned policies of these child nodes.  When the
policy for child node $M_{j_i}$ is executed, it will run until it
enters a state in $T_{j_i}$.  Hence, any policy learned for $M_j$ must
pass through some subset of these terminal state sets $\{T_{j_1},
\ldots, T_{j_k}\}$. 

The HAM method shares these same two constraints and in addition, it
imposes a partial policy on each node, so that the policy for any
subtask $M_i$ must be a deterministic refinement of the given
non-deterministic initial policy for node $i$.  

In the ``option'' approach, the policy is even further constrained.
In this approach, there are only two non-primitive levels in the
hierarchy, and the subtasks at the lower level are given complete
policies by the programmer.  Hence, any learned policy must be
constructed by ``concatenating'' the given lower level policies in
some order. 

The purpose of imposing these constraints on the policy is to
incorporate prior knowledge and thereby reduce the size of the space
that must be searched to find a good policy.  However, these
constraints may make it impossible to learn the optimal policy. 

If we can't learn the optimal policy, the next best target would be to
learn the best policy that is consistent with (i.e., can be
represented by) the given hierarchy. 

\begin{definition}  A {\em hierarchically optimal policy} for MDP $M$
is a policy that achieves the highest cumulative reward among all
policies consistent with the given hierarchy. 
\end{definition}

Parr \citeyear{p-hclmdp-98} proves that his HAMQ learning algorithm
converges with probability 1 to a hierarchically optimal policy.
Similarly, given a fixed set of options, Sutton, Precup, and Singh
\citeyear{sps-bmsm:lprkmts-98} prove that their SMDP learning
algorithm converges to a hierarchically optimal value function.
(Incidentally, they also show that if the primitive actions are also
made available as ``trivial'' options, then their SMDP method
converges to the optimal policy.  However, in this case, it is hard to
say anything formal about how the options speed the learning process.
They may in fact hinder it \cite{hmbkd-hsmdpm-98}.)

With the MAXQ method, we will seek an even weaker form of optimality:
recursive optimality.

\begin{definition} A {\em recursively optimal policy} for MDP $M$ with
MAXQ decomposition $\{M_0, \ldots, M_k\}$ is a hierarchical policy
$\pi = \{\pi_0, \ldots, \pi_k\}$ such that for each subtask $M_i$, the
corresponding policy $\pi_i$ is optimal for the SMDP defined by the set
of states $S_i$, the set of actions $A_i$, the state transition
probability function $P^{\pi}(s',N|s,a)$, and the reward function given
by the {\em sum} of the original reward function $R(s'|s,a)$ and the
pseudo-reward function $\tilde{R}_i(s')$. 
\end{definition}

Note that in this definition, the state transition probability
distribution is defined by the locally optimal policies $\{\pi_j\}$ of
all subtasks that are descendants of $M_i$ in the MAXQ graph.  Hence,
recursive optimality is a kind of local optimality in which the policy
at each node is optimal given the policies of its children. 

The reason to seek recursive optimality rather than hierarchical
optimality is that recursive optimality makes it possible to solve
each subtask without reference to the context in which it is
executed.  This context-free property makes it easier to share and
re-use subtasks.  It will also turn out to be essential for the
successful use of state abstraction.  

\begin{figure}
\begin{minipage}{1.6in}
{\epsfxsize=1.5in
\centerps{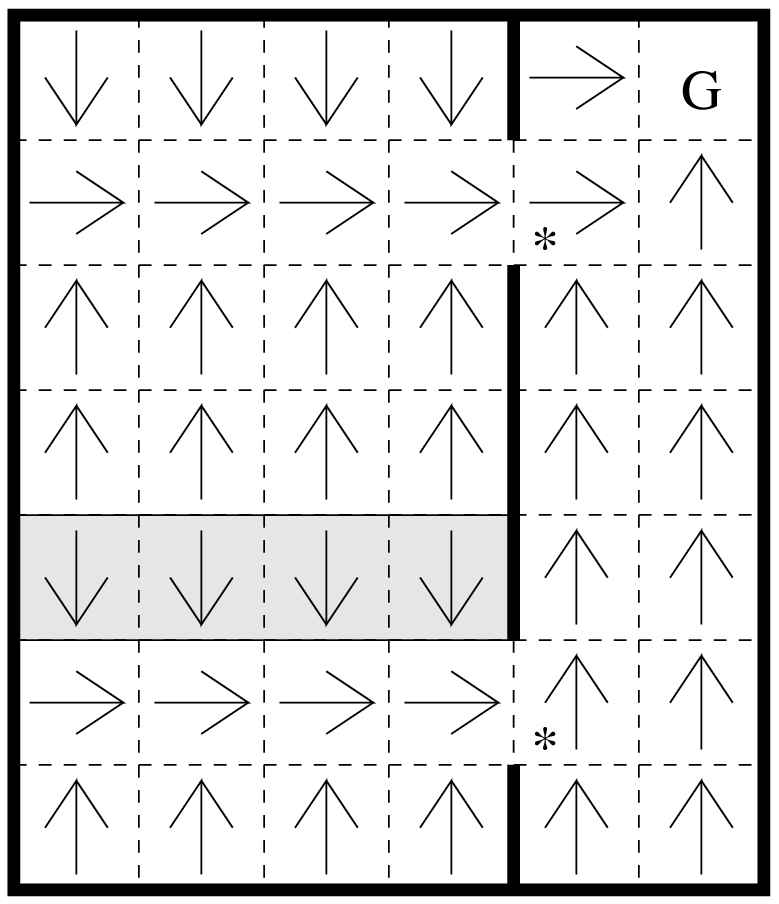}
}
\end{minipage}
\hfil
\begin{minipage}{4.7in}
{\epsfxsize=4.6in
\centerps{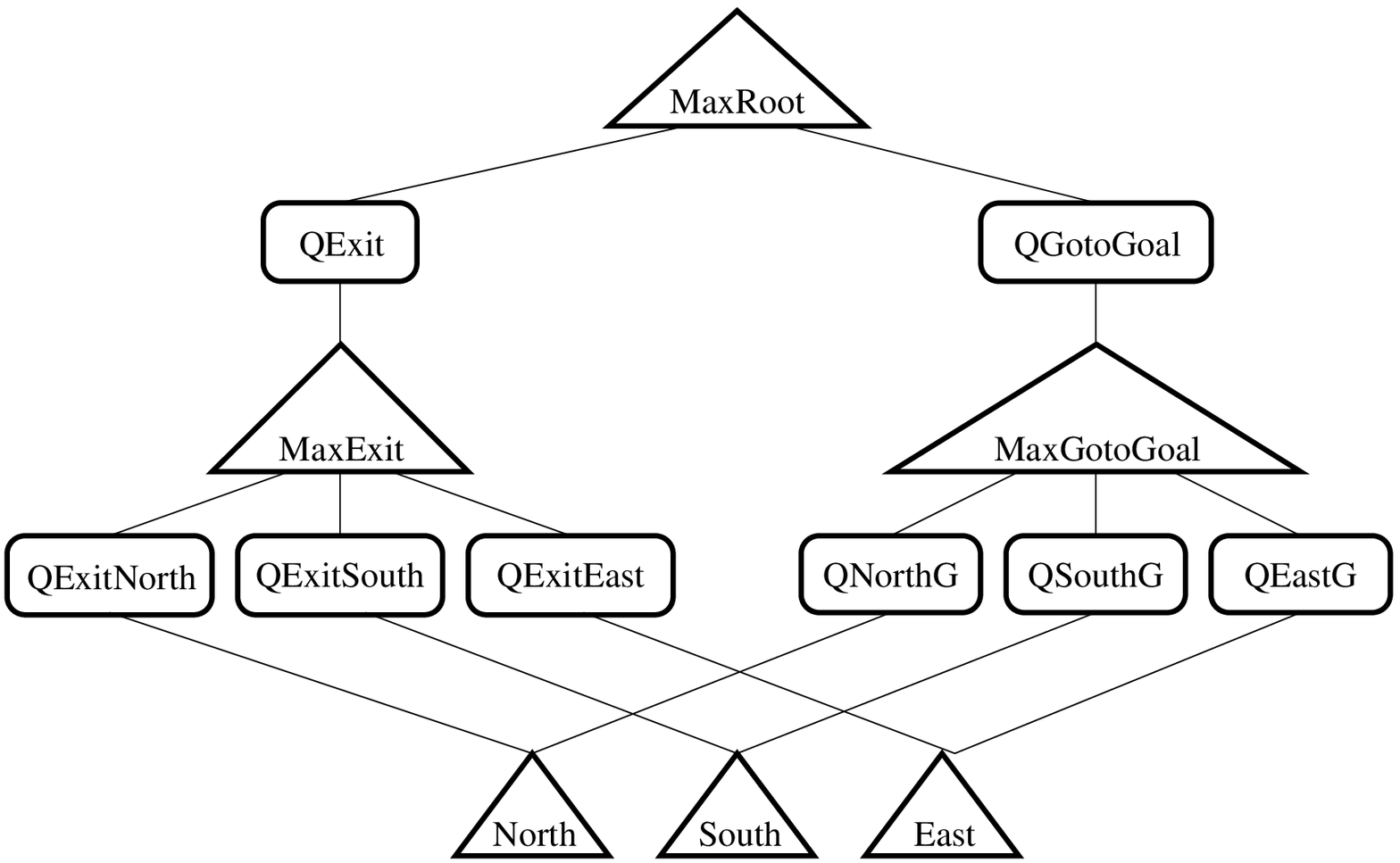}
}
\end{minipage}
\caption{A simple MDP (left) and its associated MAXQ graph (right).
The policy shown in the left diagram is recursively optimal but not
hierarchically optimal.  The shaded cells indicate points where the
locally-optimal policy is not globally optimal.}
\label{fig-two-rooms}
\end{figure}

Before we proceed to describe our learning algorithm for recursive
optimality, let us see how recursive optimality differs from
hierarchical optimality. 

It is easy to construct examples of policies that are recursively
optimal but not hierarchically optimal.  Consider the simple maze
problem and its associated MAXQ graph shown in
Figures~\ref{fig-two-rooms}.  Suppose a robot starts somewhere in the
left room, and it must reach the goal G in the right room.  The robot
has three actions, {\sf North}, {\sf South}, and {\sf East}, and these
actions are deterministic.  The robot receives a reward of $-1$ for
each move.  Let us define two subtasks:
\begin{itemize}
\item {\sf Exit}.  This task terminates when the robot exits the left
room.  We can set the pseudo-reward function $\tilde{R}$ to be 0 for
the two terminal states (i.e., the two states indicated by *'s). 

\item {\sf GotoGoal}.  This task terminates when the robot reaches the
goal G. 
\end{itemize}

The arrows in Figure~\ref{fig-two-rooms} show the locally optimal policy
within each room.  The arrows on the left seek to exit the left room
by the shortest path, because this is what we specified when we set
the pseudo-reward function to 0.  The arrows on the right follow the
shortest path to the goal, which is fine.  However, the resulting
policy is neither hierarchically optimal nor optimal.  

There exists a hierarchical policy that would always exit the left
room by the upper door.  The MAXQ value function decomposition can
represent the value function of this policy, but such a policy would
not be locally optimal (because, for example, the states in the
``shaded'' region would not follow the shortest path to a doorway). If
we consider for a moment, we can see a way to fix this problem.  The
value of the upper starred state under the optimal hierarchical policy
is $-2$ and the value of the lower starred state is $-6$.  Hence, if
we set $\tilde{R}$ to have these values, then the recursively-optimal
policy would be hierarchically optimal (and globally optimal).  In
other words, if the programmer can guess the right values for the
terminal states of a subtask, then the recursively optimal policy will
be hierarchically optimal (provided that all primitive actions are
available within the subtask). 

This basic idea was first pointed out by Dean and Lin
\citeyear{dl-dtpsd-95}.  They describe an algorithm that makes initial
guesses for the values of these starred states and then updates those
guesses based on the computed values of the starred states under the
resulting recursively-optimal policy.  They proved that this will
converge to a hierarchically optimal policy.  The drawback of their
method is that it requires repeated solution of the resulting
hierarchical learning problem, and this does not always yield a
speedup over just solving the original, flat problem.

Parr \citeyear{Parr.UAI98} proposed an interesting approach that
constructs a {\it set} of different $\tilde{R}$ functions and computes
the recursively optimal policy under each of them for each subtask.
His method chooses the $\tilde{R}$ functions in such a way that the
hierarchically optimal policy can be approximated to any desired
degree.  Unfortunately, the method is quite inefficient, because it
relies on solving a series of linear programming problems each of
which requires time polynomial in several parameters, including the
number of states $|S_i|$ within the subtask.

This discussion suggests that while, in principle, it is possible to
learn good values for the pseudo-reward function, in practice, we must
rely on the programmer to specify a single pseudo-reward function,
$\tilde{R}$.  If the programmer wishes to consider a small number of
alternative pseudo-reward functions, they can be handled by defining a
small number of subtasks that are identical except for their
$\tilde{R}$ functions, and permitting the learning algorithm to choose
the one that gives the best recursively-optimal policy.

In practice, we have employed the following simplified approach to
defining $\tilde{R}$.  For each subtask $M_i$, we define two
predicates: the termination predicate, $T_i$, and a goal predicate
$G_i$.  The goal predicate defines a subset of the terminated states
that are ``goal states'', and these have a pseudo-reward of 0.  All
other terminal states have a fixed constant pseudo-reward (e.g.,
$-100$) that is set so that it is always better to terminate in a goal
state than in a non-goal state. For the problems on which we have
tested the MAXQ method, this worked very well.

In our experiments with MAXQ, we have found that it is easy to make
mistakes in defining $T_i$ and $G_i$.  If the goal is not defined
carefully, it is easy to create a set of subtasks that lead to
infinite looping.  For example, consider again the problem in
Figure~\ref{fig-two-rooms}.  Suppose we permit a fourth action, {\sf
West} in the MDP and let us define the termination and goal predicates
for the right hand room to be satisfied iff either the robot reaches
the goal or it exits the room.  This is a very natural definition,
since it is quite similar to the definition for the left-hand room.
However, the resulting locally-optimal policy for this room will
attempt to move to the nearest of these three locations: the goal, the
upper door, or the lower door.  We can easily see that for all but a
few states near the goal, the only policies that can be constructed by
{\sf MaxRoot} will loop forever, first trying to leave the left room
by entering the right room, and then trying to leave the right room by
entering the left room.  This problem is easily fixed by defining the
goal predicate $G_i$ for the right room to be true if and only if the
robot reaches the goal G.  But avoiding such ``undesired termination''
bugs can be hard in more complex domains.

Now that we have an understanding of recursively optimal policies, we
present two learning algorithms.  The first one, called MAXQ-0,
applies only in the case when the pseudo-reward function $\tilde{R}$
is always zero.  We will first prove its convergence properties and
then show how it can be extended to give the second algorithm, MAXQ-Q,
which works with general pseudo-reward functions.

\begin{table}
\caption{The MAXQ-0 learning algorithm.}
\label{tab-maxq-0}
{\footnotesize
\begin{tabbing}
xxxxx\=xxxx\=xxxx\=xxxx\=\kill
1\> {\bf function} MAXQ-0(MaxNode $i$, State $s$)\\[.1in]
2\> {\bf if} $i$ is a primitive MaxNode \\
3\> \>  execute $i$, receive $r$, and observe result state $s'$ \\
4\> \>  $V_{t+1}(i,s) := (1-\alpha_t(i)) \cdot V_t(i,s) + \alpha_t(i)
\cdot r_t$ \\
5\> \>  {\bf return} 1 \\
6\> {\bf else} \\
7\> \>  {\bf let} $count = 0$ \\
8\> \>  {\bf while} $T_i(s)$ is false {\bf do} \\
9\> \>\>    choose an action $a$ according to the current exploration policy $\pi_x(i,s)$ \\
10\> \>\>    {\bf let} N = MAXQ-0(a, s) \\
11\> \>\>    observe result state $s'$ \\
12\> \>\>    $C_{t+1}(i,s,a) := (1-\alpha_t(i)) \cdot C_t(i,s,a) +
\alpha_t(i) \cdot \gamma^N V_t(i,s') $\\
13\> \>\>    $count := count + N$ \\
14\> \>\>    $s := s'$\\
15\> \>\>    {\bf end} \\
16\> \>  {\bf return} $count$ \\
17\> {\bf end} MAXQ-0
\end{tabbing}
} 
\end{table}

Table~\ref{tab-maxq-0} gives pseudo-code for MAXQ-0.  MAXQ-0 is a
recursive function that executes the current exploration policy
starting at Max node $i$ in state $s$.  It performs actions until it
reaches a terminal state, at which point it returns a count of the
total number of primitive actions that have been executed.  To execute
an action, MAXQ-0 calls itself recursively.  When the recursive call
returns, it updates the value of the completion function for node $i$.
It uses the count of the number of primitive actions to appropriately
discount the value of the resulting state $s'$.  At leaf nodes, MAXQ-0
updates the estimated one-step expected reward, $V(i,s)$.  The value
$\alpha_t(i)$ is a ``learning rate'' parameter that should be
gradually decreased to zero in the limit.

There are two things that must be specified in order to make this
algorithm description complete.  First, we must specify how to compute
$V_t(i,s')$ in line 12, since it is not stored in the Max node.  It is
computed by the following modified versions of the decomposition
equations:

\begin{eqnarray}
V_t(i,s)  &=&  \left\{ 
\begin{array}{ll}
\max_a Q_t(i,s,a) & \mbox{ if $i$ is composite}\\
V_t(i,s) & \mbox{ if $i$ is primitive} 
\end{array} \right. \\
Q_t(i,s,a) &=& V_t(a,s) + C_t(i,s,a).
\end{eqnarray}

These equations reflect two important changes compared with Equations
(\ref{eq-decomp-q}) and (\ref{eq-decomp-v}).  First, in the first
equation, $V_t(i,s)$ is defined in terms of the $Q$ value of the {\it
best} action $a$, rather than of the action chosen by a fixed
hierarchical policy.  Second, there are no $\pi$ superscripts, because
the current value function, $V_t(i,s)$ is not based on a fixed
hierarchical policy $\pi$.

To compute $V_t(i,s)$ using these equations, we must perform a complete
search of all paths through the MAXQ graph starting at node $i$ and
ending at the leaf nodes.  Table~\ref{tab-eval-greedy} gives
pseudo-code for a recursive function, EvaluateMaxNode, that implements
a depth-first search.  In addition to returning $V_t(i,s)$,
EvaluateMaxNode also returns the action at the leaf node that achieves
this value.  This information is not needed for MAXQ-0, but it will be
useful later when we consider non-hierarchical execution of the
learned recursively-optimal policy.

\begin{table}
\caption{Pseudo-code for Greedy Execution of the MAXQ Graph}
\label{tab-eval-greedy}
\vspace*{.1in}
\hrule
{\footnotesize
\begin{quote}
\begin{tabbing}
xxx\=xxxx\=xxxx\=xxxx\=\kill
1 \>function EvaluateMaxNode$(i,s)$\\[.1in]
2 \>\>  if $i$ is a primitive Max node\\
3 \>\>\>     return \tuple{V_t(i,s), i}\\
4 \>\> else\\
5 \>\>\>     for each $j \in A_i$, \\
6 \>\>\>\>       let \tuple{V_(j,s), a_j} = EvaluateMaxNode$(j,s)$\\
7 \>\>\>     let $j^{hg} = \argmax_j V_t(j,s) + C_t(i,s,j)$       \\
8 \>\>\>     return \tuple{V_t(j^{hg},s), a_{j^{hg}}}\\
  \> {\bf end} // EvaluateMaxNode
\end{tabbing}
\end{quote}
}
\hrule
\end{table}

The second thing that must be specified to complete our definition of
MAXQ-0 is the exploration policy, $\pi_x$.  We require that $\pi_x$ be
an ordered GLIE policy.  

\begin{definition}
An {\em ordered GLIE policy} is a GLIE policy (Greedy in the Limit of
Infinite Exploration) that converges in the limit to an {\em ordered greedy
policy}, which is a greedy policy that imposes an arbitrary fixed
order $\omega$ on the available actions and breaks ties in favor of the
action $a$ that appears earliest in that order.
\end{definition}

We need this property in order to ensure that MAXQ-0 converges to a
uniquely-defined recursively optimal policy.  A fundamental problem
with recursive optimality is that in general, each Max node $i$ will
have a choice of many different locally optimal policies given the
policies adopted by its descendant nodes.  These different locally
optimal policies will all achieve the same locally optimal value
function, but they can give rise to different probability transition
functions $P(s',N|s,i)$.  The result will be that the Semi-Markov
Decision Problem defined at the next level above node $i$ in the MAXQ
graph will differ depending on which of these various locally optimal
policies is chosen by node $i$.  However, if we establish a fixed
ordering over the Max nodes in the MAXQ graph (e.g., a left-to-right
depth-first numbering), and break ties in favor of the lowest-numbered
action, then this defines a unique policy at each Max node.  And
consequently, by induction, it defines a unique policy for the entire
MAXQ graph.  Let us call this policy $\pi^*_r$.  We will use the $r$
subscript to denote recursively optimal quantities under an ordered
greedy policy.  Hence, the corresponding value function is $V^*_r$,
and $C^*_r$ and $Q^*_r$ denote the corresponding completion
function and action-value function.  We now prove that the MAXQ-0
algorithm converges to $\pi^*_r$.

\begin{theorem} \label{theorem-converge}
Let $M = \mtuple{S, A, P, R, P_0}$ be either an episodic MDP for which
all deterministic policies are proper or a discounted infinite horizon
MDP with discount factor $\gamma$.  Let $H$ be a MAXQ graph defined
over subtasks $\{M_0, \ldots, M_k\}$ such that the pseudo-reward
function $\tilde{R}_i(s'|s,a)$ is zero for all $i$, $s$, $a$, and
$s'$.  Let $\alpha_t(i) > 0$ be a sequence of constants for each Max
node $i$ such that
\begin{equation}
\lim_{T\rightarrow \infty} \sum_{t=1}^T \alpha_t(i) = \infty \;\;\;\;
\mbox{and} \;\;\;\;
\lim_{T\rightarrow \infty} \sum_{t=1}^T \alpha_t^2(i) < \infty 
\label{alpha-diverge-converge}
\end{equation}
Let $\pi_x(i,s)$ be an ordered GLIE policy at each node $i$ and state $s$
and assume that $|V_t(i,s)|$ and $|C_t(i,s,a)|$ are bounded for all
$t$, $i$, $s$, and $a$.  Then with probability 1, algorithm MAXQ-0
converges to $\pi^*_r$, the unique recursively optimal policy for $M$
consistent with $H$ and $\pi_x$. 
\end{theorem}

\paragraph*{Proof:}
The proof follows an argument similar to those introduced to prove the
convergence of $Q$ learning and $SARSA(0)$
\cite{bt-ndp-96,nc:Jaakkola+Jordan+Singh:1994}.  We will employ the
following result from stochastic approximation theory:

\begin{lemma} (Proposition 4.5 from Bertsekas and Tsitsiklis, 1996)
\label{lemma-sa}
Consider the iteration
\[
r_{t+1}(x) := (1-\alpha_t(x)) r_t(x) + \alpha_t(x)((U r_t)(x) +
w_t(x) + u_t(x)).
\]
Let ${\cal F}_t = \{r_0(x), \ldots, r_t(x), w_0(x), \ldots,
w_{t-1}(x), \alpha_0(x), \ldots, \alpha_t(x), \forall x\}$ be
the entire history of the iteration.

If 
\begin{enumerate}
\item[(a)] The $\alpha_t(i) \geq 0$ satisfy conditions
(\ref{alpha-diverge-converge})

\item[(b)] For every $i$ and $t$ the noise terms $w_t(i)$ satisfy
$E[w_t(i) | {\cal F}_t] = 0$

\item[(c)] Given any norm $|| \cdot ||$ on $R^n$, there exist constants
$A$ and $B$ such that $E[w_t^2(i) | {\cal F}_t] \leq A + B||r_t||^2$.

\item[(d)] There exists a vector $r^*$, a positive vector $\xi$, and a
scalar $\beta \in [0,1)$, such that for all $t$,
\[||Ur_t - r^*||_{\xi} \leq \beta ||r_t - r^*||_{\xi}\]

\item[(e)] There exists a nonnegative random sequence $\theta_t$ that
converges to zero with probability 1 and is such that for all $t$
\[|u_t(x)| \leq \theta_t(||r_t||_{\xi} + 1)\]
\end{enumerate}
then $r_t$ converges to $r^*$ with probability 1.
The notation $|| \cdot ||_{\xi}$ denotes a weighted maximum norm
\[ || A ||_{\xi} = \max_x \frac{|A(x)|}{\xi(x)}. \]
\end{lemma}

The structure of the proof of Theorem \ref{theorem-converge} will be
inductive, starting at the leaves of the MAXQ graph and working toward
the root.  We will employ a different time clock at each node $i$ to
count the number of update steps performed by MAXQ-0 at that node.
The variable $t$ will always refer to the time clock of the current
node $i$.

To prove the base case for any primitive Max node, we note that line 4
of MAXQ-0 is just the standard stochastic approximation algorithm for
computing the expected reward for performing action $a$ in state $s$,
and therefore it converges under the conditions given above.

To prove the recursive case, consider any composite Max node $i$ with
child node $j$.  Let $P_t(s',N|s,j)$ be the transition probability
distribution for performing child action $j$ in state $s$ at time $t$
(i.e., while following the exploration policy in all descendent nodes
of node $j$).  By the inductive assumption, MAXQ-0 applied to $j$ will
converge to the (unique) recursively optimal value function
$V^*_r(j,s)$ with probability 1.  Furthermore, because MAXQ-0 is
following an ordered GLIE policy for $j$ and its descendants,
$P_t(s',N|s,j)$ will converge to $P^*_r(s',N|s,j)$, the unique
transition probability function for executing child $j$ under the
locally optimal policy $\pi^*_r$.  What remains to be shown is that
the update assignment for $C$ (line 12 of the MAXQ-0 algorithm) will
converge to the optimal $C^*_r$ function with probability 1.

To prove this, we will apply Lemma~\ref{lemma-sa}.  We will identify
the $x$ in the lemma with a state-action pair $(s,a)$.  The vector
$r_t$ will be the completion-cost table $C_t(i,s,a)$ for all $s,a$ and
fixed $i$ after $t$ update steps. The vector $r^*$ will be the optimal
completion-cost $C^*_r(i,s,a)$ (again, for fixed $i$).  Define the
mapping $U$ to be
\[
(UC)(i,s,a) = \sum_{s'} P^*_r(s',N|s,a) \gamma^N (\max_{a'} [C(i,s',a') + V^*_r(a',s')])
\]
This is a $C$ update under the MDP $M_i$ assuming that all descendant
value functions, $V^*_r(a,s)$, and transition probabilities, $P^*_r(s',N|s,a)$,
have converged.

\def\sbar{\overline{s}}
\def\nbar{\overline{N}}

To apply the lemma, we must first express the $C$ update formula in
the form of the update rule in the lemma.  Let $\sbar$ be the state
that results from performing $a$ in state $s$.  Line 12 can be written
\begin{eqnarray*}
C_{t+1}(i,s,a) &:=& (1-\alpha_t(i))\cdot C_t(i,s,a) + \alpha_t(i) \cdot \gamma^{\nbar} 
(\max_{a'} [C_t(i,\sbar,a') + V_t(a',\sbar)])\\
  &:=& (1-\alpha_t(i)) \cdot C_t(i,s,a) + \alpha_t(i) \cdot [(UC_t)(i,s,a) + w_t(i,s,a) + u_t(i,s,a)]\\
\end{eqnarray*}
where
\begin{eqnarray*}
w_t(i,s,a) &=& \gamma^{\nbar} \left(\max_{a'} [C_t(i,\sbar,a') +  V_t(a',\sbar)]\right) - \\
&&              \sum_{s',N} P_t(s',N|s,a) \gamma^N \left(\max_{a'} [C_t(i,s',a') + V_t(a',s')]\right)\\
u_t(i,s,a) &=& \sum_{s',N} P_t(s',N|s,a) \gamma^N \left(\max_{a'}
              [C_t(i,s',a') + V_t(a',s')]\right) - \\
 &&             \sum_{s',N} P^*_r(s',N|s,a) \gamma^N \left(\max_{a'} [C_t(i,s',a') + V^*_r(a',s')]\right)
\end{eqnarray*}
Here $w_t(i,s,a)$ is the difference between doing an update at node
$i$ using the single {\it sample point} $\sbar$ drawn according to
$P_t(s',N|s,a)$ and doing an update using the full distribution
$P_t(s',N|s,a)$.  The value of $u_t(i,s,a)$ captures the difference
between doing an update using the current probability transitions
$P_t(s',N|s,a)$ and current value functions of the children
$V_t(a',s')$ and doing an update using the optimal probability
transitions $P^*_r(s',N|s,a)$ and the optimal values of the children
$V^*_r(a',s')$.

We now verify the conditions of Lemma \ref{lemma-sa}.

Condition (a) is assumed in the conditions of the theorem with
$\alpha_t(s,a) = \alpha_t(i)$. 

Condition (b) is satisfied because $\sbar$ is sampled from
$P_t(s',N|s,a)$, so the expected value of the difference is zero. 

Condition (c) follows directly from the assumption that the
$|C_t(i,s,a)|$ and $|V_t(i,s)|$ are bounded.

Condition (d) is the condition that $U$ is a weighted max norm
pseudo-contraction.  We can derive this by starting with the weighted
max norm for $Q$ learning.  It is well known that $Q$ is a weighted
max norm pseudo-contraction \cite{bt-ndp-96} in both the episodic case
where all deterministic policies are proper (and the discount factor
$\gamma = 1$) and in the infinite horizon discounted case (with
$\gamma < 1$).  That is, there exists a
positive vector $\xi$ and a scalar $\beta \in [0,1)$, such that for
all $t$,
\begin{equation}
||TQ_t - Q^*||_{\xi} \leq \beta ||Q_t - Q^*||_{\xi},
\label{eq-t-contraction}
\end{equation}
where $T$ is the operator
\[
(TQ)(s,a) = \sum_{s',N} P(s',N|s,a) \gamma^N [R(s'|s,a) + \max_{a'} Q(s',a')].
\]
Now we will show how to derive the contraction for the $C$ update
operator $U$.  Our plan is to show first how to express the $U$
operator for learning $C$ in terms of the $T$ operator for updating
$Q$ values.  Then we will replace $TQ$ in the contraction equation for
$Q$ learning with $UC$, and show that $U$ is a weighted max-norm
contraction under the same weights $\xi$ and the same $\beta$. 

Recall from Eqn.~(\ref{eq-decomp-q}) that $Q(i,s,a) = C(i,s,a) + V(a,s)$.
Furthermore, the $U$ operator performs its updates using the optimal
value functions of the child nodes, so we can write this as $Q_t(i,s,a)
= C_t(i,s,a) + V^*(a,s)$.  Now once the children of node $i$ have
converged, the $Q$-function version of the Bellman equation for MDP
$M_i$ can be written as
\[
Q(i,s,a) = \sum_{s',N} P^*_r(s',N|s,a) \gamma^N [V^*_r(a,s) + \max_{a'} Q(i,s',a')].
\]
As we have noted before, $V^*_r(a,s)$ plays the role of the immediate
reward function for $M_i$.  Therefore, for node $i$, the $T$ operator
can be rewritten as
\[(TQ)(i,s,a) = \sum_{s',N} P^*_r(s'|s,a) \gamma^N [V^*_r(a,s) + \max_{a'} Q(i,s',a')].
\]
Now we replace $Q(i,s,a)$ by $C(i,s,a) + V^*_r(a,s)$, and obtain
\[
(TQ)(i,s,a) = \sum_{s',N} P^*_r(s',N|s,a) \gamma^N (V^*_r(a,s) + \max_{a'} [C(i,s',a') + V^*_r(a',s')]).
\]
Note that $V^*_r(a,s)$ does not depend on $s'$ or $N$, so we can move it
outside the expectation and obtain
\begin{eqnarray*}
(TQ)(i,s,a) &=& V^*_r(a,s) +  \sum_{s',N} P^*_r(s'|s,a) \gamma^N (\max_{a'}[C(i,s',a') + V^*_r(a',s')])\\
&=& V^*_r(a,s) + (UC)(i,s,a)
\end{eqnarray*}
Abusing notation slightly, we will express this in vector form as
$TQ(i) = V^*_r + UC(i)$.  Similarly, we can write $Q_t(i,s,a) = C_t(i,s,a)
+V^*_r(a,s)$ in vector form as $Q_t(i) = C_t(i) + V^*_r$.  

Now we can substitute these two formulas into the max norm pseudo-contraction
formula for $T$, Eqn.~(\ref{eq-t-contraction}) to obtain
\[||V^*_r + UC_t(i) - (C^*_r(i) + V^*_r)||_{\xi} \leq \beta ||V^*_r + C_t(i) - (V^*_r + C^*_r(i))||_{\xi}.
\]
The $V^*$ terms cancel on both sides of the equation, and we get
\[||UC_t(i) - C^*_r(i)||_{\xi} \leq \beta ||C_t(i) - C^*_r(i)||_{\xi}.
\]

Finally, it is easy verify (e), the most important condition.  By
assumption, the ordered GLIE policies in the child nodes converge with
probability 1 to locally optimal policies for the children.  Therefore
$P_t(s',N|s,a)$ converges to $P^*_r(s',N|s,a)$ for all $s', N, s, $ and
$a$ with probability 1 and $V_t(a,s)$ converges with probability 1 to
$V^*_r(a,s)$ for all child actions $a$.  Therefore, $|u_t|$ converges to
zero with probability 1.  We can trivially construct a sequence
$\theta_t = |u_t|$ that bounds this convergence, so
\[
|u_t(s,a)| \leq \theta_t \leq \theta_t(||C_t(s,a)||_{\xi} + 1).
\]
We have verified all of the conditions of Lemma~\ref{lemma-sa}, so we
can conclude that $C_t(i)$ converges to $C^*_r(i)$ with probability 1.  By
induction, we can conclude that this holds for all nodes in the MAXQ
including the root node, so the value function represented by the MAXQ
graph converges to the unique value function of the recursively optimal
policy $\pi^*_r$. \qed

Algorithm MAXQ-0 can be extended to accelerate learning in the higher
nodes of the graph by a technique that we call ``all states
updating''.  When an action $a$ is chosen for Max node $i$ in state
$s$, the execution of $a$ will move the environment through a sequence
of states $s = s_1, \ldots, s_{N}, s_{N+1} = s'$.  If $a$ was indeed the
best abstract action to choose in $s_1$, then it should also be the
best action to choose (at node $i$) in states $s_2$ through $s_N$.
Hence, we can execute a version of line 12 in MAXQ-0 for each of these
intermediate states as shown in this replacement pseudo-code:

\begin{quote}
\begin{tabbing}
xxxxx\=xxxx\=xxxx\=xxxx\=\kill
12a\> \>\>    {\bf for} $j$ {\bf from} 1 {\bf to} $N$ {\bf do} \\
12b\>\>\>\>      $C_{t+1}(i,s_j,a) := (1-\alpha_t(i)) \cdot C_t(i,s_j,a) +
                \alpha_t(i) \cdot \gamma^{(N+1-j)} max_{a'}\; Q_t(i,s',a')$ \\
12c\>\>\>\>     {\bf end} // for
\end{tabbing}
\end{quote}

In our implementation, as each composite action is executed by MAXQ-0,
it constructs a linked list of the sequence of primitive states that
were visited.  This list is returned when the composite action
terminates.  The parent Max node can then process each state in this
list as shown above.  The parent Max node appends the state
lists that it receives from its children and passes them to its parent
when it terminates.  All experiments in this paper employ all states
updating.

Kaelbling \citeyear{k-hrl:pr-93} introduced a related, but more
powerful, method for accelerating hierarchical reinforcement learning
that she calls ``all goals updating.''  This method is suitable for a
MAXQ hierarchy containing only a root task and one level of composite
tasks.  To understand all goals updating, suppose that for each
primitive action, there are several composite tasks that could have
invoked that primitive action.  In all goals updating, whenever a
primitive action is executed, the equivalent of line 12 of MAXQ-0 is
applied in every composite task that could have invoked that primitive
action.  Sutton, Precup, and Singh \citeyear{sps-bmsm:lprkmts-98}
prove that each of the composite tasks will converge to the optimal
$Q$ values under all goals updating.

All goals updating would work in the MAXQ hierarchy for composite
tasks all of whose children are primitive actions.  However, as we
have seen, at higher levels in the hierarchy, node $i$ needs to obtain
samples of result states drawn according to $P^*(s',N|s,a)$ for
composite tasks $a$.  All goals updating cannot provide these samples,
so it cannot be applied at these higher levels. 

Now that we have shown the convergence of MAXQ-0, let us design a
learning algorithm for arbitrary pseudo-reward functions,
$\tilde{R}_i(s)$.  We could just add the pseudo-reward into MAXQ-0,
but this has the effect of changing the MDP $M$ to have a different
reward function.  The pseudo-rewards ``contaminate'' the values of all
of the completion functions computed in the hierarchy.   The resulting
learned policy will not be recursively optimal for the original MDP. 

This problem can be solved by learning {\it two} completion functions.
The first one, $C(i,s,a)$ is the completion function that we have been
discussing so far in this paper.  It computes the expected reward for
completing task $M_i$ after performing action $a$ in state $s$ and
then following the learned policy for $M_i$.  It is computed without
any reference to $\tilde{R}_i$.  This completion function will be used
by parent tasks to compute $V(i,s)$, the expected reward for
performing action $i$ starting in state $s$.

The second completion function $\tilde{C}(i,s,a)$ is a completion
function that we will use only ``inside'' node $i$ in order to
discover the locally optimal policy for task $M_i$.  This function
will incorporate rewards both from the ``real'' reward function,
$R(s'|s,a)$ and from the pseudo-reward function $\tilde{R}_i(s)$. 

We will employ two different update rules to learn these two
completion functions.  The $\tilde{C}$ function will be learned using
an update rule similar to the Q learning rule in line 12 of MAXQ-0.
But the $C$ function will be learned using an update rule similar to
SARSA(0)---its purpose is to learn the value function for the policy
that is discovered by optimizing $\tilde{C}$.  Pseudo-code for the
resulting algorithm, MAXQ-Q is shown in Table~\ref{tab-maxq-q}.

\begin{table}
\caption{The MAXQ-Q learning algorithm.}
\label{tab-maxq-q}
{\footnotesize
\begin{tabbing}
xxxxx\=xxxx\=xxxx\=xxxx\=\kill
1 \> {\bf function} MAXQ-Q(MaxNode $i$, State $s$)\\[.1in]
2 \> {\bf let} $seq = ()$  be the sequence of states visited while executing $i$\\
3 \> {\bf if} $i$ is a primitive MaxNode \\
4 \> \>  execute $i$, receive $r$, and observe result state $s'$ \\
5 \> \>  $V_{t+1}(i,s) := (1-\alpha_t(i)) \cdot V_t(i,s) + \alpha_t(i) \cdot r_t$ \\
6 \> \>  push $s$ into the beginning of $seq$\\
7 \> {\bf else} \\
8 \> \>  {\bf let} $count = 0$ \\
9 \> \>  {\bf while} $T_i(s)$ is false {\bf do} \\
10 \> \>\>    choose an action $a$ according to the current exploration policy $\pi_x(i,s)$ \\
11 \> \>\>    {\bf let} $childSeq$ = MAXQ-Q$(a, s)$, where $childSeq$
is the sequence of states visited \\
   \> \>\>\>   while executing action $a$.\\
12 \> \>\>    observe result state $s'$ \\
13 \> \>\>    {\bf let} $a^* = \argmax_{a'}\; [\tilde{C}_t (i,s',a') + V_t(a',s')]$\\
14 \> \>\>    {\bf let} $N$ = length$(childSeq)$\\
15 \> \>\>    {\bf for each} $s$ {\bf in} $childSeq$ {\bf do}\\
16 \> \>\>\> $\tilde{C}_{t+1}(i,s,a) := (1-\alpha_t(i)) \cdot \tilde{C}_t(i,s,a) 
   + \alpha_t(i) \cdot \gamma^N [\tilde{R}_i(s') + \tilde{C}_t (i,s',a^*) +
   V_t(a^*,s)]$\\
17 \> \>\>\>   $C_{t+1}(i,s,a) := (1-\alpha_t(i)) \cdot C_t(i,s,a) + \alpha_t(i) \cdot \gamma^N [C_t (i,s',a^*) + V_t(a^*,s')]$\\
18 \> \>\>\>  $N := N - 1$\\
19 \> \>\>\>  {\bf end} // for\\
20 \> \>\>   {\bf append} $childSeq$ onto the front of $seq$\\
21 \> \>\>   $s := s'$\\
22 \> \>\>    {\bf end} // while \\
23 \> \>  {\bf end} // else \\
24 \> {\bf return} $seq$ \\
25 \> {\bf end} MAXQ-0
\end{tabbing}
} 
\end{table}

The key step is at lines 16 and 17.  In line 16, MAXQ-Q first updates
$\tilde{C}$ using the value of the greedy action, $a^*$, in the
resulting state.  This update includes the pseudo-reward $\tilde{R}_i$. 
Then in line 17, MAXQ-Q updates $C$ using this {\it
same greedy action} $a^*$, even if this would not be the greedy action
according to the ``uncontaminated'' value function.  This update, of
course, does not include the pseudo-reward function. 

It is important to note that whereever $V_t(a,s)$ appears in this
pseudo-code, it refers to the ``uncontaminated'' value function of
state $s$ when executing the Max node $a$.  This is computed
recursively in exactly the same way as in MAXQ-0.

Finally, note that the pseudo-code also incorporates all-states
updating, so each call to MAXQ-Q returns a list of all of the states
that were visited during its execution, and the updates of lines 16
and 17 are performed for each of those states.  The list of states is
ordered most-recent-first, so the states are updated starting with the
last state visited and working backward to the starting state, which
helps speed up the algorithm. 

When MAXQ-Q has converged, the resulting recursively optimal policy is
computed at each node by choosing the action $a$ that maximizes
$\tilde{Q}(i,s,a) = \tilde{C}(i,s,a) + V(a,s)$ (breaking ties
according to the fixed ordering established by the ordered GLIE
policy).  It is for this reason that we gave the name ``Max nodes'' to
the nodes that represent subtasks (and learned policies) within the
MAXQ graph.  Each Q node $j$ with parent node $i$ stores both
$\tilde{C}(i,s,j)$ and $C(i,s,j)$, and it computes both
$\tilde{Q}(i,s,j)$ and $Q(i,s,j)$ by invoking its child Max node $j$.
Each Max node $i$ takes the maximum of these $Q$ values and computes
either $V(i,s)$ or computes the best action, $a^*$ using $\tilde{Q}$.

\begin{corollary}
Under the same conditions as Theorem~\ref{theorem-converge}, MAXQ-Q
converges the unique recursively optimal policy for MDP $M$ defined by
MAXQ graph $H$, pseudo-reward functions $\tilde{R}$, and ordered GLIE
exploration policy $\pi_x$.
\end{corollary}
\paragraph*{Proof:}
The argument is identical to, but more tedious than, the proof of
Theorem \ref{theorem-converge}.  The proof of convergence of the
$\tilde{C}$ values is identical to the original proof for the $C$
values, but it relies on proving convergence of the ``new'' $C$ values
as well, which follows from the same weighted max norm
pseudo-contraction argument.  \qed

\section{State Abstraction}

There are many reasons to introduce hierarchical reinforcement
learning, but perhaps the most important reason is to create
opportunities for state abstraction.  When we introduced the simple
taxi problem in Figure~\ref{fig-taxi}, we pointed out that within each
subtask, we can ignore certain aspects of the state space.  For
example, while performing a ${\sf MaxNavigate}(t)$, the taxi should
make the same navigation decisions regardless of whether the passenger
is in the taxi.  The purpose of this section is to formalize the
conditions under which it is safe to introduce such state abstractions
and to show how the convergence proofs for MAXQ-Q can be extended to
prove convergence in the presence of state abstraction.  Specifically,
we will identify five conditions that permit the ``safe'' introduction
of state abstractions.

Throughout this section, we will use the taxi problem as a running
example, and we will see how each of the five conditions will permit
us to reduce the number of distinct values that must be stored in
order to represent the MAXQ value function decomposition.  To
establish a starting point, let us compute the number of values that
must be stored for the taxi problem {\it without} any state
abstraction. 

The MAXQ representation must have tables for each of the $C$ functions
at the internal nodes and the $V$ functions at the leaves.  First, at
the six leaf nodes, to store $V(i,s)$, we must store 500 values at
each node (because there are 500 states; 25 locations, 4 possible
destinations for the passenger, and 5 possible current locations for
the passenger (the four special locations and inside the taxi
itself)).  Second, at the root node, there are two children, which
requires $2 \times 500 = 1000$ values.  Third, at the {\sf MaxGet} and
{\sf MaxPut} nodes, we have 2 actions each, so each one requires 1000
values, for a total of 2000.  Finally, at ${\sf MaxNavigate}(t)$,
we have four actions, but now we must also consider the target
parameter $t$, which can take four possible values.  Hence, there are
effectively 2000 combinations of states and $t$ values for each
action, or 8000 total values that must be represented.  In total,
therefore, the MAXQ representation requires 14,000 separate quantities
to represent the value function.

To place this number in perspective, consider that a flat Q learning
representation must store a separate value for each of the six
primitive actions in each of the 500 possible states, for a total of
3,000 values.  Hence, we can see that without state abstraction, the
MAXQ representation requires more than four times the memory of a flat
Q table!

\subsection{Five Conditions that Permit State Abstraction}

We now introduce five conditions that permit the introduction of state
abstractions.  For each condition, we give a definition and then prove
a lemma which states that if the condition is satisfied, then the
value function for some corresponding class of policies can be
represented abstractly (i.e., by abstract versions of the $V$ and $C$
functions).  For each condition, we then provide some rules 
for identifying when that condition can be satisfied and give examples
from the taxi domain. 

We begin by introducing some definitions and notation. 

\begin{definition} 
Let $M$ be a MDP and $H$ be a MAXQ graph defined over $M$.  Suppose
that each state $s$ can be written as a vector of values of a set of
state variables.  At each Max node $i$, suppose the state variables
are partitioned into two sets $X_i$ and $Y_i$, and let $\chi_i$ be a
function that projects a state $s$ onto only the values of the
variables in $X_i$.  Then $H$ combined with $\chi_i$ is called a {\em
state-abstracted MAXQ graph}.
\end{definition}

In cases where the state variables can be partitioned, we will often
write $s = (x,y)$ to mean that a state $s$ is represented by a vector
of values for the state variables in $X$ and a vector of values for
the state variables in $Y$.  Similarly, we will sometimes write
$P(x',y',N|x,y,a)$, $V(a,x,y)$, and $\tilde{R}_a(x',y')$ in place of
$P(s',N|s,a)$, $V(a,s)$, and $\tilde{R}_a(s')$, respectively.

\begin{definition}
An {\em abstract hierarchical policy} for MDP $M$ with
state-abstracted MAXQ graph $H$ and associated abstraction functions
$\chi_i$, is a hierarchical policy in which each policy $\pi_i$
(corresponding to subtask $M_i$) satisfies the condition that for any
two states $s_1$ and $s_2$ such that $\chi_i(s_1) = \chi_i(s_2)$,
$\pi_i(s_1) = \pi_i(s_2)$.  (When $\pi_i$ is a stationary stochastic
policy, this is interpreted to mean that the probability distributions
for choosing actions are the same in both states.)
\end{definition}

In order for MAXQ-Q to converge in the presence of state abstractions,
we will require that at all times $t$ its (instantaneous) exploration
policy is an abstract hierarchical policy.  One way to achieve this is
to construct the exploration policy so that it only uses information
from the relevant state variables in deciding what action to perform.
Boltzmann exploration based on the (state-abstracted) Q values,
$\epsilon$-greedy exploration, and counter-based exploration based on
abstracted states are all abstract exploration policies.
Counter-based exploration based on the full state space is not an
abstract exploration policy.

Now that we have introduced our notation, let us describe and analyze
the five abstraction conditions.  We have identified three different
kinds of conditions under which abstractions can be introduced.  The
first kind involves eliminating irrelevant variables within a subtask
of the MAXQ graph.  Under this form of abstraction, nodes toward the
leaves of the MAXQ graph tend to have very few relevant variables, and
nodes higher in the graph have more relevant variables.  Hence, this
kind of abstraction is most useful at the lower levels of the MAXQ
graph.

The second kind of abstraction arises from ``funnel'' actions.  These
are macro actions that move the environment from some large number of
initial states to a small number of resulting states.  The completion
cost of such subtasks can be represented using a number of values
proportional to the number of resulting states.  Funnel actions tend
to appear higher in the MAXQ graph, so this form of abstraction is most
useful near the root of the graph. 

The third kind of abstraction arises from the structure of the MAXQ
graph itself.  It exploits the fact that large parts of the state
space for a subtask may not be reachable because of the termination
conditions of its ancestors in the MAXQ graph. 

We begin by describing two abstraction conditions of the first type.
Then we will present two conditions of the second type.  And finally,
we describe one condition of the third type.

\subsubsection{Condition 1: Max Node Irrelevance}

The first condition arises when a set of state variables is irrelevant
to a Max node. 

\begin{definition} \label{definition-max-node-irrelevance}
Let $M_i$ be a Max node in a MAXQ graph $H$ for MDP $M$.  A set of
state variables $Y$ is {\em irrelevant to node $i$} if the state
variables of $M$ can be partitioned into two sets $X$ and $Y$ such
that for any stationary abstract hierarchical policy $\pi$ executed by
the descendants of $i$, the following two properties hold:
\begin{itemize}
\item the state transition probability distribution
$P^{\pi}(s',N|s,a)$ at node $i$ can be factored into the product of
two distributions:
\begin{equation} \label{eq-p-independence}
P^{\pi}(x',y',N|x,y,a) = P^{\pi}(y'|y,a) \cdot P^{\pi}(x',N|x,a),
\end{equation}
where $y$ and $y'$ give values for the variables in $Y$, and $x$ and
$x'$ give values for the variables in $X$. 

\item for any pair of states $s_1=(x,y_1)$ and $s_2=(x,y_2)$ such that
$\chi(s_1) = \chi(s_2) = x$, and any child action $a$, $V^{\pi}(a,s_1)
= V^{\pi}(a,s_2)$ and $\tilde{R}_i(s_1) = \tilde{R}_i(s_2)$.
\end{itemize}
\end{definition}

\begin{lemma} \label{lemma-max-node-irrelevance}
Let $M$ be an MDP with full-state MAXQ graph $H$, and suppose that
state variables $Y_i$ are irrelevant for Max node $i$.  Let $\chi_i(s) =
x$ be the associated abstraction function that maps $s$ onto the
remaining relevant variables $X_i$.  Let $\pi$ be any abstract
hierarchical policy.  Then the action-value function $Q^{\pi}$ at node
$i$ can be represented compactly, with only one value of the
completion function $C^{\pi}(i,s,j)$ for each equivalence class of
states $s$ that share the same values on the relevant variables.

Specifically $Q^{\pi}(i,s,j)$ can be
computed as follows:
\[ Q^{\pi}(i,s,j) = V^{\pi}(j,\chi_i(s)) + C^{\pi}(i,\chi_i(s),j) \]
where
\[
C^{\pi}(i,x,j) = \sum_{x',N} P^{\pi}(x',N|x,j) \cdot \gamma^N[
V^{\pi}(\pi(x'),x') + \tilde{R}_i(x') + C^{\pi}(i,x',\pi(x'))],
\]
where $V^{\pi}(j',x') = V^{\pi}(j',x',y_0)$, $\tilde{R}_i(x') =
\tilde{R}_i(x',y_0)$, and $\pi(x) = \pi(x,y_0)$ for some arbitrary value
$y_0$ for the irrelevant state variables $Y_i$.  
\end{lemma}
\paragraph{Proof:}
Define a new MDP $\chi_i(M_i)$ at node $i$ as follows:
\begin{itemize}
\item States: $X = \{x \:|\: \chi_i(s) = x,\mbox{ for some } s \in S\}$.
\item Actions: $A$.
\item Transition probabilities: $P^{\pi}(x',N|x,a)$
\item Reward function: $V^{\pi}(a,x) + \tilde{R}_i(x')$
\end{itemize}
Because $\pi$ is an abstract policy, its decisions are the same for
all states $s$ such that $\chi_i(s) = x$ for some $x$.  Therefore, it is
also a well-defined policy over $\chi_i(M_i)$.  The action-value function
for $\pi$ over $\chi_i(M_i)$ is the unique solution to the following
Bellman equation:
\begin{equation} \label{eq-bellman-for-chi-M}
Q^{\pi}(i,x,j) = V^{\pi}(j,x) + \sum_{x',N} P^{\pi}(x',N|x,j) \cdot \gamma^N
        [\tilde{R}_i(x') + Q^{\pi}(i,x',\pi(x'))]
\end{equation}
Compare this to the Bellman equation over $M_i$:
\begin{equation} \label{eq-bellman-for-pi}
Q^{\pi}(i,s,j) = V^{\pi}(j,s) + \sum_{s',N} P^{\pi}(s',N|s,j) \cdot \gamma^N
        [ \tilde{R}_i(s') + Q^{\pi}(i,s',\pi(s'))]
\end{equation}
and note that $V^{\pi}(j,s) = V^{\pi}(j,\chi(s)) = V^{\pi}(j,x)$ and
$\tilde{R}_i(s') = \tilde{R}_i(\chi(s')) = \tilde{R}_i(x')$.  Furthermore,
we know that the distribution $P^{\pi}$ can be factored into separate
distributions for $Y_i$ and $X_i$.  Hence, we can rewrite
(\ref{eq-bellman-for-pi}) as
\[
Q^{\pi}(i,s,j) = V^{\pi}(j,x) + \sum_{y'} P(y'|y,j) \sum_{x',N}
        P^{\pi}(x',N|x,j) \cdot \gamma^N [ \tilde{R}_i(x') + Q^{\pi}(i,s',\pi(s'))]
\]
The right-most sum does not depend on $y$ or $y'$, so the sum over
$y'$ evaluates to 1, and can be eliminated to give
\begin{equation} \label{eq-bellman-for-pi-2}
Q^{\pi}(i,s,j) = V^{\pi}(j,x) + \sum_{x',N} P^{\pi}(x',N|x,j) \cdot \gamma^N [ \tilde{R}_i(x') + Q^{\pi}(i,s',\pi(s'))].
\end{equation}

Finally, note that equations (\ref{eq-bellman-for-chi-M}) and
(\ref{eq-bellman-for-pi-2}) are identical except for the expressions
for the $Q$ values.  Since the solution to the Bellman equation is
unique, we must conclude that
\[
Q^{\pi}(i,s,j) = Q^{\pi}(i,\chi(s),j).
\]
We can rewrite the right-hand side to obtain
\[
Q^{\pi}(i,s,j) = V^{\pi}(j,\chi(s)) + C^{\pi}(i,\chi(s),j),
\]
where
\[
C^{\pi}(i,x,j) = \sum_{x',N} P(x',N|x,j) \cdot \gamma^N[
V^{\pi}(\pi(x'),x') + \tilde{R}_i(x') + C^{\pi}(i,x',\pi(x'))].
\]
\qed

Of course we are primarily interested in being able to discover and
represent the {\it optimal} policy at each node $i$.  The following
corollary shows that the optimal policy is an abstract policy, and
hence, that it can be represented abstractly. 

\begin{corollary} \label{corollary-optimal-is-abstract}
Consider the same conditions as
Lemma~\ref{lemma-max-node-irrelevance}, but with the change that the
abstract hierarchical policy $\pi$ is executed only by the descendants
of node $i$, but not by node $i$. Let $\rho$ be an ordering over
actions.  Then the optimal ordered policy $\pi^*_{\rho}$ at node $i$ is an
abstract policy, and its action-value function can be represented
abstracted.
\end{corollary} 
\paragraph*{Proof:}
Define the policy $\omega^*_{\rho}$ to be the optimal ordered policy
over the abstract MDP $\chi(M)$, and let $Q^*(i,x,j)$ be the
corresponding optimal action-value function.  Then by the same
argument given above, $Q^*$ is also a solution to the optimal Bellman
equation for the original MDP.  This means that the policy
$\pi^*_{\rho}$ defined by $\pi^*_{\rho}(s) = \omega^*(\chi(s))$ is an
optimal ordered policy, and by construction, it is an abstract policy.
\qed

As stated, this condition appears quite difficult to satisfy, since it
requires that the state transition probability distribution factor
into $X$ and $Y$ components for all possible abstract hierarchical
policies.  However, in practice, this condition is often satisfied. 

For example, let us consider the ${\sf Navigate}(t)$ subtask.  The
source and destination of the passenger are irrelevant to the
achievement of this subtask.  Any policy that successfully completes
this subtask will have the same value function regardless of the
source and destination locations of the passenger.  (Any policy that
does not complete the subtask will have the same value function also,
but all states will have a value of $-\infty$.)  By abstracting away
the passenger source and destination, we obtain a huge savings in
space.  Instead of requiring 8000 values to represent the $C$
functions for this task, we require only 400 values (4 actions, 25
locations, 4 possible values for $t$).

One rule for noticing cases where this abstraction condition holds is
to examine the subgraph rooted at the given Max node $i$.  If a set of
state variables is irrelevant to the leaf state transition
probabilities and reward functions and also to all pseudo-reward
functions and termination conditions in the subgraph, then those
variables satisfy the Max Node Irrelevance condition:

\begin{lemma}
Let $M$ be an MDP with associated MAXQ graph $H$, and let $i$ be a Max
node in $H$.  Let $X_i$ and $Y_i$ be a partition of the state variables
for $M$.  A set of state variables $Y_i$ is irrelevant to node $i$ if
\begin{itemize}
\item For each primitive leaf node $a$ that is a descendant of $i$, 
$P(x',y'|x,y,a) = P(y'|y,a) P(x'|x,a)$ and $R(x',y'|x,y,a) =
R(x'|x,a)$,
\item For each internal node $j$ that is equal to node $i$ or is a
descendent of $i$ , $\tilde{R}_j(x',y') = \tilde{R}_j(x')$ and
the termination predicate $T_j(x',y')$ is true iff $T_j(x')$.
\end{itemize}
\end{lemma}
\paragraph*{Proof:}
We must show that any abstract hierarchical policy will give rise to
an SMDP at node $i$ whose transition probability distribution factors
and whose reward function depends only on $X_i$.  By definition, any
abstract hierarchical policy will choose actions based only upon
information in $X_i$.  Because the primitive probability transition
functions factor into an independent component for $X_i$ and since the
termination conditions at all nodes below $i$ are based only on the
variables in $X_i$, the probability transition function
$P_i(x',y',N|x,y,a)$ must also factor into $P_i(y'|y,a)$ and
$P_i(x',N|x,a)$.  Similarly, all of the reward functions $V(j,x,y)$
must be equal to $V(j,x)$, because all rewards received within the
subtree (either at the leaves or through pseudo-rewards) depend only
on the variables in $X_i$.  Therefore, the variables in $Y_i$ are
irrelevant for Max node $i$. \qed

In the Taxi task, the primitive navigation actions, {\sf North}, {\sf
South}, {\sf East}, and {\sf West} only depend on the location of the
taxi and not on the location of the passenger.  The pseudo-reward
function and termination condition for the ${\sf MaxNavigate}(t)$ node only
depend on the location of the taxi (and the parameter $t$).  Hence,
this lemma applies, and the passenger source and destination are
irrelevant for the {\sf MaxNavigate} node. 

\subsubsection{Condition 2: Leaf Irrelevance}

The second abstraction condition describes situations under which we
can apply state abstractions to leaf nodes of the MAXQ graph.  For
leaf nodes, we can obtain a stronger result than
Lemma~\ref{lemma-max-node-irrelevance} by using a slightly weaker
definition of irrelevance.

\begin{definition} {\bf (Leaf Irrelevance)} A set of state variables
$Y$ is {\em irrelevant for a primitive action} $a$ of a MAXQ graph if
for all states $s$ the expected value of the reward function,
\[V(a,s) = \sum_{s'} P(s'|s,a) R(s'|s,a)\]
does not depend on any of the values of the state
variables in $Y$.  In other words, for any pair of states $s_1$ and
$s_2$ that differ only in their values for the variables in $Y$, 
\[\sum_{s_1'} P(s_1'|s_1,a) R(s_1'|s_1,a) = \sum_{s_2'} P(s_2'|s_2,a) R(s_2'|s_2,a).\]
\end{definition}

If this condition is satisfied at leaf $a$, then the following lemma
shows that we can represent its value function $V(a,s)$ compactly.

\begin{lemma} \label{lemma-leaf-irrelevance}
Let $M$ be an MDP with full-state MAXQ graph $H$, and suppose that
state variables $Y$ are irrelevant for leaf node $a$.  Let $\chi(s) =
x$ be the associated abstraction function that maps $s$ onto the
remaining relevant variables $X$.  Then we can represent $V(a,s)$ for
any state $s$ by an abstracted value function $V(a,\chi(s)) = V(a,x)$.
\end{lemma}
\paragraph{Proof:}
According to the definition of Leaf Irrelevance, any two states that
differ only on the irrelevant state variables have the same value for
$V(a,s)$.  Hence, we can represent this unique value by $V(a,x)$. 
\qed

Here are two rules for finding cases where Leaf Irrelevance applies. 
The first rule shows that if the probability distribution factors,
then we have Leaf Irrelevance. 

\begin{lemma}
Suppose the probability transition function for primitive action $a$,
$P(s'|s,a)$, factors as $P(x',y'|x,y,a) = P(y'|y,a) P(x'|x,a)$ and the
reward function satisfies $R(s'|s,a) = R(x'|x,a)$.  Then the variables
in $Y$ are irrelevant to the leaf node $a$.
\end{lemma}
\paragraph*{Proof:}
Plug in to the definition of $V(a,s)$ and simplify. 
\begin{eqnarray*}
V(a,s) &=& \sum_{s'} P(s'|s,a) R(s'|s,a)\\
       &=& \sum_{x',y'} P(y'|y,a) P(x'|x,a) R(x'|x,a) \\
       &=& \sum_{y'} P(y'|y,a) \sum_{x'} P(x'|x,a) R(x'|x,a)\\
       &=& \sum_{x'} P(x'|x,a) R(x'|x,a)
\end{eqnarray*}
Hence, the expected reward for the action $a$ depends only on the
variables in $X$ and not on the variables in $Y$.  \qed

\begin{lemma}
Let $R(s'|s,a) = r_a$ be the reward function for action $a$ in MDP $M$,
which is always equal to a constant $r_a$.  Then the entire state $s$
is irrelevant to the primitive action $a$.
\end{lemma}
\paragraph*{Proof:}
\begin{eqnarray*}
V(a,s) &=& \sum_{s'} P(s'|s,a) R(s'|s,a)\\
       &=& \sum_{s'} P(s'|s,a) r_a\\
       &=& r_a.
\end{eqnarray*}
This does not depend on $s$, so the entire state is irrelevant to the
primitive action $a$. \qed

This lemma is satisfied by the four leaf nodes {\sf North}, {\sf
South}, {\sf East}, and {\sf West} in the taxi task, because their
one-step reward is a constant $(-1)$.  Hence, instead of requiring 2000
values to store the $V$ functions, we only need 4 values---one for
each action.  Similarly, the expected rewards of the {\sf Pickup} and
{\sf Putdown} actions each require only 2 values, depending on whether
the corresponding actions are legal or illegal.  Hence, together, they
require 4 values, instead of 1000 values. 

\subsubsection{Condition 3: Result Distribution Irrelevance}

Now we consider a condition that results from ``funnel'' actions.

\begin{definition} {\bf (Result Distribution Irrelevance).}
A set of state variables $Y_j$ is {\em irrelevant for the result
distribution of action} $j$ if, for all abstract policies $\pi$
executed by node $j$ and its descendants in the MAXQ hierarchy, the
following holds: for all pairs of states $s_1$ and $s_2$ that differ
only in their values for the state variables in $Y_j$,
\[P^{\pi}(s',N|s_1,j) = P^{\pi}(s',N|s_2,j)\]
for all $s'$ and $N$. 
\end{definition}

\begin{lemma} \label{lemma-result-distribution-irrelevance}
Let $M$ be an MDP with full-state MAXQ graph $H$, and suppose that the
set of state variables $Y_j$ is irrelevant to the result distribution of
action $j$, which is a child of Max node $i$. Let $\chi_{ij}$ be the
associated abstraction function: $\chi_{ij}(s) = x$.  Then we can define an
abstract completion cost function $C^{\pi}(i,\chi_{ij}(s),j)$ such that for
all states $s$,
\[
C^{\pi}(i,s,j) = C^{\pi}(i,\chi_{ij}(s),j).
\]
\end{lemma}
\paragraph{Proof:}
The completion function for fixed policy $\pi$ is defined as follows:
\begin{equation} \label{eq-bellman-rdi}
C^{\pi}(i,s,j) = \sum_{s',N} P(s',N|s,j) \cdot \gamma^N
[\tilde{R}_i(s') + Q^{\pi}(i,s')].
\end{equation}
Consider any two states $s_1$ and $s_2$, such that $\chi_{ij}(s_1) =
\chi_{ij}(s_2) = x$.  Under Result Distribution Irrelevance, their
transition probability distributions are the same.  Hence, the
right-hand sides of (\ref{eq-bellman-rdi}) have the same value, and we
can conclude that 
\[
C^{\pi}(i,s_1,j) = C^{\pi}(i,s_2,j).
\]
Therefore, we can define an abstract completion function, $C^{\pi}(i,x,j)$
to represent this quantity.  \qed

It might appear that this condition would rarely be satisfied, and
indeed, for infinite horizon discounted problems, this is true.
Consider, for example, the {\sf Get} subroutine under an optimal
policy for the taxi task.  No matter what location that taxi has in
state $s$, the taxi will be at the passenger's starting location when
the {\sf Get} finishes executing (i.e., because the taxi will have
just completed picking up the passenger).  Hence, the starting
location is irrelevant to the resulting location of the taxi.  In the
discounted cumulative reward setting, however, the number of steps $N$
required to complete the {\sf Get} action will depend very much on the
starting location of the taxi.  Consequently, $P(s',N|s,a)$ is not
necessarily the same for any two states $s$ with different starting
locations even though $s'$ is always the same.

The important lesson to draw from this is that discounting interferes
with introducing state abstractions based on ``funnel''
operators---the MAXQ framework is therefore less effective when
applied in the discounted setting.

However, if we restrict attention to the episodic, undiscounted
setting, then the result distribution, $P(s'|s,a)$, no longer depends
on $N$, and the Result Distribution Irrelevance condition is
satisfied.  Fortunately, the Taxi task is an undiscounted,
finite-horizon task, so we can represent $C({\sf Root}, s, {\sf Get})$
using 16 distinct values, because there are 16 equivalence classes of
states (4 source locations times 4 destination locations).  This is
much less than the 500 quantities in the unabstracted representation.

``Funnel'' actions arise in many hierarchical reinforcement learning
problems.  For example, abstract actions that move a robot to a
doorway or that move a car onto the entrance ramp of a freeway have
this property.  The Result Distribution Irrelevance condition is
applicable in all such situations as long as we are in the
undiscounted setting.  

\subsubsection{Condition 4: Termination}

The fourth condition is closely related to the ``funnel' property.  It
applies when a subtask is guaranteed to cause its parent task to
terminate in a goal state.  In a sense, the subtask is funneling the
environment into the set of states described by the goal predicate of
the parent task.

\begin{lemma} {\bf (Termination).} \label{lemma-termination}
Let $M_i$ be a task in a MAXQ graph such that for all states $s$ where
the goal predicate $G_i(s)$ is true, the pseudo-reward function
$\tilde{R}_i(s) = 0$.  Suppose there is a child task $a$ and state $s$
such that for all hierarchical policies $\pi$,
\[ \lforall{s'} P^{\pi}_i(s',N|s,a) > 0  \limplies  G_i(s').\]
(i.e., if $s'$ is a possible result state of applying $a$ in $s$, then
$s'$ is a goal terminal state for task $i$.)  

Then for any policy executed at node $i$, the completion cost $C(i, s,
a)$ is zero and does not need to be explicitly represented.
\end{lemma}
\paragraph*{Proof:}  By the assumptions in the lemma, with probability
1 the completion cost is zero for any action that results in a goal
terminal state.  \qed

For example, in the Taxi task, in all states where the taxi is holding
the passenger, the {\sf Put} subroutine will succeed and result in a
goal terminal state for {\sf Root}.  This is because the termination
predicate for {\sf Put} (i.e., that the passenger is at his or her
destination location) implies the goal condition for {\sf Root} (which
is the same).  This means that $C({\sf Root}, s, {\sf Put})$ is
uniformly zero, for all states $s$ where {\sf Put} is not terminated.

It is easy to detect cases where the Termination condition is
satisfied.  We only need to compare the termination predicate of a
subtask with the goal predicate of the parent task.  If the first
implies the second, then the termination condition is satisfied. 

\subsubsection{Condition 5: Shielding}

The shielding condition arises from the structure of the MAXQ graph. 

\begin{lemma} {\bf (Shielding).} \label{lemma-shielding}
Let $M_i$ be a task in a MAXQ graph and $s$ be a state such that for
all paths from the root of the graph down to node $M_i$ there exists a
subtask $j$ (possibly equal to $i$) whose termination predicate
$T_j(s)$ is true, then the $Q$ nodes of $M_i$ do not need to represent
$C$ values for state $s$.
\end{lemma}
\paragraph*{Proof:} Task $i$ cannot be executed in state $s$, so no
$C$ values need to be estimated. \qed

As with the Termination condition, the Shielding condition can be
verified by analyzing the structure of the MAXQ graph and identifying
nodes whose ancestor tasks are terminated. 

In the Taxi task, a simple example of this arises in the {\sf Put}
task, which is terminated in all states where the passenger is not in
the taxi.  This means that we do not need to represent $C({\sf Root},
s, {\sf Put})$ in these states.  The result is that, when combined
with the Termination condition above, we do not need to explcitly
represent the completion function for {\sf Put} at all!

\subsubsection{Dicussion}

By applying these five abstraction conditions, we obtain the following
``safe'' state abstractions for the Taxi task:
\begin{itemize}
\item {\sf North}, {\sf South}, {\sf East}, and {\sf West}.  These
terminal nodes require one quantity each, for a total of four values.
(Leaf Irrelevance).

\item {\sf Pickup} and {\sf Putdown} each require 2 values (legal and illegal
states), for a total of four.  (Leaf Irrelevance.)

\item ${\sf QNorth}(t)$, ${\sf QSouth}(t)$, ${\sf QEast}(t)$, and ${\sf
QWest}(t)$ each require 100 values (four values for $t$ and 25
locations).  (Max Node Irrelevance.)

\item ${\sf QNavigateForGet}$ requires 4 values (for the four possible
source locations).  (The passenger destination is Max Node Irrelevant for
{\sf MaxGet}, and the taxi starting location is Result Distribution
Irrelevant for the {\sf Navigate} action.)


\item ${\sf QPickup}$ requires 100 possible values, 4 possible source
locations and 25 possible taxi locations.  (Passenger destination is
Max Node Irrelevant to {\sf MaxGet}.)

\item ${\sf QGet}$ requires 16 possible values (4 source locations, 4
destination locations). (Result Distribution Irrelevance.)

\item ${\sf QNavigateForPut}$ requires only 4 values (for the four
possible destination locations). (The passenger source and destination
are Max Node Irrelevant to {\sf MaxPut}; the taxi location is Result
Distribution Irrelevant for the {\sf Navigate} action.)

\item ${\sf QPutdown}$ requires 100 possible values (25 taxi
locations, 4 possible destination locations).  (Passenger source is
Max Node Irrelevant for {\sf MaxPut}.)

\item ${\sf QPut}$ requires 0 values.  (Termination and Shielding.)
\end{itemize}

This gives a total of 632 distinct values, which is much less than the
3000 values required by flat Q learning.  Hence, we can see that by
applying state abstractions, the MAXQ representation can give a much
more compact representation of the value function.  A key thing to
note is that these state abstractions cannot be exploited with the flat
representation of the value function. 

What prior knowledge is required on the part of a programmer in order
to introduce these state abstractions?  It suffices to know some
general constraints on the one-step reward functions, the one-step
transition probabilities, and termination predicates, goal predicates,
and pseudo-reward functions within the MAXQ graph.  Specifically, the
Max Node Irrelevance and Leaf Irrelevance conditions require simple
analysis of the one-step transition function and the reward and
pseudo-reward functions.  Opportunities to apply the Result
Distribution Irrelevance condition can be found by identifying
``funnel'' effects that result from the definitions of the termination
conditions for operators.  Similarly, the Shielding and Termination
conditions only require analysis of the termination predicates of the
various subtasks.  Hence, applying these five conditions to introduce
state abstractions is a straightforward process, and once a model of
the one-step transition and reward functions has been learned, the
abstraction conditions can be checked to see if they were satisfied.

\comment{
If we are only interested in the value function of a fixed
hierarchical policy, there is one other opportunity to reduce the
number of values that need to be stored.  Specifically, we only need
to represent the $C(i,s,j)$ function for states $s$ and actions $j$
such that $\pi_i(s) = j$.  This can give additional substantial
reductions in memory.  For example, we might have a good hand-coded
policy that will never try to {\sf Pickup} the passenger unless the
taxi is at the source location.  This means that we only need four
distinct values for $C({\sf Get}, s, {\sf Pickup})$ instead of 100.
But if we are interested in non-hierarchical execution, as described
below, then we will need to represent the completion function for all
pairs of states and actions (abstracted according to the conditions
presented above).}

\subsection{Convergence of MAXQ-Q with State Abstraction}

We have shown that state abstractions can be safely introduced into
the MAXQ value function decomposition under the five conditions
described above.  However, these conditions only guarantee that the
value function of any fixed abstract hierarchical policy can be
represented---they do not show that the optimal policy can be
represented, nor do they show that the MAXQ-Q learning algorithm will
find the optimal policy.  The goal of this section is to prove these
two results: (a) that the ordered recursively-optimal policy is an
abstract policy and (b) that MAXQ-Q will converge to this policy when
applied to a MAXQ graph with safe state abstractions.

\begin{lemma}  \label{lemma-ro-policy-is-abstract}
Let $M$ be an MDP with full-state MAXQ graph $H$ and abstract-state
MAXQ graph $\chi(H)$ where the abstractions satisfy the five
conditions given above.  Let $\rho$ be an ordering over all actions in
the MAXQ graph.  Then the following statements are true:
\begin{itemize}
\item The unique ordered recursively-optimal policy $\pi^*_r$ defined by $M$,
$H$, and $\rho$ is an abstract policy (i.e., it depends only on the
relevant state variables at each node),
\item The $C$ and $V$ functions in $\chi(H)$ can represent the
projected value function of $\pi^*_r$. 
\end{itemize}
\end{lemma}
\paragraph*{Proof:}
The five abstraction lemmas tell us that if the ordered
recursively-optimal policy is abstract, then the $C$ and $V$ functions
of $\chi(H)$ can represent its value function.  Hence, the heart of
this lemma is the first claim.  The last two forms of abstraction
(Shielding and Termination) do not place any restrictions on abstract
policies, so we ignore them in this proof. 

The proof is by induction on the levels of the MAXQ graph, starting at
the leaves.  As a base case, let us consider a Max node $i$ all of
whose children are primitive actions.  In this case, there are no
policies executed {\it within} the children of the Max node.  Hence if
variables $Y_i$ are irrelevant for node $i$, then we can apply our
abstraction lemmas to represent the value function of any policy at
node $i$---not just abstract policies.  Consequently, the value
function of any optimal policy for node $i$ can be represented, and it
will have the property that $Q^*(i,s_1,a) = Q^*(i,s_2,a)$ for any
states $s_1$ and $s_2$ such that $\chi_i(s_1) = \chi_i(s_2)$.

Now let us impose the action ordering $\rho$ to compute the optimal
{\it ordered} policy.  Consider two actions $a_1$ and $a_2$ such that
$\rho(a_1,a_2)$ (i.e., $\rho$ prefers $a_1$), and suppose that there
is a ``tie'' in the $Q^*$ function at state $s_1$ such that the values
\[
Q^*(i,s_1,a_1) = Q^*(i,s_1,a_2)
\]
and they are the only two actions that maximize $Q^*$ in this state.
Then the optimal ordered policy must choose $a_1$.  Now in all other
states $s_2$ such that $\chi_i(s_1) = \chi_i(s_2)$, we know that the
$Q^*$ values will be the same.  Hence, the same tie will exist between
$a_1$ and $a_2$, and hence, the optimal ordered policy must make the
same choice in all such states.  Hence, the optimal ordered policy for
node $i$ is an abstract policy.

Now let us turn to the recursive case at Max node $i$.  Make the
inductive assumption that the ordered recursively-optimal policy is
abstract within all descendant nodes and consider the locally optimal
policy at node $i$.  If $Y$ is a set of state variables that are
irrelevant to node $i$, Corollary \ref{corollary-optimal-is-abstract}
tells us that $Q^*(i,s_1,j) = Q^*(i,s_2,j)$ for all states $s_1$ and
$s_2$ such that $\chi_i(s_1) = \chi_i(s_2)$.  Similarly, if $Y$ is a
set of variables irrelevant to the result distribution of a particular
action $j$, then Lemma \ref{lemma-result-distribution-irrelevance}
tells us the same thing.  Hence, by the same ordering argument given
above, the ordered optimal policy at node $i$ must be abstract.  By
induction, this proves the lemma.  \qed

With this lemma, we have established that the combination of an MDP
$M$, an abstract MAXQ graph $H$, and an action ordering defines a
unique recursively-optimal ordered abstract policy.  We are now ready
to prove that MAXQ-Q will converge to this policy.

\begin{theorem}  
Let $M = \mtuple{S, A, P, R, P_0}$ be either an episodic MDP for which
all deterministic policies are proper or a discounted infinite horizon
MDP with discount factor $\gamma < 1$.  Let $H$ be an unabstracted MAXQ
graph defined over subtasks $\{M_0, \ldots, M_k\}$ with pseudo-reward
functions $\tilde{R}_i(s'|s,a)$.  Let $\chi(H)$ be a state-abstracted MAXQ
graph defined by applying state abstractions $\chi_i$ to each node $i$
of $H$ under the five conditions given above.  Let
$\pi_x(i,\chi_i(s))$ be an abstract ordered GLIE exploration policy at
each node $i$ and state $s$ whose decisions depend only on the
``relevant'' state variables at each node $i$.  Let $\pi^*_r$ be the
unique recursively-optimal hierarchical policy defined by $\pi_x$,
$M$, and $\tilde{R}$.  Then with probability 1, algorithm MAXQ-Q
applied to $\chi(H)$ converges to $\pi^*_r$ provided that the learning
rates $\alpha_t(i)$ satisfy Equation~(\ref{alpha-diverge-converge}) and
$|V_t(i,\chi_i(s))|$ and $|C_t(i,\chi_i(s),a)|$ are bounded for all $t$,
$i$, $\chi_i(s)$, and $a$.
\end{theorem}
\paragraph*{Proof:}
Rather than repeating the entire proof for MAXQ-Q, we will only
describe what must change under state abstraction.  The last two forms
of state abstraction refer to states whose values can be inferred from
the structure of the MAXQ graph, and therefore do not need to be
represented at all.  Since these values are not updated by MAXQ-Q, we
can ignore them.  We will now consider the first three forms of state
abstraction in turn. 

We begin by considering primitive leaf nodes. Let $a$ be a leaf node
and let $Y$ be a set of state variables that are Leaf Irrelevant for
$a$.  Let $s_1 = (x,y_1)$ and $s_2 = (x,y_2)$ be two states that
differ only in their values for $Y$.  Under Leaf Irrelevance, the
probability transitions $P(s_1'|s_1,a)$ and $P(s_2'|s_2,a)$ need not
be the same, but the expected reward of performing $a$ in both states
must be the same.  When MAXQ-Q visits an abstract state $x$, it does
not ``know'' the value of $y$, the part of the state that has been
abstracted away.  Nonetheless, it draws a sample according to
$P(s'|x,y,a)$, receives a reward $R(s'|x,y,a)$, and updates its
estimate of $V(a,x)$ (line 5 of MAXQ-Q).  Let $P_t(y)$ be the
probability that MAXQ-Q is visiting $(x,y)$ given that the
unabstracted part of the state is $x$.  Then Line 5 of MAXQ-Q is
computing a stochastic approximation to
\[\sum_{s',N,y} P_t(y) P_t(s',N|x,y,a) R(s'|x,y,a). \]
We can write this as
\[\sum_y P_t(y) \sum_{s',N}  P_t(s',N|x,y,a) R(s'|x,y,a).\]
According to Leaf Irrelevance, the inner sum has the same value for
all states $s$ such that $\chi(s) = x$.  Call this value $r_0(x)$.  This
gives
\[\sum_y P_t(y) r_0(x),\]
which is equal to $r_0(x)$ for any distribution $P_t(y)$.   Hence,
MAXQ-Q converges under Leaf Irrelevance abstractions. 

Now let us turn to the two forms of abstraction that apply to internal
nodes: Node Irrelevance and Result Distribution Irrelevance.  Consider
the SMDP defined at each node $i$ of the abstracted MAXQ graph at time
$t$ during MAXQ-Q.  This would be an ordinary SMDP with transition
probability function $P_t(x',N|x,a)$ and reward function $V_t(a,x) +
\tilde{R}_i(x')$ except that when MAXQ-Q draws samples of state
transitions, they are drawn according to the distribution
$P_t(s',N|s,a)$ over the original state space.  To prove the theorem,
we must show that drawing $(s',N)$ according to this second
distribution is equivalent to drawing $(x',N)$ according to the first
distribution.

For Max Node Irrelevance, we know that for all abstract policies
applied to node $i$ and its descendants, the transition probability
distribution factors as
\[
P(s',N|s,a) = P(y'|y,a) P(x',N|x,a).
\]
Because the exploration policy is an abstract policy, $P_t(s',N|s,a)$
factors in this way.  This means that the $X_i$ and $Y_i$ components of
the state are independent of each other, and hence, sampling from
$P_t(s',N|s,a)$ gives samples for $P_t(x',N|x,a)$.  Therefore, MAXQ-Q will
converge under Max Node Irrelevance abstractions. 

Finally, consider Result Distribution Irrelevance.  Let $j$ be a child
of node $i$, and suppose $Y_j$ is a set of state variables that are
irrelevant to the result distribution of $j$.  When the SMDP at node
$i$ wishes to draw a sample from $P_t(x',N|x,j)$, it does not ``know''
the current value of $y$, the irrelevant part of the current state.
However, this does not matter, because Result Distribution Irrelevance
means that for all possible values of $y$, $P_t(x',y',N|x,y,j)$ is the
same.  Hence, MAXQ-Q will converge under Result Distribution
Irrelevance abstractions.

In each of these three cases, MAXQ-Q will converge to a
locally-optimal ordered policy at node $i$ in the MAXQ graph.  By
Lemma \ref{lemma-ro-policy-is-abstract}, this can be extended to
produce a locally-optimal ordered policy for the unabstracted SMDP at
node $i$.  Hence, by induction, MAXQ-Q will converge to the unique
ordered recursively optimal policy $\pi^*_r$ defined by MAXQ-Q $H$,
MDP $M$, and ordered exploration policy $\pi_x$. \qed

\subsection{The Hierarchical Credit Assignment Problem}

There are still some situations where we would like to introduce state
abstractions but where the five properties described above do not
permit them.  Consider the following modification of the taxi
problem.  Suppose that the taxi has a fuel tank and that each time the
taxi moves one square, it costs one unit of fuel.  If the taxi runs
out of fuel before delivering the passenger to his or her destination,
it receives a reward of $-20$, and the trial ends.  Fortunately, there
is a filling station where the taxi can execute a {\sf Fillup}
action to fill the fuel tank.

To solve this modified problem using the MAXQ hierarchy, we can
introduce another subtask, {\sf Refuel}, which has the goal of moving
the taxi to the filling station and filling the tank.  {\sf MaxRefuel}
is a child of {\sf MaxRoot}, and it invokes ${\sf Navigate}(t)$ (with
$t$ bound to the location of the filling station) to move the taxi to
the filling station. 

The introduction of fuel and the possibility that we might run out of
fuel means that we must include the current amount of fuel as a
feature in representing every $C$ value (for internal nodes) and $V$
value (for leaf nodes).  This is unfortunate, because our intuition
tells us that the amount of fuel should have no influence on our
decisions inside the ${\sf Navigate}(t)$ subtask.  The amount of fuel
{\it should} be taken into account by the top-level Q nodes, which
must decide whether to go refuel, go pick up the passenger, or go
deliver the passenger.

Given this intuition, it is natural to try abstracting away the
``amount of remaining fuel'' within the ${\sf Navigate}(t)$ subtask.
However, this doesn't work, because when the taxi runs out of fuel and
a $-20$ reward is given, the {\sf QNorth}, {\sf QSouth}, {\sf QEast},
and {\sf QWest} nodes cannot ``explain'' why this reward was
received---that is, they have no consistent way of setting their $C$
tables to predict when this negative reward will occur.  Stated more
formally, the difficulty is that the Max Node Irrelevance condition is
not satisfied because the one-step reward function $R(s'|s,a)$ for
these actions depends on the amount of fuel. 

We call this the {\it hierarchical credit assignment problem}.  The
fundamental issue here is that in the MAXQ decomposition all
information about rewards is stored in the leaf nodes of the
hierarchy.  We would like to separate out the basic rewards received
for navigation (i.e., $-1$ for each action) from the reward received
for exhausting fuel ($-20$).  If we make the reward at the leaves only
depend on the location of the taxi, then the Max Node Irrelevance
condition will be satisfied.

One way to do this is to have the programmer manually decompose the
reward function and indicate which nodes in the hierarchy will
``receive'' each reward.  Let $R(s'|s,a) = \sum_i R(i,s'|s,a)$ be a
decomposition of the reward function, such that $R(i,s'|s,a)$
specifies that part of the reward that must be handled by Max node
$i$.  In the modified taxi problem, for example, we can decompose the
reward so that the leaf nodes receive all of the original penalties,
but the out-of-fuel rewards must be handled by {\sf MaxRoot}.  Lines
16 and 17 of the MAXQ-Q algorithm are easily modified to include
$R(i,s'|s,a)$.  

In most domains, we believe it will be easy for the designer of the
hierarchy to decompose the reward function.  It has been
straightforward in all of the problems we have studied.  However, an
interesting problem for future research is to develop an algorithm
that can solve the hierarchical credit assignment problem
autonomously. 

\section{Non-Hierarchical Execution of the MAXQ Hierarchy}

Up to this point in the paper, we have focused exclusively on
representing and learning hierarchical policies.  However, often the
optimal policy for a MDP is not a strictly hierarchical policy.
Kaelbling \citeyear{k-hrl:pr-93} first introduced the idea of deriving
a non-hierarchical policy from the value function of a hierarchical
policy.  In this section, we exploit the MAXQ decomposition to
generalize her ideas and apply them recursively at all levels of the
hierarchy.

The first method is based on the dynamic programming algorithm known
as policy iteration.  The policy iteration algorithm starts with an
initial policy $\pi^0$.  It then repeats the following two steps until
the policy converges.  In the {\it policy evaluation} step, it
computes the value function $V^{\pi_k}$ of the current policy $\pi_k$.
Then, in the {\it policy improvement step}, it computes a new
policy, $\pi_{k+1}$ according to the rule
\begin{equation}
\label{eq-pi}
\pi_{k+1}(s) := \argmax_a \sum_{s'} P(s'|s,a) [R(s'|s,a) + \gamma V^{\pi_k}(s')].
\end{equation}
Howard \citeyear{h-dpmp-60} proved that if $\pi_k$ is not an optimal
policy, then $\pi_{k+1}$ is guaranteed to be an improvement.  Note
that in order to apply this method, we need to know the transition
probability distribution $P(s'|s,a)$ and the reward function
$R(s'|s,a)$.

If we know $P(s'|s,a)$ and $R(s'|s,a)$, we can use the MAXQ
representation of the value function to perform one step of policy
iteration.  We start with a hierarchical policy $\pi$ and represent
its value function using the MAXQ hierarchy (e.g., $\pi$ could have
been learned via MAXQ-Q).  Then, we can perform one step of policy
improvement by applying Equation~(\ref{eq-pi}) using $V^{\pi}(0,s')$
(computed by the MAXQ hierarchy) to compute $V^{\pi}(s')$.

\begin{corollary}
\label{cor-pi}
Let $\pi^g(s) = \argmax_a \sum_{s'} P(s'|s,a) [R(s'|s,a) + \gamma
V^{\pi}(0,s)]$, where $V^{\pi}(0,s)$ is the value function computed by
the MAXQ hierarchy.  Then, if $\pi$ was not an optimal policy, $\pi^g$
is strictly better for at least one state in $S$.
\end{corollary}
\noindent {\bf Proof:}  This is a direct consequence of Howard's
policy improvement theorem. \qed
\vspace{.1in}

Unfortunately, we can't iterate this policy improvement process,
because the new policy, $\pi^g$ is very unlikely to be a hierarchical policy
(i.e., it is unlikely to be representable in terms of local policies
for each node of the MAXQ graph).  Nonetheless, one step of policy
improvement can give very significant improvements. 

This approach to non-hierarchical execution ignores the internal
structure of the MAXQ graph.  In effect, the MAXQ hierarchy is just
viewed as a kind of function approximator for representing
$V^{\pi}$---any other representation would give the same one-step
improved policy $\pi^g$.  

The second approach to non-hierarchical execution borrows an idea from
$Q$ learning.  One of the great beauties of the $Q$ representation for
value functions is that we can compute one step of policy improvement
without knowing $P(s'|s,a)$, simply by taking the new policy to be
$\pi^g(s) := \argmax_a Q(s,a)$.  This gives us the same one-step
greedy policy as we computed above using one-step lookahead.  With the
MAXQ decomposition, we can perform these policy improvement steps {\it
at all levels of the hierarchy}. 

We have already defined the function that we need.  In
Table~\ref{tab-eval-greedy} we presented the function EvaluateMaxNode,
which, given the current state $s$, conducts a search along all paths
from a given Max node $i$ to the leaves of the MAXQ graph and finds
the path with the best value (i.e., with the maximum sum of $C$ values
along the path, plus the $V$ value at the leaf).  In addition,
EvaluateMaxNode returns the primitive action $a$ at the end of this
best path.  This action $a$ would be the first primitive action to be
executed if the learned hierarchical policy were executed starting in
the current state $s$.  Our second method for non-hierarchical
execution of the MAXQ graph is to call EvaluateMaxNode in each state,
and execute the primitive action $a$ that is returned.  The
pseudo-code is shown in Table~\ref{tab-exec-greedy}.

\begin{table}
\caption{The procedure for executing the one-step greedy policy.}
\label{tab-exec-greedy}
\vspace*{0.1in}
\hrule
{\footnotesize
\begin{center}
\begin{quote}
\begin{tabbing}
xxx\=xxxx\=xxxx\=\kill
 \>procedure ExecuteHGPolicy(s)\\[.1in]
1 \>\>repeat\\
2 \>\>\>  Let \tuple{V(0,s), a} := EvaluateMaxNode$(0, s)$\\
3 \>\>\>  execute primitive action $a$\\
4 \>\>\>  Let $s$ be the resulting state\\
  \> {\bf end} // ExecuteHGPolicy
\end{tabbing}
\end{quote}
\end{center}
}
\hrule
\end{table}

We will call the policy computed by ExecuteHGPolicy the {\it
hierarchical greedy policy}, and denote it $\pi^{hg*}$, where the
superscript * indicates that we are computing the greedy action at
each time step.  The following theorem shows that this can give a
better policy than the original, hierarchical policy.  

\begin{theorem} \label{theorem-hg}
Let $G$ be a MAXQ graph representing the value function of
hierarchical policy $\pi$ (i.e., in terms of $C^{\pi}(i,s,j)$,
computed for all $i,$ $s$, and $j$).  Let $V^{hg}(0,s)$ be the value
computed by ExecuteHGPolicy, and let $\pi^{hg*}$ be the resulting
policy.  Define $V^{hg*}$ to be the value function of $\pi^{hg*}$.
Then for all states $s$, it is the case that
\begin{equation}
\label{eq-value-improvement}
V^{\pi}(s) \leq V^{hg}(0,s) \leq V^{hg*}(s).
\end{equation}
\end{theorem}
\noindent {\bf Proof:} (sketch) The left inequality in
Equation~(\ref{eq-value-improvement}) is satisfied by construction by
line 7 of EvaluateMaxNode.  To see this, consider that the original
hierarchical policy, $\pi$, can be viewed as choosing a ``path''
through the MAXQ graph running from the root to one of the leaf nodes,
and $V^{\pi}(0,s)$ is the sum of the $C^{\pi}$ values along this
chosen path (plus the $V^{\pi}$ value at the leaf node).  In contrast,
EvaluateMaxNode performs a traversal of {\it all} paths through the
MAXQ graph and finds the {\it best} path, that is, the path with the
largest sum of $C^{\pi}$ (and leaf $V^{\pi}$) values.  Hence,
$V^{hg}(0,s)$ must be at least as large as $V^{\pi}(0,s)$.

To establish the right inequality, note that by construction
$V^{hg}(0,s)$ is the value function of a policy, call it $\pi^{hg}$,
that chooses one action greedily at each level of the MAXQ graph
(recursively), and then follows $\pi$ thereafter.  This is a
consequence of the fact that line 7 of EvaluateMaxNode has $C^{\pi}$
on its right-hand side, and $C^{\pi}$ represents the cost of
``completing'' each subroutine by following $\pi$, not by following
some other, greedier, policy.  (In Table~\ref{tab-eval-greedy},
$C^{\pi}$ is written as $C_t$.)  However, when we execute
ExecuteHGPolicy (and hence, execute $\pi^{hg*}$), we have an
opportunity to improve upon $\pi$ and $\pi^{hg}$ at each time step.
Hence, $V^{hg}(0,s)$ is an underestimate of the actual value of
$\pi^{hg*}$. \qed
\vspace{.1in}

Note that this theorem only works in one direction.  It says that if
we can find a state where $V^{hg}(0,s) > V^{\pi}(s)$, then the greedy
policy, $\pi^{hg*}$, will be strictly better than $\pi$.  However, it
could be that $\pi$ is not an optimal policy and yet the structure of
the MAXQ graph prevents us from considering an action (either
primitive or composite) that would improve $\pi$.  Hence, unlike the
policy improvement theorem of Howard, we do not have a guarantee that
if $\pi$ is suboptimal, then the hierarchically greedy policy is a
strict improvement.

In contrast, if we perform one-step policy improvement as discussed at
the start of this section, Corollary~\ref{cor-pi} guarantees that we
will improve the policy.  So we can see that in general, neither of
these two methods for non-hierarchical execution dominates the other.
Nonetheless, the first method only operates at the level of individual
primitive actions, so it is not able to produce very large
improvements in the policy.  In contrast, the hierarchical greedy
method can obtain very large improvements in the policy by changing
which actions (i.e., subroutines) are chosen near the root of the
hierarchy.  Hence, in general, hierarchical greedy execution is
probably the better method.  (Of course, the value functions of both
methods could be computed, and the one with the better estimated value
could be executed.)

Sutton, Singh, Precup and Ravindran \citeyear{sspr-istaa-99} have
simultaneously developed a closely-related method for non-hierarchical
execution of macros.  Their method is equivalent to ExecuteHGPolicy
for the special case where the MAXQ hierarchy has only one level of
subtasks.  The interesting aspect of ExecuteHGPolicy is that it
permits greedy improvements at all levels of the tree to influence
which action is chosen.

Some care must be taken in applying Theorem~\ref{theorem-hg} to a MAXQ
hierarchy whose $C$ values have been learned via MAXQ-Q.  Being an
online algorithm, MAXQ-Q will not have correctly learned the values of
{\it all} states at all nodes of the MAXQ graph.  For example, in the
taxi problem, the value of $C({\sf Put}, s, {\sf QPutdown})$ will not
have been learned very well except at the four special locations.
This is because the {\sf Put} subtask cannot be executed until the
passenger is in the taxi, and this usually means that a {\sf Get} has
just been completed, so the taxi is at the passenger's source
location.  During exploration, both children of {\sf Put} will be
tried in such states.  The {\sf PutDown} will usually fail, whereas
the {\sf Navigate} will eventually succeed (perhaps after lengthy
exploration) and take the taxi to the destination location.  Now
because of all states updating, the values for $C({\sf Put}, s, {\sf
Navigate}(t))$ will have been learned at all of the states, but the
$C$ values for the {\sf Putdown} action will not.  Hence, if we train
the MAXQ representation using hierarchical execution (as in MAXQ-Q),
and then switch to hierarchically-greedy execution, the results will
be quite bad.  In particular, we need to introduce
hierarchically-greedy execution early enough so that the exploration
policy is still actively exploring.  (In theory, a GLIE exploration
policy never ceases to explore, but in practice, we want to find a
good policy quickly, not just asymptotically).

Of course an alternative would be to use hierarchically-greedy
execution from the very beginning of learning.  However, remember that
the higher nodes in the MAXQ hierarchy need to obtain samples of
$P(s',N|s,a)$ for each child action $a$.  If the hierarchical greedy
execution interrupts child $a$ before it has reached a terminal state,
then these samples cannot be obtained.  Hence, it is important to
begin with purely hierarchical execution during training, and make a
transition to greedy execution at some point. 

The approach we have taken is to implement MAXQ-Q in such a way that
we can specify a number of primitive actions $L$ that can be taken
hierarchically before the hierarchical execution is ``interrupted''
and control returns to the top level (where a new action can be chosen
greedily).  We start with $L$ set very large, so that execution is
completely hierarchical---when a child action is invoked, we are
committed to execute that action until it terminates.  However,
gradually, we reduce $L$ until it becomes 1, at which point we have
hierarchical greedy execution.  We time this so that it reaches 1 at
about the same time our Boltzmann exploration cools to a temperature
of 0.1 (which is where exploration effectively has halted).  As the
experimental results will show, this generally gives excellent
results with very little added exploration cost. 

\section{Experimental Evaluation of the MAXQ Method}

We have performed a series of experiments with the MAXQ method with
three goals in mind: (a) to understand the expressive power of the
value function decomposition, (b) to characterize the behavior of the
MAXQ-Q learning algorithm, and (c) to assess the relative importance
of temporal abstraction, state abstraction, and non-hierarchical
execution.  In this section, we describe these experiments and present
the results.

\subsection{The Fickle Taxi Task}

Our first experiments were performed on a modified version of the taxi
task.  This version incorporates two changes to the task described in
Section~\ref{sec-taxi}.  First, each of the four navigation actions is
noisy, so that with probability 0.8 it moves in the intended
direction, but with probability 0.1 it instead moves to the right (of
the intended direction) and with probability 0.1 it moves to the left.
The second change is that after the taxi has picked up the passenger
and moved one square away from the passenger's source location, the
passenger changes his or her destination location with probability
0.3.  The purpose of this change is to create a situation where the
optimal policy is not a hierarchical policy so that the effectiveness
of non-hierarchical execution can be measured.  

We compared four different configurations of the learning algorithm:
(a) flat Q learning, (b) MAXQ-Q learning without any form of state
abstraction, (c) MAXQ-Q learning with state abstraction, and (d)
MAXQ-Q learning with state abstraction and greedy execution.  These
configurations are controlled by many parameters.  These include the
following: (a) the initial values of the Q and C functions, (b) the
learning rate (we employed a fixed learning rate), (c) the cooling
schedule for Boltzmann exploration (the GLIE policy that we employed),
and (d) for non-hierarchical execution, the schedule for decreasing
$L$, the number of steps of consecutive hierarchical execution.  We
optimized these settings separately for each configuration with the
goal of matching or exceeding (with as few primitive actions as
possible) the best policy that we could code by hand.  For Boltzmann
exploration, we established an initial temperature and then a cooling
rate.  A separate temperature is maintained for each Max node in the
MAXQ graph, and its temperature is reduced by multiplying by the
cooling rate each time that subtask terminates in a goal state. 

The following parameters were chosen.  For flat Q learning: initial Q
values of 0.123, learning rate 0.25, and Boltzmann exploration with an
initial temperature of 50 and a cooling rate of 0.9879.  (We use
initial values that end in .123 as a ``signature'' to aid debugging.)

For MAXQ-Q learning without state abstraction, we used initial values
of 0.123, a learning rate of 0.50, and Boltzmann exploration with an
initial temperature of 50 and cooling rates of .9996 at MaxRoot and
MaxPut, 0.9939 at MaxGet, and 0.9879 at MaxNavigate. 

For MAXQ-Q learning with state abstraction, we used initial values of
0.123, a learning rate of 0.25, and Boltzmann exploration with an
initial temperature of 50 and cooling rates of 0.9074 at MaxRoot,
0.9526 at MaxPut, 0.9526 at MaxGet, and 0.9879 at MaxNavigate. 

For MAXQ-Q learning with non-hierarchical execution, we used the same
settings as with state abstraction.  In addition, we initialized $L$
to 500 and decreased it by 10 with each trial until it reached 1.  So
after 50 trials, execution was completely greedy.

\begin{figure}
\centerps{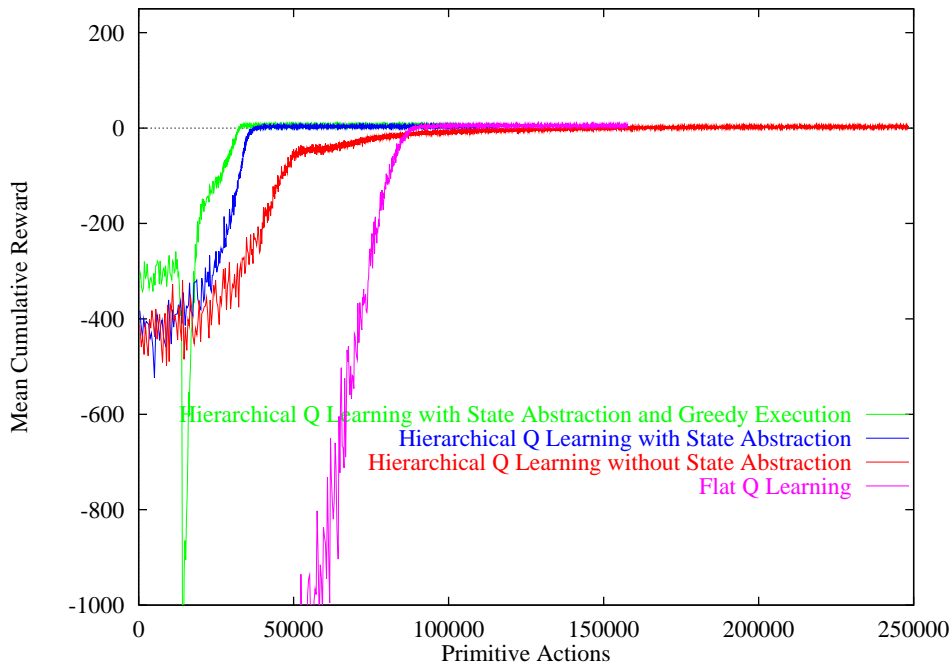}
\caption{Comparison of performance of hierarchical Q learning with
flat Q learning, with and without state abstractions, and with and
without greedy evaluation.}
\label{fig-delta-taxi}
\end{figure}

Figure~\ref{fig-delta-taxi} shows the averaged results of 100 training
trials.  The first thing to note is that all forms of MAXQ learning
have better initial performance than flat Q learning.  This is because
of the constraints introduced by the MAXQ hierarchy.  For example, 
while the agent is executing a {\sf Navigate} subtask, it will never
attempt to pickup or putdown the passenger.  Similarly, it will never
attempt to putdown the passenger until it has first picked up the
passenger (and vice versa). 

The second thing to notice is that without state abstractions, MAXQ-Q
learning actually takes longer to converge, so that the Flat Q curve
crosses the MAXQ/no abstraction curve.  This shows that without state
abstraction, the cost of learning the huge number of parameters in the
MAXQ representation is not really worth the benefits.  

\begin{figure}
\centerps{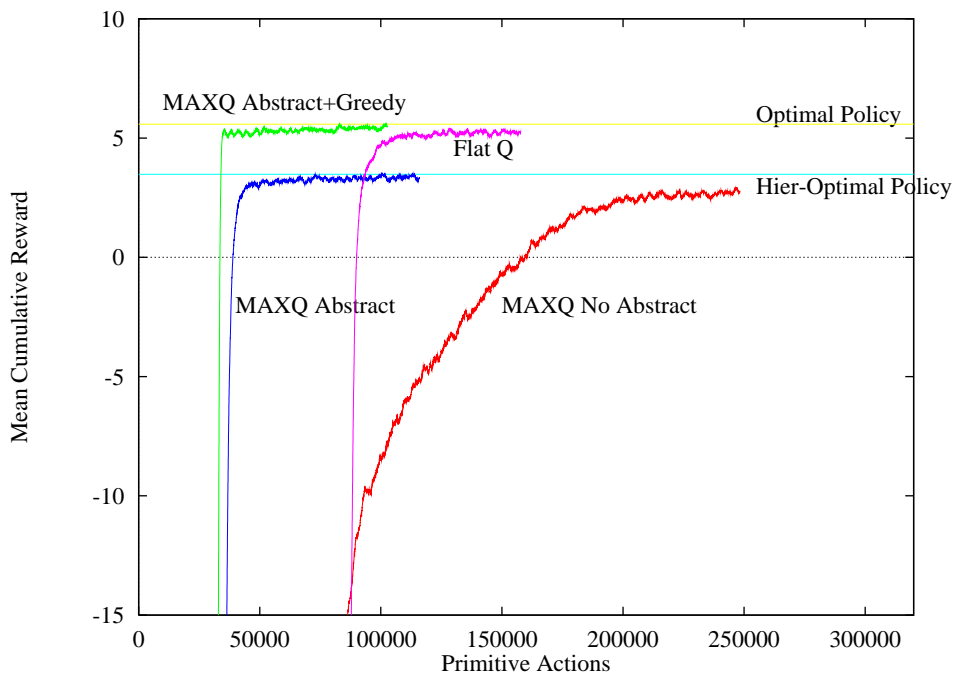}
\caption{Close-up view of the previous figure.  This figure also shows
two horizontal lines indicating optimal performance and hierarchically
optimal performance in this domain.  To make this figure more
readable, we have applied a 100-step moving average to the data
points.}
\label{fig-delta-taxi-small}
\end{figure}

The third thing to notice is that with state abstractions, MAXQ-Q
converges very quickly to a hierarchically optimal policy.  This can
be seen more clearly in Figure~\ref{fig-delta-taxi-small}, which
focuses on the range of reward values in the neighborhood of the
optimal policy.  Here we can see that MAXQ with abstractions attains
the hierarchically optimal policy after approximately 40,000 steps,
whereas flat Q learning requires roughly twice as long to reach the
same level.  However, flat Q learning, of course, can continue onward
and reach optimal performance, whereas with the MAXQ hierarchy, the
best hierarchical policy is slow to respond to the ``fickle'' behavior
of the passenger when he/she changes the destination. 

The last thing to notice is that with greedy execution, the MAXQ
policy is also able to attain optimal performance.  But as the
execution becomes ``more greedy'', there is a drop in performance,
because MAXQ-Q must learn $C$ values in new regions of the state space
that were not visited by the recursively optimal policy.  Despite this
drop in performance, greedy MAXQ-Q recovers rapidly and reaches
hierarchically optimal performance faster than purely-hierarchical
MAXQ-Q learning.  Hence, there is no added cost---in terms of
exploration---for introducing greedy execution. 

This experiment presents evidence in favor of three claims:  first, that
hierarchical reinforcement learning can be much faster than flat Q
learning;  second, that state abstraction is required by MAXQ for good
performance; and third, that non-hierarchical execution can produce
significant improvements in performance with little or no added
exploration cost. 

\subsection{Kaelbling's HDG Method}

\begin{figure}
{\epsfxsize=3in
\centerps{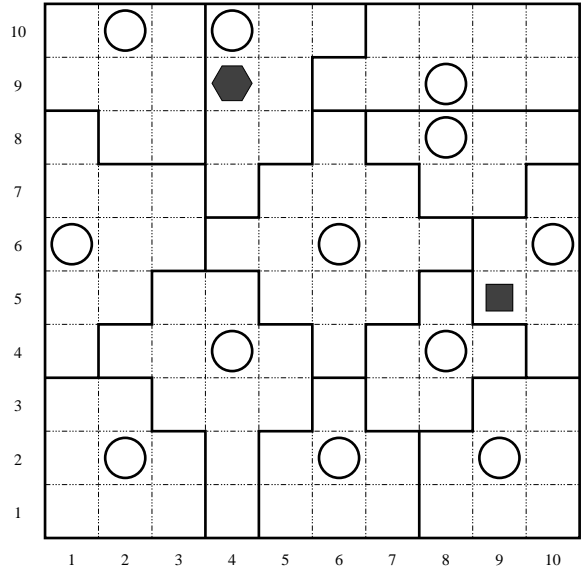}
}
\caption{Kaelbling's 10-by-10 navigation task.  Each circled state is
a landmark state, and the heavy lines show the boundaries of the
Voronoi cells.  In each episode, a start state and a goal state are
chosen at random.  In this figure, the start state is shown by the
shaded hexagon, and the goal state is shown by the shaded square. }
\label{fig-kaelbling}
\end{figure}

\begin{figure}
{\epsfxsize=468pt
\centerps{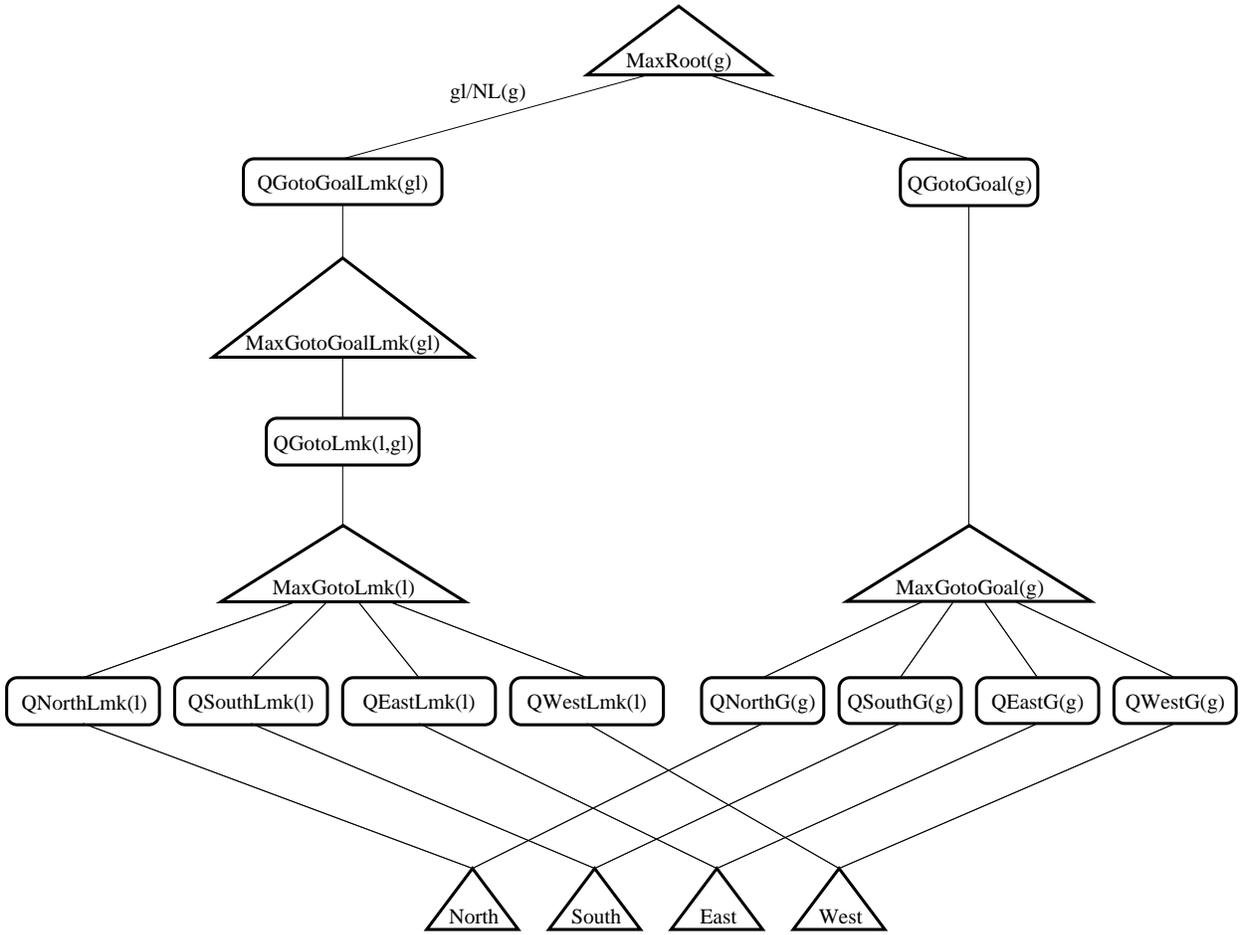}
}
\caption{A MAXQ graph for the HDG navigation task.}
\label{fig-hdg}
\end{figure}

The second task that we will consider is a simple maze task introduced
by Leslie Kaelbling \citeyear{k-hrl:pr-93} and shown in
Figure~\ref{fig-hdg}.  In each trial of this task, the agent starts in
a randomly-chosen state and must move to a randomly-chosen goal state
using the usual {\sf North}, {\sf South}, {\sf East}, and {\sf West}
operators (we employed deterministic operators).  There is a small
cost for each move, and the agent must maximize the undiscounted sum
of these costs.

Because the goal state can be in any of 100 different locations, there
are actually 100 different MDPs.  Kaelbling's HDG method starts by
choosing an arbitrary set of landmark states and defining a Voronoi
partition of the state space based on the Manhattan distances to these
landmarks (i.e., two states belong to the same Voronoi cell iff they
have the same nearest landmark).  The method then defines one subtask
for each landmark $l$.  The subtask is to move from any state in the
current Voronoi cell {\it or in any neighboring Voronoi cell} to the
landmark $l$.  Optimal policies for these subtasks are then computed.

Once HDG has the policies for these subtasks, it can solve the
abstract Markov Decision Problem of moving from each landmark state to
any other landmark state using the subtask solutions as macro actions
(subroutines).  So it computes a value function for this MDP.

Finally, for each possible destination location $g$ within a Voronoi
cell for landmark $l$, the HDG method computes the optimal policy of
getting from $l$ to $g$. 

By combining these subtasks, the HDG method can construct a good
approximation to the optimal policy as follows.  In addition to the
value functions discussed above, the agent maintains two other
functions: $NL(s)$, the name of the landmark nearest to state $s$, and
$N(l)$, a list of the landmarks that are in the cells that are
immediate neighbors of cell $l$.  By combining these, the agent can
build a list for each state $s$ of the current landmark and the
landmarks of the neighboring cells.  For each such landmark, the agent
computes the sum of three terms:
\begin{enumerate}
\item[(t1)] the expected cost of reaching that landmark, 
\item[(t2)] the expected cost of moving from that landmark to the
landmark in the goal cell, and
\item[(t3)] the expected cost of moving from the goal-cell
landmark to the goal state.
\end{enumerate}
Note that while terms (t1) and (t3) can be exact estimates, term (t2)
is computed using the landmark subtasks as subroutines.  This means
that the corresponding path must pass through the intermediate
landmark states rather than going directly to the goal landmark.
Hence, term (t2) is typically an overestimate of the required
distance.  (Also note that (t3) is the same for all choices of the
intermediate landmarks, so it does not need to be explicitly included
in the computation.)

Given this information, the agent then chooses to move toward the best
of the landmarks (unless the agent is already in the goal Voronoi
cell, in which case the agent moves toward the goal state).  For
example, in Figure~\ref{fig-kaelbling}, term (t1) is the cost of
reaching the landmark in row 7, column 4, which is 4.  Term (t2) is the
cost of getting from row 7, column 4 to the landmark at row 1 column 4
(by going from one landmark to another).  In this case, the best
landmark-to-landmark path is from row 7, column 1 to row 5 column 6,
and then to row 1 column 4.  Hence, term (t2) is 12.  Term (t3) is the
cost of getting from row 1 column 4 to the goal, which is 2.  The sum
of these is 4 + 12 + 2 = 18.  For comparison, the optimal path has
length 10.

In Kaelbling's experiments, she employed a variation of Q learning to
learn terms (t1) and (t3), and she computed (t2) at regular intervals
via the Floyd-Warshall all-sources shortest paths algorithm.

Figure~\ref{fig-hdg} shows a MAXQ approach to solving this problem.
The overall task {\sf Root}, takes one argument $g$, which specifies
the goal cell.  There are three subtasks:
\begin{itemize}

\item ${\sf GotoGoalLmk}$, go to the landmark nearest to the goal
location.  The termination for the predicate is true if the agent
reaches the landmark nearest to the goal.  The goal predicate is the
same as the termination predicate. 

\item ${\sf GotoLmk}(l)$, go to landmark $l$.  The termination
predicate for this is true if either (a) the agent reaches landmark
$l$ or (b) the agent is outside of the region defined by the Voronoi
cell for $l$ and the neighboring Voronoi cells, $N(l)$.  The goal
predicate for this subtask is true only for condition (a). 

\item ${\sf GotoGoal}(g)$, go to the goal location $g$.  The termination
predicate for this subtask is true if either the agent is in the goal
location or the agent is outside of the Voronoi cell $NL(g)$ that contains
$g$.  The goal predicate for this subtask is true if the agent is in
the goal location. 
\end{itemize}

The MAXQ decomposition is essentially the same as Kaelbling's method,
but somewhat redundant.  Consider a state where the agent is not
inside the same Voronoi cell as the goal $g$.  In such states, HDG
decomposes the value function into three terms (t1), (t2), and (t3).
Similarly, MAXQ also decomposes it into these same three terms:

\begin{itemize}
\item $V(GotoLmk(l), s, a)$ the cost of getting to landmark $l$.
(Actually the sum of $V(a,s)$ and $C(GotoLmk(l), s, a)$.)

\item $C(GotoGoalLmk(gl), s, MaxGotoLmk(l))$ the cost of getting from
landmark $l$ to the landmark $gl$ nearest the goal. 

\item $C(Root, s, GotoGoalLmk(gl))$ the cost of getting to the goal
location after reaching $gl$. 
\end{itemize}

When the agent is inside the goal Voronoi cell, then again HDG and
MAXQ store essentially the same information.  HDG stores $Q({\sf
GotoGoal}(g), s, a)$, while MAXQ breaks this into two terms: $C({\sf
GotoGoal}(g), s, a)$ and $V(a,s)$ and then sums these two quantities
to compute the $Q$ value.  

Note that this MAXQ decomposition stores some information
twice---specifically, the cost of getting from the goal landmark $gl$
to the goal is stored both as $C(Root, s, GotoGoalLmk(gl))$ and as
$C({\sf GotoGoal}(g), s, a) + V(a,s)$.

Let us compare the amount of memory required by flat Q learning, HDG,
and MAXQ.   There are 100 locations, 4 possible actions, and 100
possible goal states, so flat $Q$ learning must store 40,000 values.

To compute quantity (t1), HDG must store 4 Q values (for the four
actions) for each state $s$ with respect to its own landmark and the
landmarks in $N(NL(s))$.  This gives a total of 2,028 values that must
be stored.

To compute quantity (t2), HDG must store, for each landmark,
information on the shortest path to every other landmark.  There are
12 landmarks.  Consider the landmark at row 6, column 1.  It has 5
neighboring landmarks which constitute the five macro actions that
the agent can perform to move to another landmark.  The nearest
landmark to the goal cell could be any of the other 11 landmarks, so
this gives a total of 55 Q values that must be stored.  Similar
computations for all 12 landmarks give a total of 506 values that must
be stored.

Finally, to compute quantity (t3), HDG must store information, for each
square inside each Voronoi cell, about how to get to each of the other
squares inside the same Voronoi cell.  This requires 3,536 values.

Hence, the grand total for HDG is 6,070, which is a huge savings over
flat Q learning.

Now let's consider the MAXQ hierarchy with and without state
abstractions.  

\begin{itemize}
\item $V(a,s)$: This is the expected reward of each primitive action
in each state.  There are 100 states and 4 primitive actions, so this
requires 400 values.  However, because the reward is constant $(-1)$,
we can apply Leaf Irrelevance to store only a single value.

\item $C({\sf GotoLmk}(l), s, a)$, where $a$ is one of the four
primitive actions.  This requires the same amount of space as (t1) in
Kaelbling's representation---indeed, combined with $V(a,a)$, this
represents exactly the same information as (t1).  It requires 2,028
values.  No state abstractions can be applied. 

\item $C({\sf GotoGoalLmk}(gl), s, {\sf GotoLmk}(l))$: This is the
cost of completing the {\sf GotoGoalLmk} task after going to landmark
$l$.  If the primitive actions are deterministic, then ${\sf
GotoLmk}(l)$ will always terminate at location $l$, and hence, we only
need to store this for each pair of $l$ and $gl$.  This is exactly the
same as Kaelbling's quantity (t2), which requires 506 values.
However, if the primitive actions are stochastic---as they were in
Kaelbling's original paper---then we must store this value for each
possible terminal state of each ${\sf GotoLmk}$ action.  Each of these
actions could terminate at its target landmark $l$ or in one of the
states bordering the set of Voronoi cells that are the neighbors of
the cell for $l$.  This requires 6,600 values.  When Kaelbling stores
values only for (t2), she is effectively making the assumption that
${\sf GotoLmk}(l)$ will never fail to reach landmark $l$.  This is an
approximation which we can introduce into the MAXQ representation by
our choice of state abstraction at this node.

\item $C({\sf GotoGoal}, s, a)$: This is the cost of completing the
{\sf GotoGoal} task after making one of the primitive actions $a$.
This is the same as quantity (t3) in the HDG representation, and it
requires the same amoount of space: 3,536 values.

\item $C({\sf Root}, s, {\sf GotoGoalLmk})$:  This is the cost of
reaching the goal once we have reached the landmark nearest the goal.
MAXQ must represent this for all combinations of goal landmarks and
goals.  This requires 100 values.  Note that these values are the same
as the values of $C({\sf GotoGoal}(g), s, a) + V(a,s)$ for each of the
primitive actions.  This means that the MAXQ representation stores
this information twice, whereas the HDG representation only stores it
once (as term (t3)). 

\item $C({\sf Root}, s, {\sf GotoGoal})$.  This is the cost of
completing the {\sf Root} task after we have executed the {\sf
GotoGoal} task.  If the primitive action are deterministic, this is
always zero, because {\sf GotoGoal} will have reached the goal.
Hence, we can apply the Termination condition and not store any values
at all.  However, if the primitive actions are stochastic, then we
must store this value for each possible state that borders the Voronoi
cell that contains the goal.  This requires 96 different values.
Again, in Kaelbling's HDG representation of the value function, she is
ignoring the probability that {\sf GotoGoal} will terminate in a
non-goal state.  Because MAXQ is an exact representation of the value
function, it does not ignore this possibility.  If we (incorrectly)
apply the Termination condition in this case, the MAXQ representation
becomes a function approximation.
\end{itemize}

In the stochastic case, without state abstractions, the MAXQ
representation requires 12,760 values.  With safe state abstractions, it
requires 12,361 values.  With the approximations employed by Kaelbling
(or equivalently, if the primitive actions are deterministic), the
MAXQ representation with state abstractions requires 6,171 values. 
These numbers are summarized in Table~\ref{tab-hdg}.  We can see that,
with the unsafe state abstractions, the MAXQ representation requires
only slightly more space than the HDG representation (because of the
redundancy in storing $C({\sf Root}, s, {\sf GotoGoalLmk})$. 

\begin{table}
\caption{Comparison of the number of values that must be stored to
represent the value function using the HDG and MAXQ methods.}
\label{tab-hdg}
\begin{center}
\begin{tabular}{llrrrr}\hline
HDG & MAXQ                                     &
\multicolumn{1}{c}{HDG} & \multicolumn{1}{c}{MAXQ} &
\multicolumn{1}{c}{MAXQ} & \multicolumn{1}{c}{MAXQ} \\ 
item& item                                     & values & no abs & safe abs & unsafe abs\\ \hline
    &$V(a,s)$                                  &      0 &    400 &      1   &      1 \\
(t1)&$C({\sf GotoLmk}(l),s,a)$                 &  2,028 &  2,028 &  2,028   &  2,028 \\
(t2)&$C({\sf GotoGoalLmk},s,{\sf GotoLmk}(l))$ &    506 &  6,600 &  6,600   &    506 \\
(t3)&$C({\sf GotoGoal}(g),s,a)$                &  3,536 &  3,536 &  3,536   &  3,536 \\
    &$C({\sf Root}, s, {\sf GotoGoalLmk})$     &      0 &    100 &    100   &    100 \\
    &$C({\sf Root}, s, {\sf GotoGoal})$        &      0 &     96 &     96   &      0 \\ \hline
\multicolumn{2}{c}{Total Number of Values Required} & 6,070 & 12,760 & 12,361 & 6,171 \\ \hline
\end{tabular}
\end{center}
\end{table}

This example shows that for the HDG task, we can start with the
fully-general formulation provided by MAXQ and impose assumptions to
obtain a method that is similar to HDG.  The MAXQ formulation
guarantees that the value function of the hierarchical policy will be
represented exactly.  The assumptions will introduce approximations
into the value function representation.  This might be useful as a
general design methodology for building application-specific
hierarchical representations.  Our long-term goal is to develop such
methods so that each new application does not require inventing a new
set of techniques.  Instead, off-the-shelf tools (e.g., based on MAXQ)
could be specialized by imposing assumptions and state abstractions to
produce more efficient special-purpose systems.  

One of the most important contributions of the HDG method was that it
introduced a form of non-hierarchical execution.  As soon as the agent
crosses from one Voronoi cell into another, the current subtask is
``interrupted'', and the agent recomputes the ``current target
landmark''.  The effect of this is that (until it reaches the goal
Voronoi cell), the agent is always aiming for a landmark outside of
its current Voronoi cell.  Hence, although the agent ``aims for'' a
sequence of landmark states, it typically does not visit many of these
states on its way to the goal.  The states just provide a convenient
set of intermediate targets.  By taking these ``shortcuts'', HDG
compensates for the fact that, in general, it has overestimated the
cost of getting to the goal, because its computed value function is
based on a policy where the agent goes from one landmark to another.

The same effect is obtained by hierarchical greedy execution of the
MAXQ graph (which was directly inspired by the HDG method).  Note that
by storing the $NL$ (nearest landmark) function, Kaelbing's HDG method
can detect very efficiently when the current subtask should be
interrupted.  This technique only works for navigation problems in a
space with a distance metric.  In contrast, ExecuteHGPolicy performs a
kind of ``polling'', where it checks after each primitive action
whether it should interrupt the current subroutine and invoke a new
one.  An important goal for future research on MAXQ is to find a
general purpose mechanism for avoiding unnecessary ``polling''---that
is, a mechanism that can discover efficiently-evaluable interrupt
conditions.

\begin{figure}
\centerps{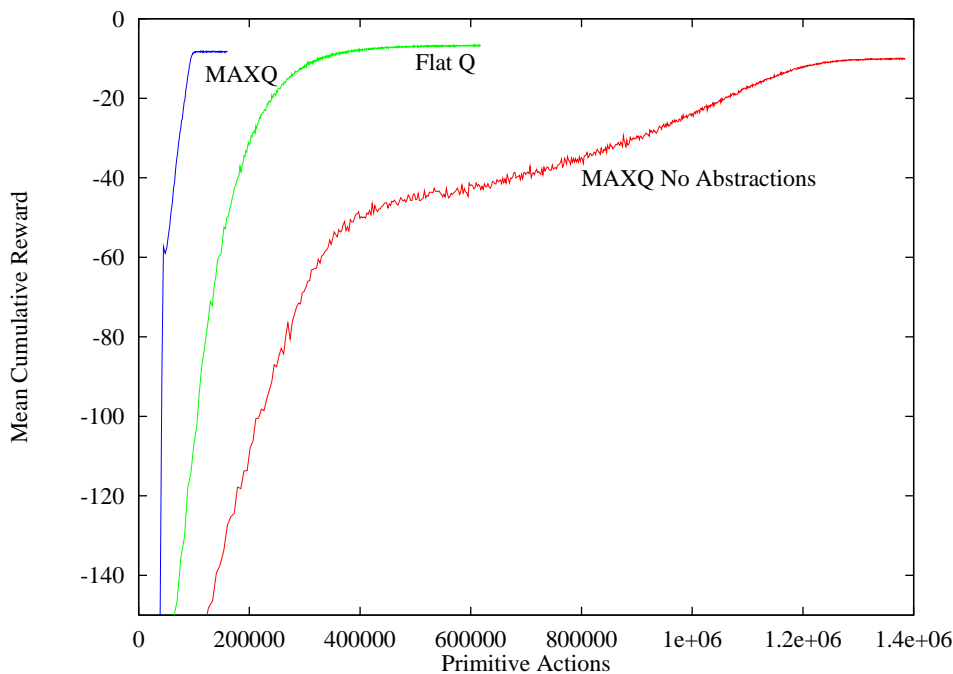}
\caption{Comparison of Flat Q learning with MAXQ-Q learning with and
without state abstraction. (Average of 100 runs.)}
\label{fig-hdg-big}
\end{figure}

\begin{figure}
\centerps{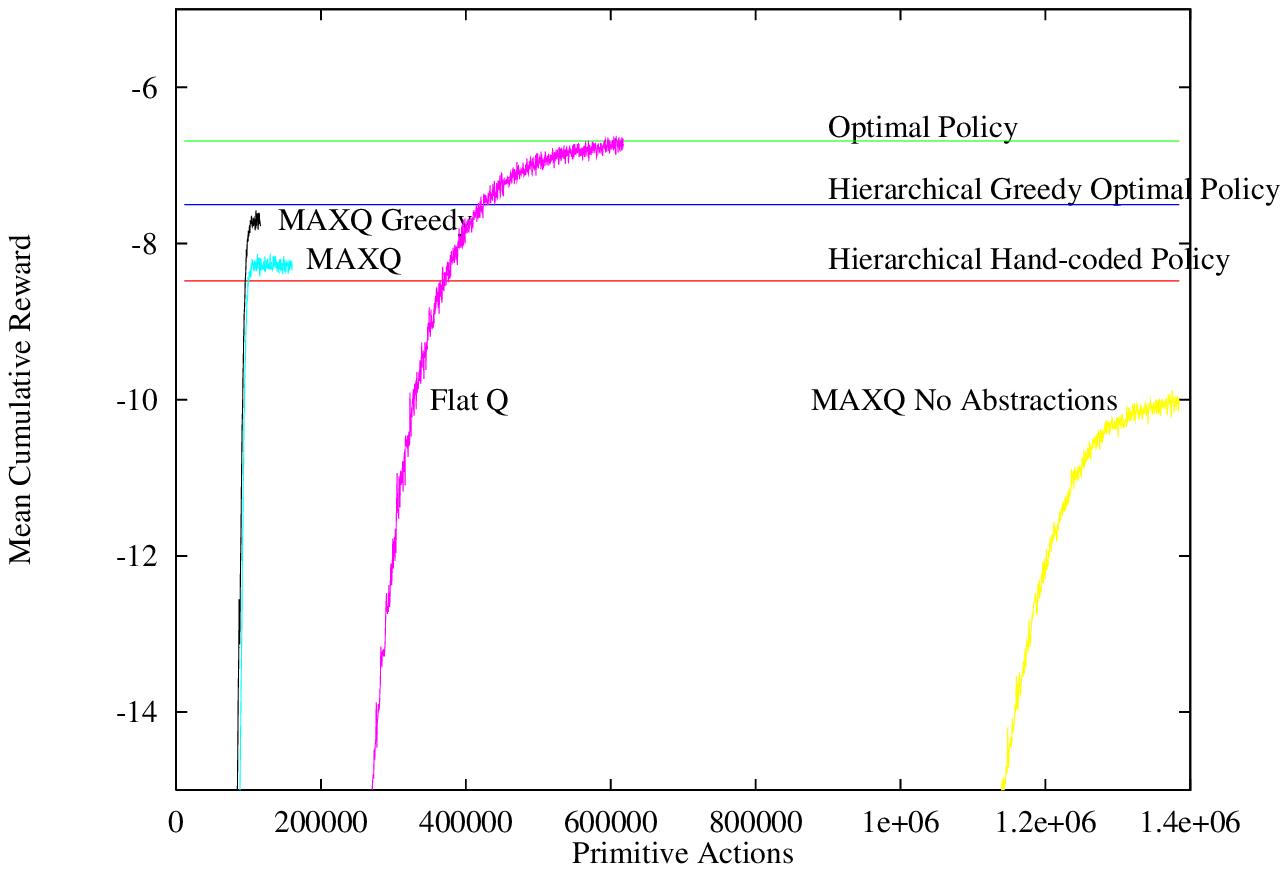}
\caption{Expanded view comparing Flat Q learning with MAXQ-Q learning
with and without state abstraction and with and without hierarchical
greedy execution. (Average of 100 runs.)}
\label{fig-hdg-small}
\end{figure}

Figure~\ref{fig-hdg-big} shows the results of our experiments with HDG
using the MAXQ-Q learning algorithm.  We employed the following
parameters: for Flat Q learning, initial values of 0.123, a learning
rate of 1.0, initial temperature of 50, and cooling rate of .9074; for
MAXQ-Q without state abstractions: initial values of $-25.123$, learning
rate of 1.0, initial temperature of 50, and cooling rates of .9074 for
{\sf MaxRoot}, .9999 for {\sf MaxGotoGoalLmk}, .9074 for {\sf
MaxGotoGoal}, and .9526 for {\sf MaxGotoLmk}; for MAXQ-Q with state
abstractions: initial values of $-20.123$, learning rate of 1.0, initial
temperature of 50, and cooling rates of .9760 for {\sf MaxRoot}, .9969
for {\sf MaxGotoGoal}, .9984 for {\sf MaxGotoGoalLmk}, and .9969 for
{\sf MaxGotoLmk}.  Hierarchical greedy execution was introduced by
starting with 3000 primitive actions per trial, and reducing this
every trial by 2 actions, so that after 1500 trials, execution is
completely greedy. 

The figure confirms the observations made in our experiments with the
Fickle Taxi task.  Without state abstractions, MAXQ-Q converges much
more slowly than flat Q learning.  With state abstractions, it
converges roughly three times as fast.  Figure~\ref{fig-hdg-small}
shows a close-up view of Figure~\ref{fig-hdg-big} that allows us to
compare the differences in the final levels of performance of the
methods.  Here, we can see that MAXQ-Q with no state abstractions was
not able to reach the quality of our hand-coded hierarchical
policy---presumably even more exploration would be required to achieve
this, whereas with state abstractions, MAXQ-Q is able to do slightly
better than our hand-coded policy.  With hierarchical greedy
execution, MAXQ-Q is able to reach the goal using one fewer action, on
the average---so that it approaches the performance of the best
hierarchical greedy policy (as computed by value iteration).  Notice
however, that the best performance that can be obtained by
hierarchical greedy execution of the best recursively-optimal policy
cannot match optimal performance.  Hence, Flat Q learning achieves a
policy that reaches the goal state, on the average, with about one
fewer primitive action.  Finally notice that as in the taxi domain,
there was no added exploration cost for shifting to greedy execution. 

\subsection{Parr and Russell: Hierarchies of Abstract Machines}

In his \citeyear{p-hclmdp-98} dissertation work, Ron Parr considered
an approach to hierarchical reinforcement learning in which the
programmer encodes prior knowledge in the form of a hierarchy of
finite-state controllers called a HAM (Hierarchy of Abstract
Machines).  The hierarchy is executed using a
procedure-call-and-return discipline, and it provides a {\it partial
policy} for the task.  The policy is partial because each machine can
include non-deterministic, ``choice'' machine states, in which the
machine lists several options for action but does not specify which
one should be chosen.  The programmer puts ``choice'' states at any
point where he/she does not know what action should be performed.
Given this partial policy, Parr's goal is to find the best policy for
making choices in the choice states.  In other words, his goal is to
learn a hierarchical value function $V(\mtuple{s,m})$, where $s$ is a
state (of the external environment) and $m$ contains all of the
internal state of the hierarchy (i.e., the contents of the procedure
call stack and the values of the current machine states for all
machines appearing in the stack).  A key observation is that it is
only necessary to learn this value function at choice states
\tuple{s,m}.  Parr's algorithm does not learn a decomposition of the
value function.  Instead, it ``flattens'' the hierarchy to create a
new Markov decision problem over the choice states \tuple{s,m}.
Hence, it is hierarchical primarily in the sense that the programmer
structures the prior knowledge hierarchically.  An advantage of this
is that Parr's method can find the optimal hierarchical policy subject
to constraints provided by the programmer.  A disadvantage is that
the method cannot be executed ``non-hierarchically'' to produce a
better policy.

\begin{figure}[!ht]
{\epsfxsize=6in
\centerps{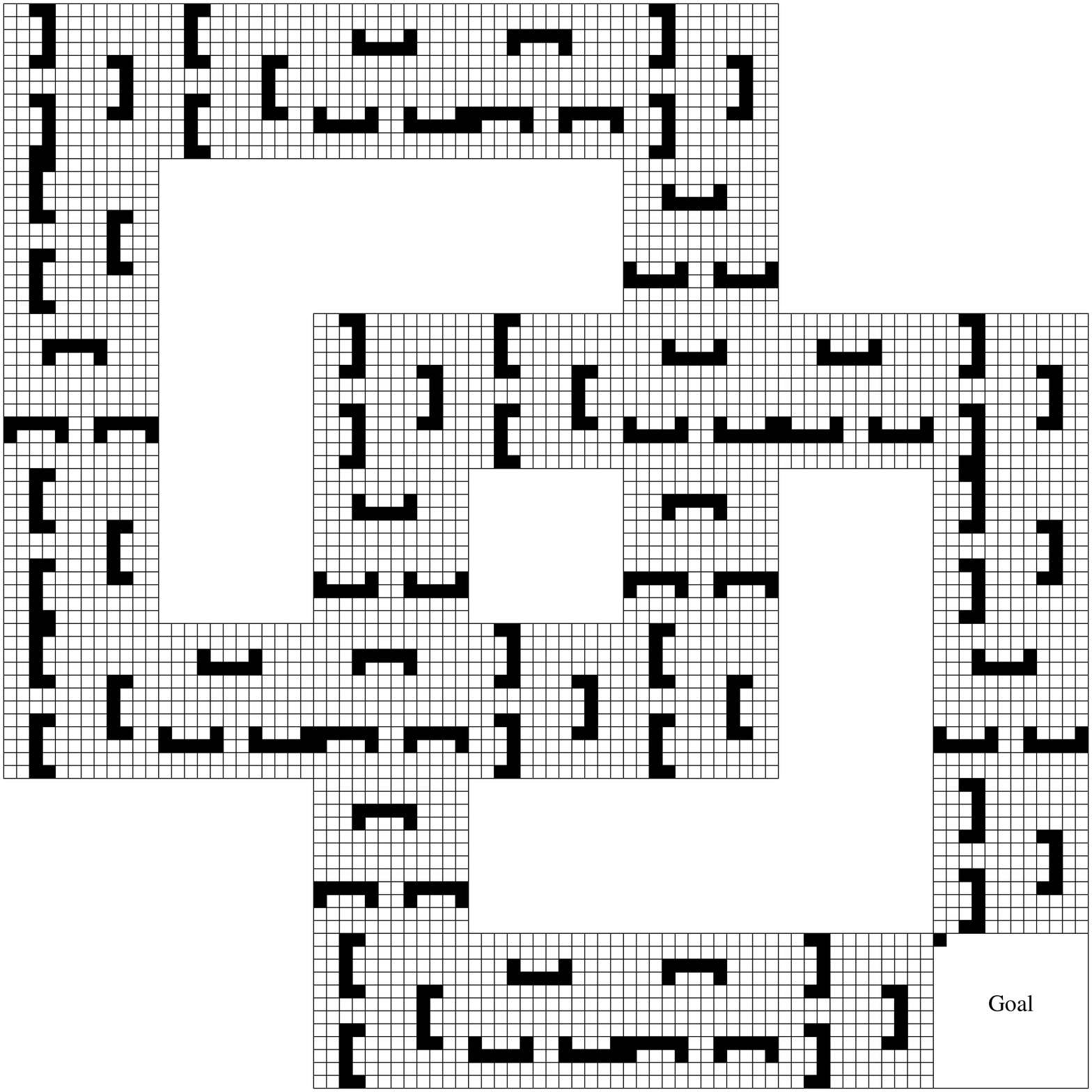}
}
\caption{Parr's maze problem.  The start state is in the upper left
corner, and all states in the lower right-hand room are terminal
states. }
\label{fig-parr-maze}
\end{figure}

Parr illustrated his work using the maze shown in
Figure~\ref{fig-parr-maze}.  This maze has a high-level structure
(i.e., as a series of hallways and intersections), and a low-level
structure (a series of obstacles that must be avoided in order to move
through the hallways and intersections).  In each trial, the agent
starts in the top left corner, and it must move to any state in the
bottom right corner room.  The agent has the usual four primitive
actions, {\sf North}, {\sf South}, {\sf East}, and {\sf West}.  The
actions are stochastic: with probability 0.8, they succeed, but with
probability 0.1 the action will move to the ``left'' and with
probability 0.1 the action will move to the ``right'' instead (e.g., a
{\sf North} action will move east with probability 0.1 and west with
probability 0.1).  If an action would collide with a wall or an
obstacle, it has no effect. 

The maze is structured as a series of ``rooms'', each
containing a 12-by-12 block of states (and various obstacles).  Some
rooms are parts of ``hallways'', because they contain walls on two
opposite sides, and they are open on the other two sides.  Other rooms
are ``intersections'', where two or more hallways meet.  

To test the representational power of the MAXQ hierarchy, we want to
see how well it can represent the prior knowledge that Parr is able to
represent using the HAM.  We begin by describing Parr's HAM for his
maze task, and then we will present a MAXQ hierarchy that captures
much of the same prior knowledge.\footnote{The author thanks Ron Parr
for providing the details of the HAM for this task.}

Parr's top level machine, {\sf MRoot}, consists of a loop with a
single choice state that chooses among four possible child machines:
${\sf MGo}(East)$, ${\sf MGo}(South)$, ${\sf MGo}(West)$, and ${\sf
MGo}(North)$.  The loop terminates when the agent reaches a goal
state.  {\sf MRoot} will only invoke a particular machine if there is
a hallway in the specified direction.  Hence, in the start state, it
will only consider ${\sf MGo}(South)$ and ${\sf MGo}(East)$.

The ${\sf MGo}(d)$ machine begins executing when the agent is in an
intersection.  So the first thing it tries to do is to exit the
intersection into a hallway in specified direction $d$.  Then it
attempts to traverse the hallway until it reaches another
intersection.  It does this by first invoking a ${\sf
ExitIntersection}(d)$ machine.  When that machine returns, it then
invokes a ${\sf MExitHallway}(d)$ machine.  When that machine returns,
{\sf MGo} also returns.

The {\sf MExitIntersection} and {\sf MExitHallway} machines are
identical except for their termination conditions.  Both machines
consist of a loop with one choice state that chooses among four
possible subroutines.  To simplify their description, suppose that
${\sf MGo}(East)$ has chosen ${\sf MExitIntersection}(East)$.  Then the
four possible subroutines are ${\sf MSniff}(East,North)$, ${\sf
MSniff}(East,South)$, ${\sf MBack}(East,North)$, and ${\sf
MBack}(East,South)$.

The ${\sf MSniff}(d,p)$ machine always moves in direction $d$ until it
encounters a wall (either part of an obstacle or part of the walls of
the maze).  Then it moves in perpendicular direction $p$ until it
reaches the end of the wall.  A wall can ``end'' in two ways: either
the agent is now trapped in a corner with walls in both directions $d$
and $p$ or else there is no longer a wall in direction $d$.  In the
first case, the {\sf MSniff} machine terminates; in the second case,
it resumes moving in direction $d$.

The ${\sf MBack}(d,p)$ machine moves one step backwards (in the
direction opposite from $d$) and then moves five steps in direction
$p$.  These moves may or may not succeed, because the actions are
stochastic and there may be walls blocking the way.  But the actions
are carried out in any case, and then the {\sf MBack} machine returns. 

The ${\sf MSniff}$ and {\sf MBack} machines also terminate if they
reach the end of a hall or the end of an intersection.

These finite-state controllers define a highly constrained partial
policy.  The {\sf MBack}, {\sf MSniff}, and {\sf MGo} machines contain
no choice states at all.  The only choice points are in {\sf MRoot},
which must choose the direction in which to move, and in {\sf
MExitIntersection} and {\sf MExitHall}, which must decide when to call
{\sf MSniff}, when to call {\sf MBack}, and which ``perpendicular''
direction to tell these machines to try when they cannot move forward.

\begin{figure}[!ht]
{\epsfxsize=5in
\centerps{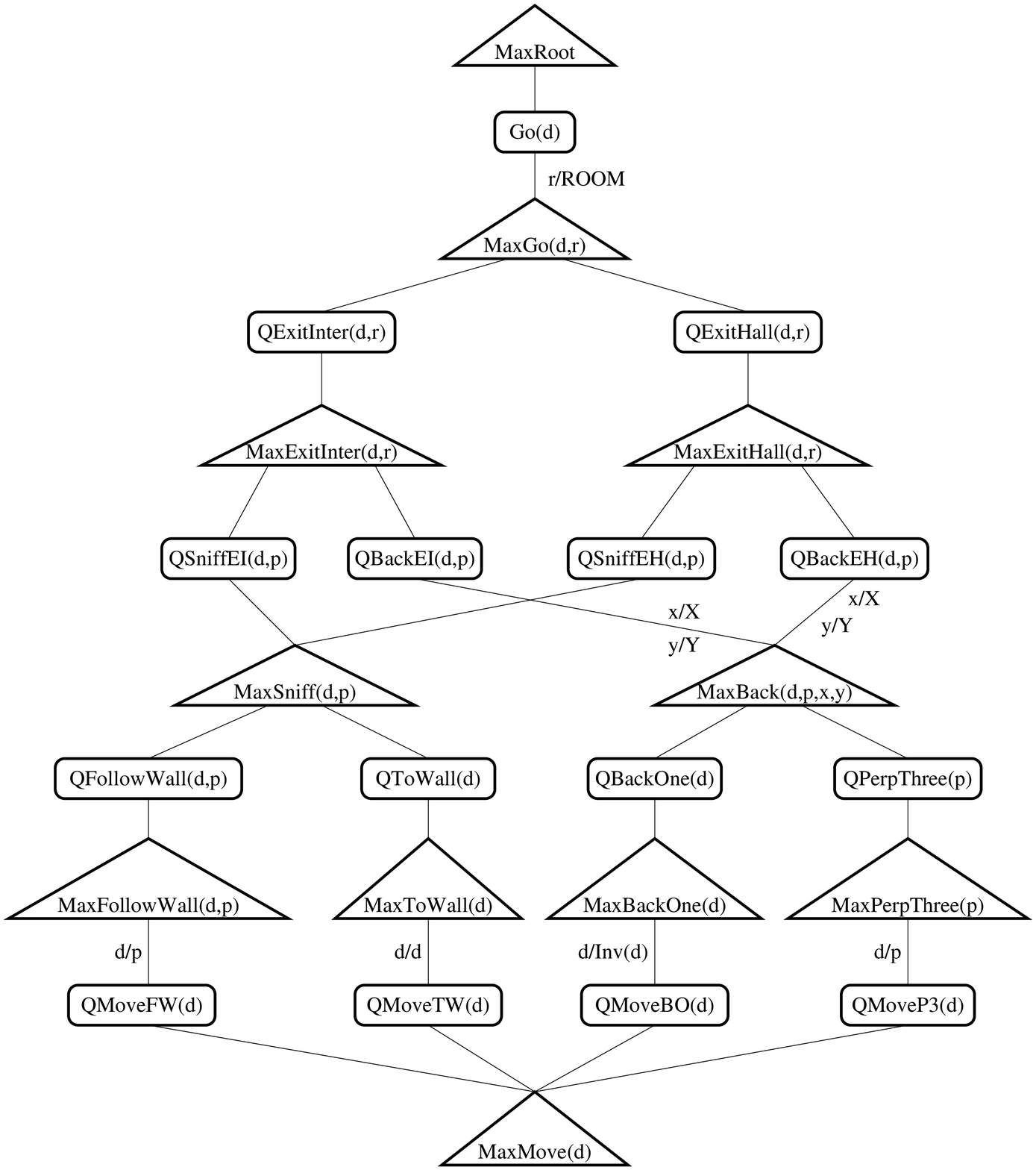}
}
\caption{MAXQ graph for Parr's maze task.}
\label{fig-parr-maxq}
\end{figure}

Figure~\ref{fig-parr-maxq} shows a MAXQ graph that encodes a similar
set of constraints on the policy.  The subtasks are defined as
follows:

\begin{itemize}
\item {\sf Root}.  This is exactly the same as the {\sf MRoot}
machine.  It must choose a direction $d$ and invoke {\sf Go}.  It
terminates when the agent enters a terminal state.  This is also its
goal condition (of course). 

\item ${\sf Go}(d,r)$.  The parameter $r$ is bound to the current
12-by-12 ``room'' in which the agent is located.  {\sf Go}
terminates when the agent enters the room at the end of the hallway in
direction $d$ or when it leaves the desired hallway (e.g., in the
wrong direction).  The goal condition for {\sf Go} is satisfied only
if the agent reaches the desired intersection. 

\item ${\sf ExitInter}(d,r)$.  This terminates when the agent has
exited room $r$.  The goal condition is that the agent exit room $r$
in direction $d$.

\item ${\sf ExitHall}(d,r)$.  This terminates when the agent has
exited the current hall (into some intersection).  The goal condition
is that the agent has entered the desired intersection in direction
$d$.

\item ${\sf Sniff}(d,r)$.  This encodes a subtask that is
equivalent to the {\sf MSniff} machine.  However, {\sf Sniff} must
have two child subtasks, {\sf ToWall} and {\sf FollowWall} that were
simply internal states of {\sf MSniff}.  This is necessary, because a
subtask in the MAXQ framework cannot contain any internal state,
whereas a finite-state controller in the HAM representation can
contain as many internal states as necessary.  In particular, it can
have one state for when it is moving forward and another state for
when it is following a wall sideways.

\item ${\sf ToWall}(d)$.  This is equivalent to part of {\sf MSniff},
and it terminates when there is a wall in ``front'' of the agent in
direction $d$.  The goal condition is the same as the termination
condition. 

\item ${\sf FollowWall}(d,p)$.  This is equivalent to the other part
of {\sf MSniff}.  It moves in direction $p$ until the wall in
direction $d$ ends (or until it is stuck in a corner with walls in
both directions $d$ and $p$).   The goal condition is the same as the
termination condition.

\item ${\sf Back}(d,p,x,y)$.  This attempts to encode the same information
as the {\sf MBack} machine, but this is a case where the MAXQ
hierarchy cannot capture the same information.  {\sf MBack} simply
executes a sequence of 6 primitive actions (one step back, five stes
in direction $p$).  But to do this, {\sf MBack} must have 6 internal
states, which MAXQ does not allow.  Instead, the {\sf Back} subtask is
has the subgoal of moving the agent at least one square backwards and
at least 3 squares in the direction $p$.  In order to determine
whether it has achieved this subgoal, it must remember the $x$ and $y$
position where it started to execute, so these are bound as parameters
to {\sf Back}.  {\sf Back} terminates if it achieves this subgoal or
if it runs into walls that prevent it from achieving the subgoal.  The
goal condition is the same as the termination condition. 

\item ${\sf BackOne}(d,x,y)$.  This moves the agent one step backwards
(in the direction opposite to $d$.  It needs the starting $x$ and $y$
position in order to tell when it has succeeded.  It terminates if it
has moved at least one unit in direction $d$ or if there is a wall in
this direction.  Its goal condition is the same as its termination
condition.

\item ${\sf PerpThree}(p,x,y)$.  This moves the agent three steps in
the direction $p$.  It needs the starting $x$ and $y$ positions in
order to tell when it has succeeded.  It terminates when it has moved
at least three units in the direction $p$ or if there is a wall in
that direction.  The goal condition is the same as the termination
condition. 

\item ${\sf Move}(d)$.  This is a ``parameterized primitive'' action.
It executes one primitive move in direction $d$ and terminates
immediately. 
\end{itemize}

From this, we can see that there are three major differences between
the MAXQ representation and the HAM representation.  First, a HAM
finite-state controller can contain internal states.  To convert them
into a MAXQ subtask graph, we must make a separate subtask for each
internal state in the HAM.  Second, a HAM can terminate based on an
``amount of effort'' (e.g., performing 5 actions), whereas a MAXQ
subtask must terminate based on some change in the state of the world.
It is impossible to define a MAXQ subtask that performs $k$ steps and
then terminate regardless of the effects of those steps (i.e., without
adding some kind of ``counter'' to the state of the MDP).  Third, it
is more difficult to formulate the termination conditions for MAXQ
subtasks than for HAM machines.  For example, in the HAM, it was not
necessary to specify that the {\sf MExitHallway} machine terminates
when it has entered a {\it different} intersection than the one where
the {\sf MGo} was executed.  However, this is important for the MAXQ
method, because in MAXQ, each subtask learns its own value function
and policy---independent of its parent tasks.  For example, without
the requirement to enter a {\it different} intersection, the learning
algorithms for MAXQ will always prefer to have {\sf MaxExitHall} take
one step backward and return to the room in which the {\sf Go} action
was started (because that is a much easier terminated state to reach).
This problem does not arise in the HAM approach, because the policy
learned for a subtask depends on the whole ``flattened'' hierarchy of
machines, and returning to the state where the {\sf Go} action was
started does not help solve the overall problem of reaching the goal
state in the lower right corner.  

To construct the MAXQ graph for this problem, we have introduced three
programming tricks: (a) binding parameters to aspects of the current
state (in order to serve as a kind of ``local memory'' for where the
subtask began executing), (b) having a parameterized primitive action
(in order to be able to pass a parameter value that specifies which
primitive action to perform), and (c) employing ``inheritance of
termination conditions''---that is, each subtask in this MAXQ graph
(but not the others in this paper) inherits the termination conditions
of all of its ancestor tasks.  Hence, if the agent is in the middle of
executing a {\sf ToWall} action when it leaves an intersection, the
{\sf ToWall} subroutine terminates because the {\sf ExitInter}
subroutine has terminated.  If this satisfies the goal condition of
{\sf ExitInter}, then it is also considered to satisfy the goal
condition of {\sf ToWall}.  This inheritance made it easier to write
the MAXQ graph, because the parents did not need to pass down to their
children all of the information necessary to define the complete
termination and goal predicates. 

There are essentially no opportunities for state abstraction in this
task, because there are no irrelevant features of the state.  There
are some opportunities to apply the Shielding and Termination
properties, however.  In particular, ${\sf ExitHall}(d)$ is guaranteed
to cause its parent task, ${\sf MaxGo}(d)$ to terminate, so it does
not require any stored $C$ values.  There are many states where
some subtasks are terminated (e.g., ${\sf Go}(East)$ in any state where
there is a wall on the east side of the room), and so no $C$ values
need to be stored.

Nonetheless, even after applying the state elimination conditions, the
MAXQ representation for this task requires much more space than a flat
representation.  An exact computation is difficult, but after applying
MAXQ-Q learning, the MAXQ representation required 52,043 values,
whereas flat Q learning requires fewer than 16,704 values.  Parr
states that his method requires only 4,300 values.

To test the relative effectiveness of the MAXQ representation, we
compare MAXQ-Q learning with flat Q learning.  Because of the very
large negative values that some states acquire (particularly during
the early phases of learning), we were unable to get Boltzmann
exploration to work well---one very bad trial would cause an action to
receive such a low Q value, that it would never be tried again.
Hence, we experimented with both $\epsilon$-greedy exploration and
counter-based exploration.  The $\epsilon$-greedy exploration policy
is an ordered, abstract GLIE policy in which a random action is chosen
with probability $\epsilon$, and $\epsilon$ is gradually decreased
over time.  The counter-based exploration policy keeps track of how
many times each action $a$ has been executed in each state $s$.  To
choose an action in state $s$, it selects the action that has been
executed the fewest times until all actions have been executed $T$
times.  Then it switches to greedy execution.  Hence, it is not a
genuine GLIE policy.  Parr employed counter-based exploration policies
in his experiments with this task.  For Flat Q learning, we chose the
following parameters: learning rate 0.50, initial value for $\epsilon$
of 1.0, $\epsilon$ decreased by 0.001 after each successful execution
of a Max node, and initial Q values of $-200.123$.  For MAXQ-Q
learning, we chose the following parameters: counter-based exploration
with $T=10$, learning rate equal to the reciprocal of the number of
times an action had been performed, and initial values for the $C$
values selected carefully to provide underestimates of the true $C$
values.  For example, the initial values for {\sf QExitInter} were
$-40.123$, because in the worst case, after completing an {\sf
ExitInter} task, it takes about 40 steps to complete the subsequent
{\sf ExitHall} task and hence, complete the {\sf Go} parent task.

\begin{figure}
\centerps{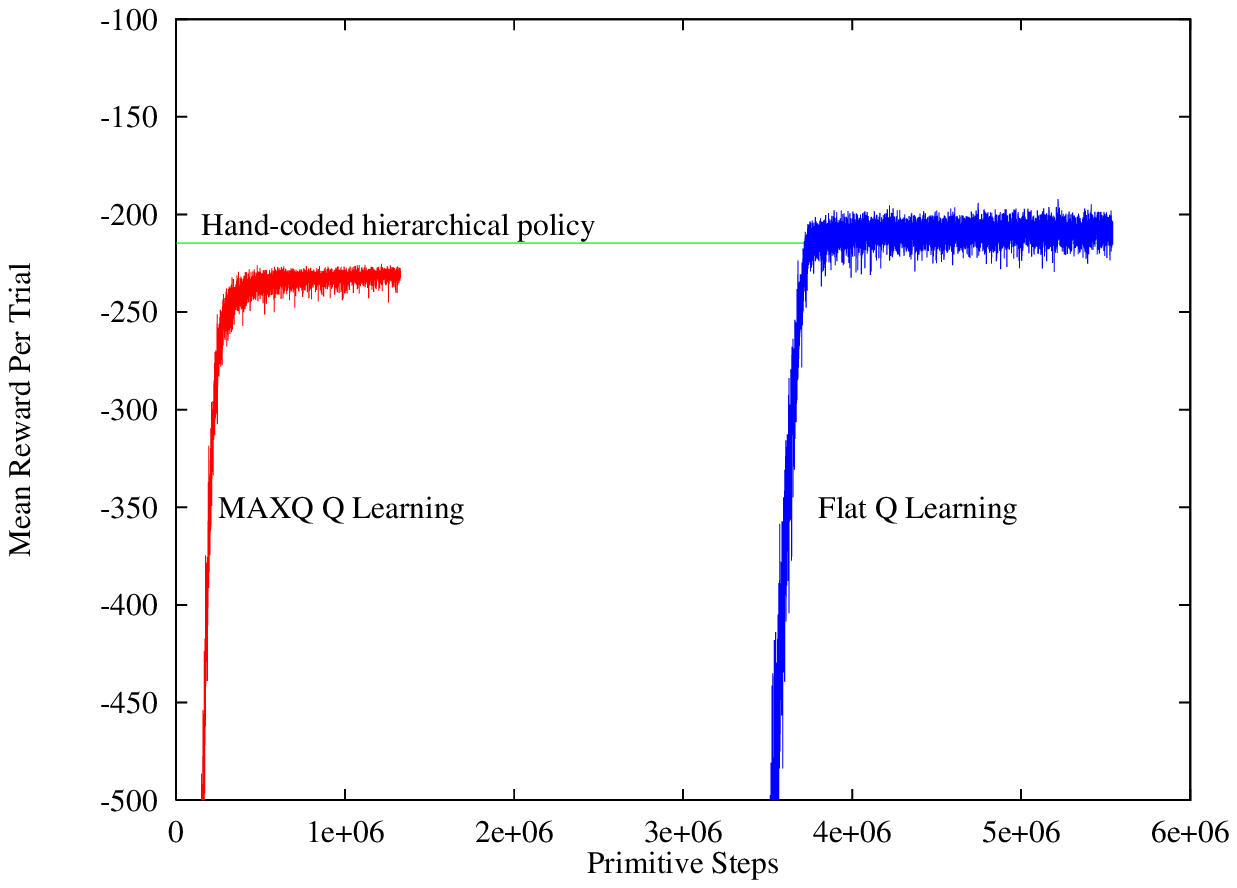}
\caption{Comparison of Flat Q learning and MAXQ-Q learning in the Parr
maze task.}
\label{fig-parr}
\end{figure}

Figure~\ref{fig-parr} plots the results.  We can see that MAXQ-Q
learning converges about 10 times faster than Flat Q learning.  We do
not know whether MAXQ-Q has converged to a recursively optimal policy.
For comparison, we also show the performance of a hierarchical policy
that we coded by hand, but in our hand-coded policy, we used knowledge
of contextual information to choose operators, so this policy is
surely better than the best recursively optimal policy.  HAMQ learning
should converge to a policy equal to or slightly better than our
hand-coded policy. 

This experiment demonstrates that the MAXQ representation can capture
most---but not all---of the prior knowledge that can be represented by
the HAMQ hierarchy.  It also shows that the MAXQ representation
requires much more care in the design of the goal conditions for the
subtasks.  

\subsection{Other Domains}

In addition to the three domains discussed above, we have developed
MAXQ graphs for Singh's \citeyear{mach:Singh:1992} ``flag task'', the
treasure hunter task described by Tadepalli and Dietterich
\cite{td-hebrl-97}, and Dayan and Hinton's \citeyear{dh-frl-93}
Fuedal-Q learning task.  All of these tasks can be easily and
naturally placed into the MAXQ framework---indeed, all of them fit
more easily than the Parr and Russell maze task.

MAXQ is able to exactly duplicate Singh's work and his decomposition
of the value function---while using exactly the same amount of space
to represent the value function.  MAXQ can also duplicate the results
from Tadepalli and Dietterich---however, because MAXQ is not an
explanation-based method, it is considerably slower and requires
substantially more space to represent the value function.

In the Feudal-Q task, MAXQ is able to give better performance than
Feudal-Q learning.  The reason is that in Feudal-Q learning, each
subroutine makes decisions using only a $Q$ function learned at that
level---that is, without information about the estimated costs of the
actions of its descendants.  In contrast, the MAXQ value function
decomposition permits each Max node to make decisions based on the sum
of its completion function, $C(i,s,j)$, and the costs estimated by its
descendants, $V(j,s)$.  Of course, MAXQ also supports non-hierarchical
execution, which is not possible for Feudal-Q, because it does not
learn a value function decomposition.

\section{Discussion: Design Tradeoffs in Hierarchical Reinforcement
Learning}

At the start of this paper, we discussed four issues concerning the
design of hierarchical reinforcement learning architectures.  In this
section, we want to highlight a tradeoff between two of those issues:
the method for defining subtasks and the use of state abstraction.

MAXQ defines subtasks using a termination predicate $T_i$ and a
pseudo-reward function $\tilde{R}$.  There are at least two drawbacks
of this method.  First, it can be hard for the programmer to define
$T_i$ and $\tilde{R}$ correctly, since this essentially requires
guessing the value function of the optimal policy for the MDP at all
states where the subtask terminates.  Second, it leads us to seek a
recursively optimal policy rather than a hierarchically optimal
policy.  Recursively optimal policies may be much worse than
hierarchically optimal ones, so we may be giving up substantial
performance.

However, in return for these two drawbacks, MAXQ obtains a very
important benefit: the policies and value functions for subtasks
become {\it context-free}.  In other words, they do not depend on
their parent tasks or the larger context in which they are invoked.
To understand this point, consider again the MDP shown in
Figure~\ref{fig-two-rooms}.  It is clear that the {\it optimal} policy
for exiting the left-hand room (the {\sf Exit} subtask) depends on the
location of the goal.  If it is at the top of the right-hand room,
then the agent should prefer to exit via the upper door, whereas if it
is at the bottom of the right-hand room, the agent should prefer to
exit by the lower door.  However, if we define the subtask of exiting
the left-hand room using a pseudo-reward of zero for both doors, then
we obtain a policy that is not optimal in either case, but a policy
that we can re-use in both cases.  Furthermore, this policy {\it does
not depend on the location of the goal}.  Hence, we can apply Max node
irrelevance to solve the {\sf Exit} subtask using only the location of
the robot and ignore the location of the goal.

This example shows that we obtain the benefits of subtask reuse and
state abstraction because we define the subtask using a termination
predicate and a pseudo-reward function.   The termination predicate
and pseudo-reward function provide a barrier that prevents
``communication'' of value information between the {\sf Exit}
subtask and its context.  

Compare this to Parr's HAM method.  The HAMQ algorithm finds the best
policy consistent with the hierarchy.  To achieve this, it must permit
information to propagate ``into'' the {\sf Exit} subtask (i.e., the
{\sf Exit} finite-state controller) from its environment.  But this
means that if any state that is reached after leaving the {\sf Exit}
subtask has different values depending on the location of the goal,
then these different values will propagate back into the {\sf Exit}
subtask.  To represent these different values, the {\sf Exit} subtask
must know the location of the goal.  In short, to achieve a
hierarchically optimal policy within the {\sf Exit} subtask, we must
(in general) represent its value function using the {\it entire} state
space.

We can see, therefore, that there is a direct tradeoff between
achieving hierarchical optimality and achieving recursive optimality.
Methods for hierarchical optimality have more freedom in defining
subtasks (e.g., using complete policies, as in the option approach, or
using partial policies, as in the HAM approach).  But they cannot
employ state abstractions within subtasks, and in general, they cannot
reuse the solution of one subtask in multiple contexts.  Methods for
recursive optimality, on the other hand, must define subtasks using
some method (such as pseudo-reward functions) that isolates the
subtask from its context.  But in return, they can apply state
abstraction and the learned policy can be reused in many contexts
(where it will be more or less optimal).

It is interesting that the iterative method described by Dean and Lin
\citeyear{dl-dtpsd-95} can be viewed as a method for moving along this
tradeoff.  In the Dean and Lin method, the programmer makes an initial
guess for the values of the terminal states of each subtask (i.e., the
doorways in Figure~\ref{fig-two-rooms}).  Based on this initial guess,
the locally optimal policies for the subtasks are computed.  Then the
locally optimal policy for the parent task is computed---while holding
the subtask policies fixed (i.e., treating them as options).  At this
point, their algorithm has computed the recursively optimal solution
to the original problem, given the initial guesses.  Instead of
solving the various subproblems sequentially via an offline algorithm,
we could use the MAXQ-Q learning algorithm.

But the method of Dean and Lin does not stop here.  Instead, it
computes new values of the terminal states of each subtask based
on the learned value function for the entire problem.  This allows it
to update its ``guesses'' for the values of the terminal states.  The
entire solution process can now be repeated.  To obtain a new
recursively optimal solution, based on the new guesses.  They prove
that if this process is iterated indefinitely, it will converge to the
recursively optimal policy (provided, of course, that no state
abstractions are used within the subtasks). 

This suggests an extension to MAXQ-Q learning that adapts the
$\tilde{R}$ values online.  Each time a subtask terminates, we could
update the $\tilde{R}$ function based on the computed value of the
terminated state.  To be precise, if $j$ is a subtask of $i$, then
when $j$ terminates in state $s'$, we should update $\tilde{R}(j,s')$
to be equal to $\tilde{V}(i,s') = \max_{a'} \tilde{Q}(i,s',a')$.
However, this will only work if $\tilde{R}(j,s')$ is represented using
the full state $s'$.  If subtask $j$ is employing state abstractions,
$x = \chi(s)$, then $\tilde{R}(j,x')$ will need to be the average
value of $\tilde{V}(i,s')$, where the average is taken over all states
$s'$ such that $x' = \chi(s')$ (weighted by the probability of
visiting those states).  This is easily accomplished by performing a
stochastic approximation update of the form
\[\tilde{R}(j,x') = (1-\alpha_t) \tilde{R}(j,x') + \alpha_t \tilde{V}(i,s')\]
each time subtask $j$ terminates.   Such an algorithm could be
expected to converge to the best hierarchical policy consistent with
the given state abstractions.

This also suggests that in some problems, it may be worthwhile to
first learn a recursively optimal policy using very aggressive state
abstractions and then use the learned value function to initialize a
MAXQ representation with a more detailed representation of the
states.  These progressive refinements of the state space could be
guided by monitoring the degree to which the values of
$\tilde{V}(i,s')$ vary for a single abstract state $x'$.  If they have
a large variance, this means that the state abstractions are failing
to make important distinctions in the values of the states, and they
should be refined.  

Both of these kinds of adaptive algorithms will take longer to
converge than the basic MAXQ method described in this paper.  But for
tasks that an agent must solve many times in its lifetime, it is
worthwhile to have learning algorithms that provide an initial useful
solution but gradually improve that solution until it is optimal.  An
important goal for future research is to find methods for diagnosing
and repairing errors (or sub-optimalities) in the initial hierarchy so
that ultimately the optimal policy is discovered. 

\section{Concluding Remarks}

This paper has introduced a new representation for the value function
in hierarchical reinforcement learning---the MAXQ value function
decomposition.  We have proved that the MAXQ decomposition can
represent the value function of any hierarchical policy under both the
finite-horizon undiscounted, cumulative reward criterion and the
infinite-horizon discounted reward criterion.  This representation
supports subtask sharing and re-use, because the overall value
function is decomposed into value functions for individual subtasks.

The paper introduced a learning algorithm, MAXQ-Q learning, and proved
that it converges with probability 1 to a recursively optimal policy.
The paper argued that although recursive optimality is weaker than
either hierarchical optimality or global optimality, it is an
important form of optimality because it permits each subtask to learn
a locally optimal policy while ignoring the behavior of its ancestors
in the MAXQ graph.  This increases the opportunities for subtask
sharing and state abstraction.

We have shown that the MAXQ decomposition creates opportunities for
state abstraction, and we identified a set of five properties (Max
Node Irrelevance, Leaf Irrelevance, Result Distribution Irrelevance,
Shielding, and Termination) that allow us to ignore large parts of the
state space within subtasks.  We proved that MAXQ-Q still converges in
the presence of these forms of state abstraction, and we showed
experimentally that state abstraction is important in practice for the
successful application of MAXQ-Q learning---at least in the Taxi and
Kaelbling HDG tasks.

The paper presented two different methods for deriving improved
non-hierarchical policies from the MAXQ value function representation,
and it has formalized the conditions under which these methods can
improve over the hierarchical policy.   The paper verified
experimentally that non-hierarchical execution gives improved
performance in the Fickle Taxi Task (where it achieves optimal
performance) and in the HDG task (where it gives a substantial
improvement). 

Finally, the paper has argued that there is a tradeoff governing the
design of hierarchical reinforcement learning methods.  At one end of
the design spectrum are ``context free'' methods such as MAXQ.  They
provide good support for state abstraction and subtask sharing but
they can only learn recursively optimal policies.  At the other end of
the spectrum are ``context-sensitive'' methods such as HAMQ, the
options framework, and the early work of Dean and Lin.  These methods
can discover hierarchically optimal policies (or, in some cases,
globally optimal policies), but their drawback is that they cannot
easily exploit state abstractions or share subtasks.  Because of the
great speedups that are enabled by state abstraction, this paper has
argued that the context-free approach is to be preferred---and that it
can be relaxed as needed to obtain improved policies. 

\subsection*{Acknowledgements}

The author gratefully acknowledges the support of the National Science
Foundation under grant number IRI-9626584, the Office of Naval
Research under grant number N00014-95-1-0557, the Air Force Office of
Scientific Research under grant number F49620-98-1-0375, and the
Spanish Council for Scientific Research.  In addition, the author is
indebted to many colleagues for helping develop and clarify the ideas
in this paper including Valentina Bayer, Leslie Kaelbling, Bill
Langford, Wes Pinchot, Rich Sutton, Prasad Tadepalli, and Sebastian
Thrun.  I particularly want to thank Eric Chown for encouraging me to
study Feudal reinforcement learning, Ron Parr for providing the
details of his HAM machines, and Sebastian Thrun encouraging me to
write a single comprehensive paper.  I also thank the anonymous
reviewers of previous drafts of this paper for their suggestions and
careful reading, which have improved the paper immeasurably.

\bibliography{c:/tex/bib,c:/tex/colt,c:/tex/nn2,c:/tex/neural-computation,c:/tex/ml,c:/tex/aij,c:/tex/nc,c:/tex/PRNN,c:/tex/uai98}
\bibliographystyle{theapa}

\end{document}